\documentclass[10pt,journal,compsoc]{IEEEtran}  
\usepackage{color}
\usepackage{xcolor}

\newcommand{\negexp}[1]{{-#1}}

\ifCLASSOPTIONcompsoc
  \usepackage[nocompress]{cite}
\else
  \usepackage{cite}
\fi
\ifCLASSINFOpdf
\else
\fi
\usepackage{amsmath}
\usepackage{amssymb}
\usepackage{braket}
\usepackage{quantikz}
\usepackage{graphicx}
\usepackage{tabularx}
\usepackage{url}
\usepackage{tikz}
\usepackage{pgfplots}
\usetikzlibrary{datavisualization}
\usetikzlibrary{datavisualization.formats.functions}
\usepackage{mathtools}
\usepackage{blkarray,booktabs,bigstrut}
\usepackage{etoolbox}
\usepackage{nicematrix}

\begin{document}

\renewcommand{\arraystretch}{2}

\title{Quantum-Inspired Machine Learning: a Survey}

\author{Larry~Huynh, Jin~Hong, Ajmal~Mian, Hajime~Suzuki, Yanqiu~Wu, and Seyit~Camtepe
\thanks{L. Huynh, J. Hong, and A. Mian are with the University of Western Australia. Emails: \{larry.huynh, jin.hong, ajmal.mian\}@uwa.edu.au.}
\thanks{H. Suzuki, Y. Wu, and S. Camtepe are with CSIRO's Data61, Marsfield, NSW, Australia. Emails: hajime.suzuki@data61.csiro.au, autumn.wu@data61.csiro.au, seyit.camtepe@data61.csiro.au.}
}

\IEEEtitleabstractindextext{
\begin{abstract}
Quantum-inspired Machine Learning (QiML) is a burgeoning field, receiving global attention from researchers for its potential to leverage principles of quantum mechanics within classical computational frameworks. However, current review literature often presents a superficial exploration of QiML, focusing instead on the broader Quantum Machine Learning (QML) field. In response to this gap, this survey provides an integrated and comprehensive examination of QiML, exploring QiML's diverse research domains including tensor network simulations, dequantized algorithms, and others, showcasing recent advancements, practical applications, and illuminating potential future research avenues. Further, a concrete definition of QiML is established by analyzing various prior interpretations of the term and their inherent ambiguities. As QiML continues to evolve, we anticipate a wealth of future developments drawing from quantum mechanics, quantum computing, and classical machine learning, enriching the field further. This survey serves as a guide for researchers and practitioners alike, providing a holistic understanding of QiML's current landscape and future directions.
\end{abstract}


}

\maketitle
\IEEEdisplaynontitleabstractindextext
\IEEEpeerreviewmaketitle


\IEEEraisesectionheading{\section{Introduction}\label{sec:introduction}}

\begin{figure}[]
    \centering
    \hspace*{-0.3in}
    \centerline{\includegraphics[width=0.4\textwidth]{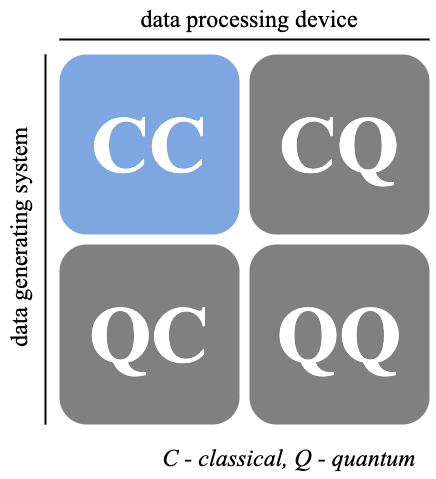}}
    \caption{Four approaches to QML \cite{schuld2021machine}, based on whether quantum or classical data/processing is used. QiML methods describe the “CC” mode (blue shading).}
    \label{fig:intro_intersections}
\end{figure}

\IEEEPARstart{T}{he} field of Quantum-Inspired Machine Learning (QiML) has seen substantial growth, garnering interest from researchers globally. A specialized subset of Quantum Machine Learning (QML), QiML focuses on developing 
classical machine learning algorithms 
inspired by principles of quantum mechanics within a classical computational framework, commonly referenced as the “classical-classical” quadrant of QML categorization as shown in Figure \ref{fig:intro_intersections}. QiML represents a multifaceted research domain, with investigations pushing to exceed conventional, classical state-of-the-art results, or exploring the expressivity provided by quantum formulations. 

To situate QiML within the context of QML, we briefly expound upon the latter. QML, more broadly, sits at the fascinating intersection of quantum computing and machine learning.
The dominant research field concerns the “classical-quantum” domain, and explores the use of quantum hardware to accelerate and enhance machine learning strategies. Here, two challenges present in classical machine learning are addressed. First, the increasing size and complexity of datasets in many fields have created computational challenges that classical machine learning struggles to manage efficiently. Secondly, quantum computing offers the potential to solve complex problems that are currently infeasible with classical computation methods \cite{ciliberto2018quantum}. Practical evaluation of QML algorithms on actual quantum hardware, however, is currently limited by factors such as the limited number of qubits, high error rates in quantum gates, difficulty in maintaining quantum states (decoherence), and challenges associated with quantum error correction \cite{cerezo2022challenges}. As a result, the QML landscape has been primarily shaped by theoretical considerations, with recent advancements in noisy-intermediate scale quantum (NISQ) devices providing an early, empirical glimpse into the potential of full-scale quantum computing \cite{preskill2018quantum}. As such, the true extent and impact of QML on the machine learning landscape remains an ongoing research topics.

QiML has evolved in tandem with QML research. Instances of often cited research domains include tensor network quantum simulations and dequantized algorithms \cite{khan2020machine, schuld2021machine}. However, in contrast with QML, discoveries in QiML are frequently backed by numerical evidence, facilitated by the independence from quantum hardware constraints, thereby enabling easier quantitative evaluations compared to other QML subsets. While QiML research is flourishing, current survey literature often neglects this field, with a larger focus given to QML as a whole. Often, QiML is only briefly mentioned or treated superficially \cite{garg2020advances, abohashima2020classification, mishra2021quantum, houssein2022machine, khan2020machine, zeguendry2023quantum}. Practical use cases of QiML, their applications, and comparative analyses with standard classical benchmarks often remain unexplored. This points to a crucial need for a standalone, in-depth review of QiML as a distinct field.

Responding to this literature gap, our survey aims to provide a comprehensive, integrated discussion on the various facets of QiML. We aim to provide an accessible and comprehensive overview of how QiML is used in practice, detailing its recent advancements and giving readers an understanding of the field's progression. The reader should note that while exploring QiML methods from the lens of quantum mechanics and categorizing methods based on sources of inspiration would be of interest, this survey approaches the field from an applications perspective. The contributions of this survey are to provide an overview of the progression of QiML and its research directions in recent years, and to identify the future directions of QiML research. Specifically, they are:

\begin{itemize}
    \item To highlight and classify existing QiML methods;
    \item To establish a concrete definition of QiML, accounting for its multi-directional research trends;
    \item To discuss the practical applications of these methods, specifically identifying the tasks to which QiML techniques have currently been applied;
    \item To discuss the limiting factors of QiML in practice, and;  
    \item To explore and discuss the potential future directions of QiML research.
\end{itemize}

\section{Quantum-Inspired Machine Learning}\label{sec:def}

In this section, we dissect the QiML term into its constituent parts --- “quantum-inspired” and “machine learning” --- for a comprehensive understanding. Following this, we unpack the QiML term itself, integrating our analysis of its individual components. Our goal is to address inconsistencies in past literature, identify common threads, and propose a precise definition for QiML. This serves as a reliable compass, guiding future research in this dynamic field.

\subsection{\textit{“Quantum-Inspired”}}

The term “quantum-inspired” was introduced by Moore and Narayanan \cite{moore1995quantum} in the context of computing for the first time in 1995. The term was used to differentiate between two types of computational methods: “pure” quantum computation and quantum-inspired computation. The former is firmly rooted in quantum mechanical concepts, such as standing waves, interference, and coherence, and can only be executed on a quantum computer. On the other hand, “quantum-inspired” computing refers to practical methods that have been derived from these concepts. These methods do not require a quantum computer, but rather utilize classical computers or algorithms to simulate quantum effects and achieve computational advantages. The categorization of these two types of methods was significant at the time, since quantum computing methods were not yet practically realizable due to the technological inability of implementing stable quantum systems with robust error correction \cite{steane1998quantum}. The potential of quantum computing was known and acknowledged, as well as the challenges of practical implementation; many pure quantum algorithms still currently operate at a theoretical level \cite{cerezo2022challenges}. This led to research efforts in exploring the utilization of quantum mechanics in classical computing. 


Han and Kim were among the pioneers giving name to quantum-inspired algorithms, and proposed the “quantum-inspired evolutionary algorithm” (QIEA). Extending upon prior works \cite{narayanan1996quantum, han2000genetic}, QIEAs describe evolutionary algorithms that are inspired by quantum mechanical concepts, specifically employing “Q-bits” and “Q-gates” to model populations and evolutionary processes \cite{han2002quantum}. These are the classical analogues of quantum qubits and quantum gates; Q-bits serve as probabilistic representations that maintain population diversity among individuals through the linear superposition of states, while Q-gates act as variational operators, driving individuals towards an optimal solution by modifying the probability distributions associated with Q-bits. Since these operate on classical computers, quantum phenomena is not observed, e.g. state collapse does not occur when the Q-bit state is “measured”. Nevertheless, significant performance improvements were observed when compared to its classical counterpart, highlighting the benefits of using quantum-inspired methods. QIEAs form a subset of quantum-inspired metaheuristics, which are optimization techniques developed for finding approximate or near-optimal solutions inspired by quantum mechanics principles but are implemented on classical computers. This research area has demonstrated notable advancements in both performance and computational efficiency compared to classical methods over recent decades \cite{zhang2011quantum, manju2014applications, varmantchaonala2022quantum, ross2019review}. 
Quantum-inspired Genetic Algorithms (QGA) \cite{narayanan1996quantum, han2001parallel} also belong to this category, employing Q-bits for probabilistic solution encoding and Q-gates for genetic operations like crossover and mutation. These features enable QGAs to maintain greater population diversity \cite{han2002quantum}. Quantum-inspired Particle Swarm Optimization (QPSO) \cite{sun2004particle} represents particles as Q-bits, allowing for a probabilistic representation of the solution space instead of fixed positions and velocities. By maintaining particles in a superposition of states, the search space diversity and exploration capabilities are enhanced. Similarly, quantum-inspired ant colony optimization (QACO) \cite{wang2007novel} utilizes Q-bits and Q-gates to improve the classical ACO's pheromone update mechanism. The traversal of the solution space via these mechanisms has been shown to augment exploration and exploitation capabilities \cite{yongjun2017application, das2023quantum}. Quantum Simulated Annealing (QSA) \cite{somma2007quantum} employs Monte Carlo methods to efficiently simulate classical annealing processes. Additional quantum concepts are also utilized, such as quantum random walks for efficient exploration of the energy landscape, quantum phase estimation for optimizing the annealing schedule, and quantum tunneling to escape local minima.

It is important to note that the research goals in developing quantum-inspired methods can vary. Typically, the objectives include achieving faster and more stable convergence, enhancing the effectiveness of the solution search, or a combination of both \cite{gharehchopogh2022quantum}. 



Some works also introduce unique additions to their specification of “quantum-inspired”. Moore and Narayanan \cite{moore1995quantum} further characterized quantum-inspired computing algorithms by stipulating that their output need only be verified by classical computing methods, and not requiring a quantum computer for the task. Manju and Nigam \cite{manju2014applications} constrain quantum-inspired computing to methods for solving engineering problems with cyclic or recurrent behavior. Further, some studies emphasize the use of quantum bits or Q-bits as the defining aspect of being quantum-inspired \cite{gharehchopogh2022quantum, dos2012nuclear}, while others characterize different quantum concepts to be the core underpinning of the “quantum inspired” definition. These minor distinctions in defining “quantum inspired” do not significantly impact the overall understanding of the term, as the core concept remains consistent with the generalization outlined earlier. 


\subsection{\textit{“Machine Learning”}}

Machine learning, as a field, has evolved significantly since its inception, with its definition and scope adapting to reflect advancements in techniques and computational capabilities. Initially conceptualized in the 1950s as the concept of “programming computers to learn from experience” by Arthur Samuel \cite{samuel1959machine}, machine learning has witnessed several transformative milestones. Throughout these developments, the core idea of learning from data to make predictions or decisions has persisted, although the specific models, techniques, and learning paradigms have diversified over time. Several foundational definitions of machine learning have been offered, each emphasizing various aspects of what constitutes the field. For example, \cite{mitchell2007machine} highlights the importance of learning from experience and improving performance over time, thus underscoring the iterative and adaptive nature of machine learning. Goodfellow et al. \cite{goodfellow2016deep} position machine learning as a sub-field of artificial intelligence, concerned with building algorithms that rely on a collection of examples of some phenomenon to be useful. A few authors define machine learning rather as a broad collection of algorithms that learn patterns over feature spaces \cite{murphy2012machine, wittek2014quantum}. Clear commonalities have been agreed upon, such as the importance of learning from patterns inherent within data \cite{mitchell2007machine, goodfellow2016deep} via automatic processes \cite{domingos2012few, murphy2012machine} without explicit programming \cite{mahesh2020machine}, and the ability to improve performance based on the experience or data it is exposed to. While the specific techniques and approaches within machine learning have evolved and diversified, its fundamental definition has remained consistently cohesive and integral to the field.


\subsection{\textit{“Quantum-Inspired Machine Learning”}}\label{sec:def_qiml}

Both quantum computing and machine learning have gained tremendous popularity in the past couple of decades with significant advances made compared to when they were first introduced. More recently, researchers have turned their attention to an inner subset at the intersection of these fields; “quantum-inspired machine learning” (QiML). The interpretation of this term by researchers, however, has varied significantly in the literature. Hence, this subsection aims to explore different perspectives and approaches taken by researchers in their attempts to define “quantum-inspired machine learning,” shedding light on the nuances and challenges involved in characterizing this rapidly developing field. We will thus also argue for a concrete description of the quantum-inspired machine learning term, which will better promote clarity, and assist in characterizing methodologies within this, and related domains.

The terms “quantum-inspired”, or “quantum-like” machine learning have, in early reviews, focused on describing optimization techniques inspired by quantum phenomena, and run on classical computers \cite{wittek2014quantum}, likely in the absence of parameterized, iterative pattern recognition models more akin to classical machine learning methods. Other authors have corroborated this idea, with more explicit mention of machine learning rather than optimization \cite{abohashima2020classification, garg2020advances, houssein2022machine, zeguendry2023quantum}. These definitions are consistent with the expected combination of the individual terms “quantum-inspired” and “machine learning”. In recent years, the umbrella of QiML has been extended to include tensor network machine learning models that parallel classical methodologies \cite{carrasquilla2020machine, alchieri2021introduction}, as well as "dequantized" algorithms, which aim to develop classical analogs of quantum algorithms with comparable computational advantages \cite{schuld2021machine}. Although tensor network techniques agree with the concept of quantum inspiration (given that tensor decompositions aspire to model complex quantum wavefunctions), the classification of dequantized algorithms as "quantum-inspired" may not wholly align with prevailing definitions of QiML. In the case of dequantized algorithms, the "inspiration" derives not from quantum mechanics itself but from the scrutiny of claims of quantum supremacy, or rather by the quantum algorithms themselves. In this sense, the relationship to quantum mechanics is indirect; the focus is on understanding the potential of quantum algorithms in classical settings.


While the prevailing definitions of QiML offer some flexibility and accommodate a variety of quantum applications to machine learning, they may also inadvertently lead to imprecise categorizations. Notably, certain techniques might be labeled as QiML when they may not accurately belong to this domain. Consider, for example, a quantum kernel used in a quantum support vector machine or a quantum variational circuit, methods typically relying on quantum circuit implementation, that has been simulated on classical hardware, as in \cite{wu2021application, riaz2023accurate} using tools like Qiskit \cite{Qiskit} or PennyLane \cite{bergholm2018pennylane}. These algorithms are now implemented on classical hardware and do not necessarily require quantum hardware, yet are often considered part of the broader quantum machine learning (QML) context, not QiML. The main reason for this distinction is not immediately clear, but may be related to the nature of the algorithms themselves, and that quantum circuit implementations have been colloquial considered as QML, regardless of implementation device. Another explanation could be tied to the efficiency of implementation: if the translation of a quantum computing process to a classical setting is not intrinsically efficient, regardless of the circuit or input size, and does not scale beyond the limits of classical computation, the technique could be more appropriately classified as QML. 

Addressing these inconsistencies is crucial for a more accurate understanding of the evolving landscape of QiML, as it will allow researchers to effectively build upon previous work, avoid confusion, and foster more precise communication within the academic community. We now attempt to pin down an appropriate definition of QiML. In doing so, the terms QML and QiML warrant clarification to better understand their roles in the interdisciplinary area between quantum computing and machine learning. Broadly, QML represents the integration of quantum computing and machine learning principles, forming an umbrella term that includes related concepts within this intersection, one of which is QiML. To gain a deeper understanding of the broader QML landscape, we examine a typology introduced by Schuld and Petruccione \cite{schuld2021machine}, depicted in Figure \ref{fig:intro_intersections}. This typology categorizes QML based on whether the data source is a classical (C) or quantum (Q) system and whether the data processing device is classical (C) or quantum (Q). Four distinct categories emerge, each highlighting a different aspect of the interplay between quantum computing and machine learning:

\begin{itemize}
    \item \textbf{CC:} Classical data and classical processing. This category is where QiML primarily resides. Methods in this category are inspired by quantum mechanics but still use classical data and processing.
    \item \textbf{QC:} Quantum data and classical processing. In this category, machine learning techniques are used to analyze quantum data or measurement outcomes from quantum systems and experiments.
    \item \textbf{CQ:} Classical data and quantum processing. Quantum computing is utilized to process conventional data, often with the objective of developing quantum algorithms for data mining.
    \item \textbf{QQ:} Quantum data and quantum processing. This category explores processing quantum data with quantum devices, either by inputting experimental measurements into a quantum computer or using a quantum computer to simulate and subsequently analyze the behavior of quantum systems.
\end{itemize}

These four categories have been corroborated and utilized in various studies within the field of quantum machine learning, as demonstrated in the literature \cite{aimeur2006machine, cerezo2022challenges, zeguendry2023quantum}. Some researchers have also proposed a contemporary categorization scheme that reflects the evolving landscape of QML \cite{abohashima2020classification, zeguendry2023quantum}. Here, QML is divided into three distinct categories:
\begin{itemize}
    \item \textbf{Quantum Machine Learning}: all quantum adaptations of classical ML algorithms that necessitate quantum computation for their execution.
    \item \textbf{Quantum-inspired Machine Learning}: the integration of quantum computing concepts to enhance traditional machine learning algorithms, without requiring actual quantum computation.
    \item \textbf{Hybrid Classical-Quantum Machine Learning}: the fusion of classical and quantum algorithms, aiming to optimize performance and minimize learning costs by exploiting the strengths of both approaches.
\end{itemize}

These perspectives are consistent with prior QiML definitions, while also specifying classical computation as a crucial component. To consolidate this aspect with conventional understanding, and also accommodate for the existence of dequantized algorithms, we now propose a concrete definition of “quantum-inspired machine learning”. 

{\bf Definition 1:} Quantum-Inspired Machine Learning (QiML) refers to machine learning algorithms that draw inspiration from principles of quantum mechanics or quantum computing constructs, but do not necessitate quantum processing and can be executed on classical hardware.

This definition encapsulates the following pivotal aspects:
\begin{itemize}
\item The foundational principles of machine learning;
\item Inspiration from quantum phenomena or quantum computing, including quantum algorithms, thus acknowledging the significance of dequantized algorithms within QiML;
\item The capacity for problem representation and computation on classical hardware, thereby incorporating the ability to simulate quantum hardware.
\end{itemize}

Our aim is to provide clarity and guidance for future research in this rapidly evolving field. This definition acknowledges the growing intersection between quantum computing and machine learning and fosters a more focused and constructive discourse within the QiML landscape.








\section{Selection Criteria}

This review aims to collate and analyze recent advancements in the rapidly evolving field of quantum-inspired machine learning. We focus on contemporary studies; only studies published in the recent years are considered (2017-2023). A systematic search strategy was employed to retrieve literature from multiple databases, including Google Scholar, and other academic search engines. The search was performed using the specific key phrases, in order to refine the search space, as these research areas have been considered QiML in the literature:

\begin{itemize}
    \item “quantum-inspired”,
    \item “dequantized algorithms”,
    \item “tensor networks”,
    \item “variational quantum algorithms”
\end{itemize}
We combine each of these terms with “machine learning” to ensure the retrieval of publications specifically relevant to QiML. We also exclude terms that may capture works rooted in combinatorial optimization, using the key phrase “-‘combinatorial optimization’” and other related search terms such as “-‘heuristic algorithms’”, “-‘optimization algorithms”, “-‘combinatorial algorithms”, with additional manual vetting of papers that bypass this filter --- in this review, we will consider such methods disjoint from machine learning, and more aligned with the field of metaheuristics. Initial search returned 2,300 results. We also use the key phrases “dequantized algorithms”, “tensor networks” and “variational quantum algorithms” in order to refine the search space, as these research areas have been considered QiML in the literature. 

\begin{figure}[!ht]
    \centering
    \centerline{\includegraphics[width=0.5\textwidth]{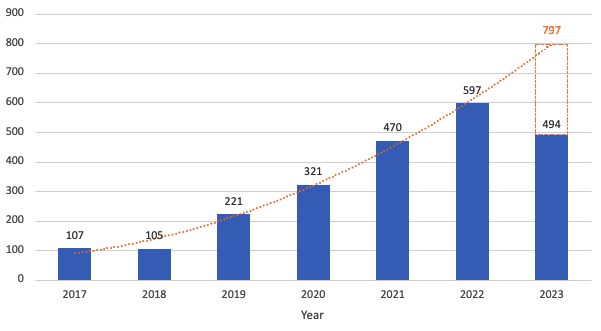}}
    \caption{Yearly plot displaying the frequency of works found via the search term “‘quantum-inspired’ AND ‘machine learning’ -‘combinatorial optimization’”}
    \label{fig:sc_timeline}
\end{figure}

Categorizing the selected studies was based on the following criteria. First, papers are categorized based on the types of techniques involved in accomplishing machine learning tasks. It focused on the various quantum-inspired algorithms, methodologies, and models used in these studies and their unique aspects that contribute to the advancement of machine learning tasks. Secondly, the applicability and practical implementation of these quantum-inspired methods is of importance. As a fast-growing field, it is essential to discern not only the theoretical advancements but also where these methods have been applied in empirical experimentation. Identifying such applications provides insights into the current state of quantum-inspired machine learning and its potential in solving complex problems across various domains; works that demonstrate these real-world applications and emphasize the practical benefits and challenges of quantum-inspired techniques were given priority. Works that introduced novel methods, algorithms, or theoretical insights into QiML were also particularly valued.

\section{Current QiML Techniques}\label{sec:qiml}

\begin{figure*}[!ht]
    \centering
    \centerline{\includegraphics[width=0.95\textwidth]{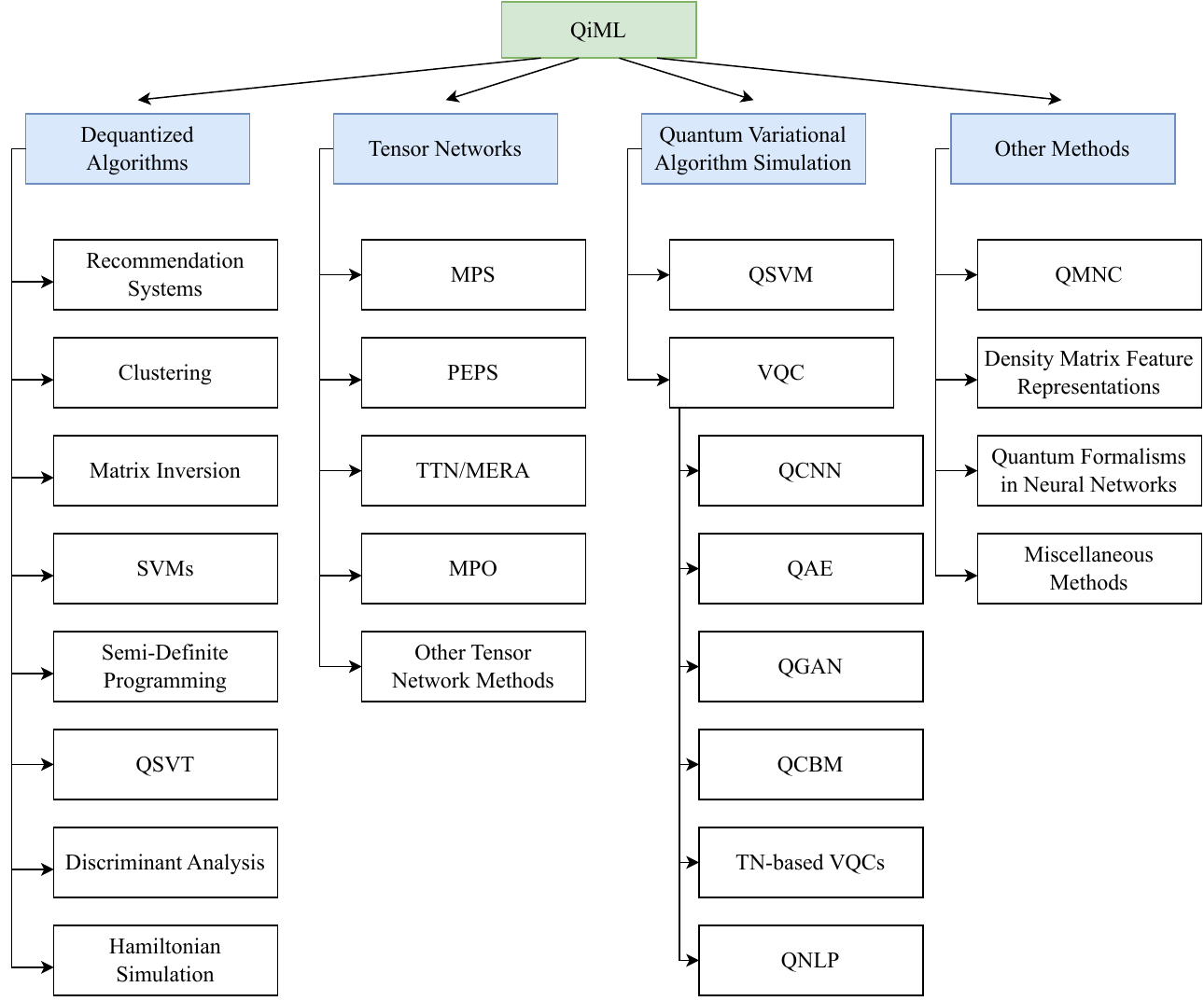}}
    \caption{Categories of QiML Methods}
    \label{fig:qiml_categories}
\end{figure*}

We classify works in QiML based on their underlying methodologies and purposes. Three overarching categories have been identified: “Dequantized Algorithms” (Section \ref{sec:dequantized_algorithms}), “Tensor Networks” (Section \ref{sec:tensor_networks}), and “Quantum Variational Algorithm Simulation” (Section \ref{sec:quantum_variational_algorithm_simulation}). Methods that do not fail into these categories are grouped and labeled as “Other QiML Methods” (Section \ref{sec:other}). Within each category, methods can be further grouped by their application domains, with the exception of Dequantized Algorithms, which are grouped via method due to lack of practical application domains. Figure \ref{fig:qiml_categories} presents these categorizations in an organizational chart.

\subsection{Dequantized Algorithms}\label{sec:dequantized_algorithms}
\begin{table}[!ht]
    \centering
    \caption{Variable Descriptions for Dequantized Algorithms}
    \begin{tabularx}{0.4\textwidth}{|c|X|}
        \hline
        \textbf{Variable} & \textbf{Description} \\
        \hline
        $\|A\|_F$ & Frobenius norm of matrix $A$ \\
        \hline
        $\|A\|_2$ & Spectral norm of $A$ \\
        \hline
        $\|A\|$ & Operator norm of $A$ \\
        \hline
        $\|v\|$ & L2 norm of vector $v$ \\
        \hline
        $k$ & Rank of matrix $A$ \\
        \hline
        $\kappa$ & Condition number of $A$ \\ 
        \hline
        $d$ & Degree of the polynomial associated with $A$ in the QSVT \\
        \hline
        $\sigma$ & Singular value threshold \\
        \hline
        $\varepsilon$ & Approximation error \\
        \hline
    \end{tabularx}
    \label{tab:legend}
\end{table}
In recent years, algorithms termed as “dequantized” have been developed. These algorithms aim to determine whether the speedups claimed by quantum machine learning algorithms are genuinely attributed to the inherent power of quantum computation or are merely a byproduct of strong assumptions regarding input and output encoding/decoding (i.e. via state preparation). By scrutinizing these assumptions and their implications, researchers can more accurately assess the practicality of quantum algorithms and identify potential drawbacks that may impede their real-world applications. This line of inquiry has lead to the identification of classical counterparts to quantum-based machine learning methods, which can be implemented on classical hardware and achieve performance levels comparable to their quantum analogs. By designing algorithms that can be efficiently executed on classical resources, the costly and arguably impractical requirements for quantum state preparation and hardware can both be circumvented \cite{tang2019overviewblog}.

Ewin Tang presented the seminal work in this area in 2019 \cite{tang2019quantum}. In 
an 
attempt to prove that no classical algorithm could match the runtime of the quantum recommendation system developed by Kerenidis and Prakash \cite{kerenidis2016quantum}, the author was instead able to devise a classical algorithm that matched the fast runtime. The classical algorithm ran in time $O(poly(k)log(mn))$, only polynomially slower than the quantum algorithm's $O(poly(k)polylog(mn))$, thus there is no quantum advantage observed. This was a significant result, as this quantum algorithm was once thought to be one of the most promising candidates for demonstrably exponential improvements in quantum machine learning \cite{preskill2018quantum}. The key observation is that the exponential speedup achieved by the quantum algorithm relies on specific input assumptions about the user-product preference matrix. These were, in general, the prevailing assumptions for many quantum machine learning algorithms at the time; that either computing the corresponding quantum state $\ket{v}$ from some input vector $v$ is arbitrarily fast, or that the necessary quantum states come into the system already prepared. However, the author emphasizes that the cost of state preparation is nontrivial; if state preparation cannot be performed in poly-logarithmic time, then the claimed exponential speedup of the associated quantum algorithm can not be realized in practice, as it does not account for the time constraints imposed by state preparation.

\begin{figure*}[!ht]
    \centering
    \centerline{\includegraphics[width=1\textwidth]{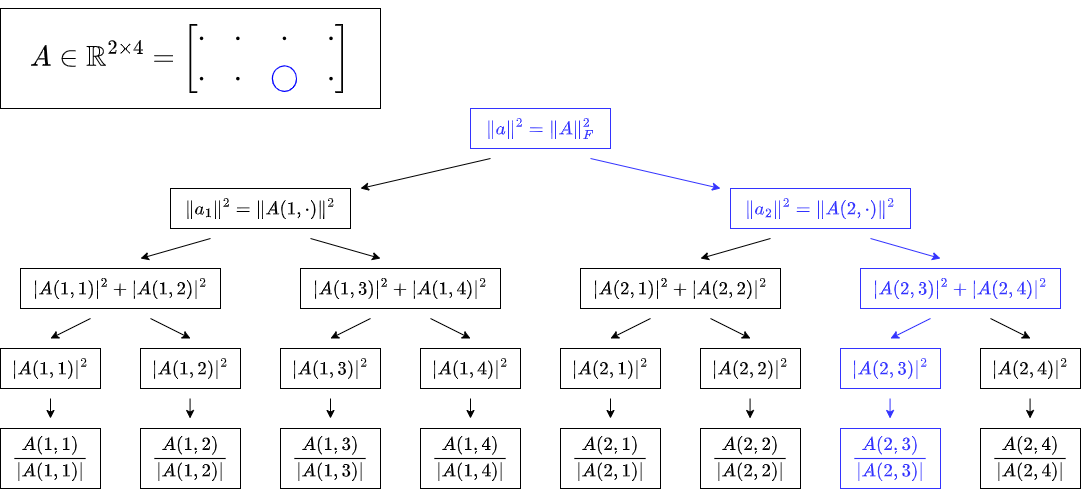}}
    \caption{QRAM-like Data Structure for a matrix $A \in \mathbb{C}^{2 \times 4}$ \cite{chia2022sampling}. Sampling from this structure involves traversing down the tree based on L2-norm probability. The terminating nodes contain the sign of each element $A(i, j)$. An example is shown for $A(2, 3)$ (blue highlighting), sampled with probability $\|a_2\|^2 / \|A\|^2_F \times |A(2, 3)|^2 / \|a_2\|^2$.}
    \label{fig:qram}
\end{figure*}

The quantum recommendation system by Kerenidis and Prakash \cite{kerenidis2016quantum} describes an explicit data structure used to quickly prepare quantum states, as seen in Figure \ref{fig:qram}. This is implemented via a set of binary search trees, one for each row (user) in the preference matrix. Quantum states are encoded using the square amplitudes of the coefficients of the given row vector $\ket{v}$ in a hierarchical manner: $\ket{v} = \sum_{i=1}^n v_i \ket{i}$ for $v \in \mathbb{C}^n$. The amplitudes and signs are stored as leaf nodes in the tree; the root node contains $\|v\|^2$. This structure can be seen as a classical analogue to QRAM \footnote{We disregard the exact notion of QRAM for quantum settings, as the data structure is only interfaced with classically; no quantum operations are involved.}. Sampling from this requires randomly traversing down through each subsequent child node with probability proportional to its weight. This data structure allows for $O(log(mn))$ query access, as well as time $O(log^2(mn))$ for online updates to preference matrix elements $A_{ij}$. However, the model assumes quantum access to this data structure with prepared quantum states. Tang demonstrated that this data structure employed to meet the state preparation assumptions can also fulfill classical L2-norm sampling assumptions, allowing for “sample and query” access (SQ access) to the data. As a result, a classical algorithm aiming to “equal” the performance of the quantum algorithm can take advantage of these assumptions. In this way, more reasonable comparisons can be made between QML algorithms with state preparation assumptions and classical counterparts with sampling assumptions.

Specifically, SQ access to $x$ --- $SQ(x)$ is possible if, in time $O(T)$, the given data structure supports the following operations:
\begin{itemize}
    \item Sample: sample the $i$-th entry from $x$ with probability $x_{i}^2 / \|x\|$
    \item Query: output entries $x_i$ ($i \in [n]$) of $x$, and;
    \item Norm: determine the L2-norm $\|x\|$.
\end{itemize}
A data structure only supporting query operations over $x$ will be denoted as $Q(x)$. These assumptions serve as the classical analogue to quantum state preparation, and are often easier to satisfy.

This work also introduces three dequantized linear algebra protocols, which have been widely used by subsequent studies to dequantize various machine learning techniques.

\textbf{Protocol 1: Inner Product Estimation} (from Proposition 4.2 in \cite{tang2019quantum}):
For $x, y \in \mathbb{C}^n$ given $SQ(x)$ and $Q(y)$, the inner product $\braket{x|y}$ can be approximated with an additive error of $\|x\| \|y\|_\varepsilon$, and a probability of at least $1-\delta$, using $O(1/\varepsilon^2 log(1/\delta))$ queries and samples.


\textbf{Protocol 2: “Thin-Matrix” Multiplication} (from Proposition 4.3 in \cite{tang2019quantum}):
For $w \in \mathbb{C}^k$ and $V \in \mathbb{C}^{n \times k}$, given $SQ(V^\dagger)$\footnote{The $\dagger$ symbol denotes the complex conjugate transpose: $A^\dagger := (A^\intercal)^* = (A^*)^\intercal$} (sample and query access to columns of $V$) and $Q(w)$, sampling from the linear combination of $V$'s columns $Vw$ can be done in $O(k
^2C(V,w))$ query and time complexity, where
\begin{align}
    C(V,w) := \frac{\sum_{i=1}^{k} |w_i V^{(i)}|^2}{|Vw|^2}
\end{align}
The cancellation measure $C(V,w)$ quantifies the extent of cancellation in the matrix-vector product $Vw$, with a value of 1 indicating no cancellation in orthogonal columns and undefined values in cases of maximum cancellation, such as linearly dependent columns. 

\textbf{Protocol 3: Low-Rank Approximation} (from Theorem 4.4 in \cite{tang2019quantum}):
For matrix $A \in \mathbb{C}^{m \times n}$, a threshold $k$, and an error parameter $\varepsilon$, a low-rank approximation of A can be described with probability of at least $1-\delta$, in $O(poly(k, 1/\varepsilon))$ query and time complexity. Specifically in \cite{tang2019quantum}, the low-rank description is $SQ(S, \hat{U}, \hat{\Sigma})$ for matrices $S \in \mathbb{C}^{\ell \times n}$, $\hat{U} \in \mathbb{C}^{\ell \times k}$, and $\hat{\Sigma} \in \mathbb{C}^{k \times k}$ (with $\ell = poly(k, 1/\varepsilon))$. $S$ is a normalized sub-matrix of $A$, while $\hat{U}$, and $\hat{\Sigma}$ result from the singular value decomposition of $S$. This implicitly describes the low-rank approximation of $A$, denoted by $D$:
\begin{equation}
    D := A\hat{V}\hat{V}^\intercal = A(S^\dagger\hat{U}\hat{\Sigma}^{-1})(S^\dagger\hat{U}\hat{\Sigma}^{-1})^\dagger.
\end{equation}

\subsubsection{Recommendation Systems}

We now briefly describe the dequantized recommendation systems approach in \cite{tang2019quantum}. Like its quantum version, the algorithm finds a low rank approximation of the hidden preference matrix $A$, from which the preferences for a user $i$ are sampled. Specifically, given sample and query access $SQ(A)$ to $A$ in the data structure, the algorithm leverages L2-norm sampling to obtain a succinct description of matrix A: $SQ(S, \hat{U}, \hat{\Sigma})$ (Protocol 3). SQ access is shown to be available to these components; $\hat{V} := S^\intercal \hat{U} \hat{\Sigma}^{-1}$ is obtained implicitly (the approximate largest $k$ right singular vectors of $A$). The low-rank approximation of $A$ is thus $D$, the projection of $A$ onto the subspace of the right singular vectors $\hat{V}$, satisfying an acceptable error bound. Given, $SQ(S, \hat{U}, \hat{\Sigma})$, $SQ(D)$ and thus $SQ(D_i)$ is available. Protocol 1 is used to approximate $A_i\hat{V}$ using $k$ inner product estimations. Rejection sampling then allows for finding $D_i = A_i\hat{V}\hat{V}^\intercal$ in time independent of dimensions (Protocol 2). The algorithm's quality bounds on recommendations match those of the quantum version. Further, this regime produces an exponential improvement over the next best classical recommendations systems algorithm, with a running time of $\tilde{O}(\|A\|^{24}_F / \sigma^{24} \varepsilon^{12})$\footnote{The notation $\tilde{O}(\cdot)$ is used to suppress poly-logarithmic factors in variables.}.

Finally, the author provides a resulting guideline for performing any comparisons between QML and classical algorithms: 

\begin{quote}
\textit{When QML algorithms are compared to classical ML algorithms in the context of finding speedups, any state preparation assumptions in the QML model should be matched with L2-norm sampling assumptions in the
classical ML model} \cite{tang2019quantum} .
\end{quote}

Improvements over this work in the recommendation systems task come after several advancements in the dequantized algorithms field, particularly with the advent of the dequantized Quantum Singular Value Transform (QSVT) (Section \ref{sec:deq_qsvt}). Chia et al. \cite{chia2022sampling} used their dequantized QSVT framework to obtain a complexity bound of $\tilde{O}(\|A\|^6_F \|A\|^{10} / \sigma^{16} \varepsilon^{6})$, which introduces a dependence on $\|A\|$, but otherwise substantially improves upon \cite{tang2019quantum}. Chepurko et al. \cite{chepurko2022quantum} presents an algorithm that achieves a lower bound run-time complexity of $\Omega(k^3/\varepsilon^6)$ for generating a low-rank $k$ approximation of an input matrix $M$, guaranteeing that $||A - M||_F^2 \leq (1 + \varepsilon)||A - A_k||_F^2$. Their methodology employs `$\lambda$-importance' sampling sketches that over sample ridge leverage score sketching, which relies on certain assumptions regarding the size of the Frobenius norm of $A$, $||A||_F$, and the residual $||A - A_k||_F$. The authors note that this is directly comparable to \cite{tang2019quantum}, which requires a run-time that is polynomially large in $k$, $\kappa$, and $\varepsilon^{-1}$. While providing relative error bounds for low-rank approximations, their method has an additive error, with a bound more like $||A - A_k||_F + \varepsilon||A||_F$.

Bakshi and Tang \cite{bakshi2023improved} similarly employed their dequantized QSVT framework, and achieved a complexity of $\tilde{O}(\|A\|^4_F / \sigma^{9} \varepsilon^{2})$. The authors note that direct comparison with \cite{chepurko2022quantum} is difficult due to the additional assumptions on the size of $\|A_k\|_F$ and the residual $\|A - A_k\|_F$ required by ridge leverage score sketching. Introducing the bound $k \leq \|A\|^2_F / \sigma^2$ converts these error bounds to be more “QSVT-like”, which elucidates the run-time in \cite{chepurko2022quantum} to be $\tilde{O}(\|A\|^6_F / \sigma^{6} \varepsilon^{6})$, which is improved upon in terms of $\|A\|_F$ and $\varepsilon$, yet loses a factor of $\sigma^3$.

\subsubsection{Clustering and Dimensionality Reduction}

Following their work in \cite{tang2019quantum}, Tang further went on to dequantize both quantum supervised clustering and principle component analysis (PCA) \cite{tang2021quantum}. In that work, the dequantization model (SQ access input model with L2-norm sampling assumptions) is formalized, which directly comes from the prior work on the recommendation systems problem:

\textit{A quantum protocol's $\mathbb{S}: O(T)$-time state preparation of $\ket{\phi_1}, \dots, \ket{\phi_c} \rightarrow \ket{\psi}$ is “dequantized” if a classical algorithm of the form $\mathbb{C_S}: O(T)$-time $SQ(\phi_1, \dots, \phi_c) \rightarrow SQ^v (\psi)$ can be described with similar guarantees to $\mathbb{S}$ up to polynomial slowdown} \cite{tang2021quantum}.

Under this framework, only a quadratic speedup (due to amplitude amplification) is observed by the quantum nearest-centroid supervised clustering method proposed in \cite{lloyd2013quantum}, which aims to find the distance from a point $u \in \mathbb{C}^n$ to the centroid of a cluster of points given by $V \in \mathbb{C}^{mxn}$ (and let $\bar{V}$ be $V$ with unit normalized rows). Assuming SQ access to both $u$ and rows of $V$, the problem reduces to approximating $\|Mw\|$, where:
\begin{align}
    M := 
    \begin{bmatrix} 
        u/\|u\| \\[6pt]
        \frac{1}{\sqrt{n}} \bar{V}
    \end{bmatrix}
    \ \text{and} \
    w :=     
    \begin{bmatrix} 
        \|u\| - \frac{1}{\sqrt{n}} \bar{V}
    \end{bmatrix}
\end{align}
At this point, the quantum version will perform a swap test to construct $\ket{wM}$. This can be effectively dequantized by reformulating $\|Mw\|$ into $w^{\dagger}M^{\dagger}Mw$, which can be given as the inner product $\braket{a|b}$ of two tensors $a$, $b \in \mathbb{C}^{m \times n \times n}$ constructed from $M$ and $w$:

\begin{align}
    a:=\sum_{i=1}^d \sum_{j=1}^{n+1} \sum_{k=1}^{n+1} M_{j i} \| M_{k, *} \| \ket{i}\ket{j}\ket{k};
\end{align}

\begin{align}
    b:=\sum_{i=1}^d \sum_{j=1}^{n+1} \sum_{k=1}^{n+1} \frac{w_j^{\dagger} w_k M_{k i}}{\|M_{k, *}\|}\ket{i}\ket{j}\ket{k} .
\end{align}

Then, given $SQ(a)$ and using Protocol 1, the desired approximation is achieved in $O(T(\|w\|^2/\varepsilon^2)log(1/\delta))$ time.

The Quantum Principle Component Analysis (QPCA) method \cite{lloyd2014quantum} was dequantized via a similar methodology to that in \cite{tang2019quantum}, since the low rank approximation effectively produces a dimensionality reduction of the given matrix. In QPCA, the algorithm outputs estimates for both the top-$k$ singular values $\hat{\sigma}_{1}^2, \dots , \hat{\sigma}_{k}^2$ and singular vectors $\ket{\hat{v}_{1}^2}, \dots, \ket{\hat{v}_{k}^2}$, using density matrix exponentiation to find these principal components. The dequantized version uses Protocol 3 to find $SQ(S, \hat{U}, \hat{\Sigma})$ and produces the approximate large singular vectors $\hat{V} := S^{\dagger} \hat{U} \hat{\Sigma}^{-1}$. The $\sigma_i$'s are taken from $\Sigma$, and the $v_i$'s are given implicitly: $\hat{v}_i = S^{\dagger} \hat{U}_{*,i} / \hat{\sigma}_i$. 

In the QSVT dequantization method of \cite{chia2022sampling}, the bounds in \cite{tang2021quantum} for supervised clustering were reproduced, and improved for PCA from $\tilde{O}(\|X\|^{36}_F / \|X\|^{12} \lambda^{12}_k \eta^6 \varepsilon^{12})$ to $\tilde{O}(\|X\|^{6}_F / \|X\|^{2} \lambda^{2}_k \eta^6 \varepsilon^{6})$, a substantial improvement in almost all parameters.

\subsubsection{Matrix Inversion and Solving Linear Systems}

The task of matrix inversion is an essential subroutine in many machine learning optimization operations. Solving the linear system $Ax = b$ for some $A \in \mathbb{R}^{m \times n}$ typically requires solving the pseudoinverse of $A$: $\bar{x} = A^{+}b$ (finding an $\bar{x}$ minimizing $\|Ax - b\|$ when $A$ is not invertible). In classical settings, computing the pseudoinverse can be computationally difficult, especially for large or ill-conditioned matrices. Known methods of doing so, such as the Moore-Penrose pseudoinverse, may take, at worst, close to $O(n^3)$ time (where $m \approx n, m < n$) \cite{courrieu2008fast}; a severe bottleneck in large data applications \cite{geron2022hands}.

In the quantum setting, matrix inversion is performed using the famed HHL algorithm \cite{harrow2009quantum}. Dequantizing this in the sparse matrix inversion setting is generally difficult due to its BQP completeness \cite{harrow2009quantum}. However, many machine learning contexts operate in low rank settings, wherein which dequantization becomes possible, as demonstrated by the low-rank matrix inversion algorithms independently proposed by Gilyén et al. \cite{gilyen2018quantum} and Chia et al. \cite{chia2018quantum}. These algorithms exploit the low-rank structure of the input matrix to perform the inversion efficiently using classical techniques. 
In \cite{gilyen2018quantum} an exponential speed-up was shown to be possible for classical low-rank matrix inversion. Given $SQ(A)$ and $SQ(b)$, both in constant time, $SQ(A^+b)$ is approximated in time $O(polylog(mn))$, equivalent to the quantum version presented in \cite{harrow2009quantum}, thus showing no quantum advantage. Specifically, the approximation to $\bar{x}$ is given by:
\begin{align}
    \begin{split}
        \tilde{x} & \approx \sum_{\ell=1}^k \frac{1}{\tilde{\sigma}_{\ell}^4}
        \ket{R^{\dagger} w^{(\ell)}}
        \bra{R^{\dagger} w^{(\ell)}}A^{\dagger} b 
        \\ & \approx \sum_{\ell=1}^k \frac{1}{\tilde{\sigma}_{\ell}^2}
        \ket{\tilde{v}_A^{(\ell)}}
        \bra{\tilde{v}_A^{(\ell)}} A^{\dagger} b \approx
        (A^{\dagger} A)^{+}A^{\dagger} b=A^{+} b
    \end{split}
\end{align}

$SQ(A)$ allows for L2-norm sampling of row indices $r$ from $A \in \mathbb{C}^{m \times n}$; $s \in [r]$ indices are then sampled from uniformly, from which the column indices $c$ are L2-norm sampled to find $C$ (Protocol 3); $R \in \mathbb{C}^{r \times n}$ is implicitly defined from the $r$ rows. The Singular Value Decomposition (SVD) of $C$ is then computed to obtain its left singular vectors $u^1, \dots , u^k$ and singular values $\widetilde\sigma_1, \dots , \widetilde\sigma_k$, which are shown to be good approximations for the right singular vectors of $R$, and implicitly $A$. Specifically, this approximation is good with probability at least $1 - \eta$ given $r \geq \frac{4 ln (2n / \eta)}{\varepsilon^2}$. The left singular vectors $\widetilde{v}_{A}^{(\ell)} := \left(\sum_{s=1}^{r} R^{\dagger}_{i_s} u^{(\ell)}_s / \widetilde\sigma_\ell\right)$ of $A$ are then approximated by the projection $(\bra{\tilde{v}_{A}^{(\ell)}}A^{\dagger}/\widetilde{\sigma}_{\ell})$. Then an estimate of $\bra{\tilde{v}_{A}^{(\ell)}}A^{\dagger}\ket{b}$ is obtained via Protocol 1 to additive error, since $\bra{\tilde{v}_{A}^{(\ell)}}A^{\dagger}\ket{b}$ = $\text{Tr}\left[ \ket{A^{\dagger}b}\bra{\tilde{v}_{A}^{(\ell)}} \right]$ is an inner product. Protocol 2 produces samples from the linear combinations of these to achieve the approximate solution $\bar{x}$.



Chia et al. \cite{chia2018quantum} present a similar method. The objective is to approximate $A^\dagger A$ from the singular values and singular vectors of a small sub-matrix of $A$, from which the solution $A^{-1}b$ can then be sampled from in sub-linear time, given $SQ(A)$. The left singular values and vectors are produced from $A$ using a similar sub-sampling technique in \cite{gilyen2018quantum} to arrive at $A \in \mathbb{C}^{m \times n} \rightarrow R \in \mathbb{C}^{p \times n} \rightarrow C \in \mathbb{C}^{p \times p} \rightarrow W \in \mathbb{C}^{k \times k}$ where $W$'s $u^1, \dots , u^k$ left singular vectors and $\hat{\sigma}_1, \dots , \hat{\sigma}_k$, singular values are good approximations of $A$'s. $V \in \mathbb{C}^{n \times k}$ is then formed from column vectors $\frac{S^\dagger}{\hat{\sigma}_i}\hat{u}_i$, with $D \in \mathbb{C}^{k \times k}$ being a diagonal matrix containing the $\hat{\sigma}_i$'s. This allows for the description of $\hat{A}^{\sim 2} = VD^{-2}V^\dagger$, the approximation of the square matrix $A^\dagger A$ in the normal equation $(A^{\dagger} A)^{+}A^{\dagger} b$. The problem is now to find $\bar{x} = VD^{-2}V^\dagger A^+ b$. An extension of Protocols 1 and 2 is proposed that shows that approximating $x^\dagger Ay$ up to additive error is possible given $SQ(A)$, $SQ(x)$ and $SQ(y)$. This is used to find $w \in \mathbb{C}^{k} = V^\dagger A^+ b$. The authors then show that both querying an entry in $V_{i,*}D^{-2}w$ (via Protocol 1) and sampling the solution $VD^{-2}w$ (via Protocol 2) can be done in sub-linear time.

The works by \cite{gilyen2018quantum} and \cite{chia2018quantum} achieve running times of $\tilde{O}(k^6 \|A\|^6_F \|A\|^{16} / \sigma^{22}\varepsilon^6)$ (where $k \leq \frac{\|A\|^2_F}{\sigma^2}$ is rank($A$)) and $\tilde{O}(\|A\|^6_F \|A\|^{22} / \sigma^{28}\varepsilon^6)$ respectively. The dependence on $k$ in the former makes direct comparison between these complexities difficult, though the algorithm in \cite{gilyen2018quantum} operates in the more restricted setting where $A$ is strictly rank $k$. Chia et al. go on to employ their dequantized QSVT \cite{chia2022sampling} (see Section \ref{sec:deq_qsvt}) to re-derive an algorithm for the task that runs in the same time as \cite{chia2018quantum}, in a more general setting where $\sigma$ can be chosen as an arbitrary threshold, and not necessarily the minimum singular value of $A$.

Several subsequent works have produced quantum-inspired classical low-rank linear regression algorithms that improve on these complexity bounds. Gilyén et al. \cite{gilyen2022improved}, in the same setting, provide two algorithms; one that outputs a measurement of $|x_i|$ in the computational basis and another that outputs an entry of $x$ in $\tilde{O}\left(\|A\|^6_F \|A\|^6 / \sigma^{12}\varepsilon^4 \right)$ and $\tilde{O}\left(\|A\|^6_F \|A\|^2 / \sigma^8 \varepsilon^4 \right)$ time, respectively, a large improvement on all prior methods. The problem considered is extended to the ridge regression setting: finding a solution $x$ given $f(x) := \frac{1}{2}\left(\|Ax - b\|^2 + \lambda\|x\|^2\right)$. 
The key contribution is the construction of a sparse description of the solution vector $b$, after which a stochastic gradient descent (SGD) based optimization process exploits sample-query accesses $SQ(A)$ and $Q(b)$ to find $x$. This avoids explicitly computing the matrix pseudo-inverse, resulting in large improvements in running time. The authors note that this method is viable only when $b$ is sparse, but show that via matrix sketching techniques, this requirement can be bounded up to a certain degree. This work highlights the advantages of using iterative approaches in conjunction with quantum-inspired linear algebra techniques.

Chepurko et al. \cite{chepurko2022quantum} proposed an algorithm for the ridge regression problem using alternative techniques from randomized numerical linear algebra. A key distinction is the use of Projection-Cost Preserving (PCP) sketches, with analysis via ridge leverage score sampling techniques, as opposed to the L2-norm sampling common in prior quantum-inspired algorithms. The authors argue that sampling according to the squared row norms of a matrix $A$ is akin to sampling from a distribution close to the leverage score distribution. Consequently, by oversampling by a factor of the square of the condition number $(\kappa^2 (A))$, the same guarantees can be achieved as with leverage score sampling. PCP sketches produces the low-rank approximation of $A$, which is then processed through the SVD, and a QR factorization to produce an approximation of $A$. This small, sub-matrix decomposition solves a linear system of equations using a conjugate gradient method to find the solution, as opposed to inverting the matrix directly. A running time of $\tilde{O}(\|A\|^4_F \log(c) / \sigma^8 \varepsilon^4)$ is achieved, where $\sigma$ is the minimum singular value of $A$, $c$ is the number of columns in $A$, and assuming some sizable regularization parameter $\lambda = \Theta(\|A\|^2_2)$\footnote{The symbol \( \Theta \) denotes that the behavior of \( \lambda \) is tightly bound to the squared 2-norm of matrix \( A \), neither growing much faster nor much slower than that term.}; a roughly $\|A\|^2_F / \|A\|^2$ improvement on \cite{gilyen2022improved}.

Shao and Montanaro \cite{shao2022faster} propose two algorithms for solving linear systems: One based on the randomized Kaczmarz method, and one based on the randomized coordinate descent method. Both are iterative methods. The authors explore their algorithms under various settings of $A$, such as when it is dense, sparse, or semi-positive definite (SPD), and note that their latter algorithm only operates in the SPD setting. The former algorithm focuses on the dual form of the Kaczmarz method, specifically iterating on $y$, obtained by introducing $y$ such that $x = A^\intercal y$ for consistent linear systems. The goal is to find a sparse approximate solution for $y$, which, when plugged back into $x = A^\intercal y$ provides an approximate solution for the original linear system, similar to \cite{gilyen2022improved}. This sparsity, guaranteed by the iterative method, simplifies querying an entry of $x$, and an entry of $x$ can be sampled using rejection sampling. The resulting algorithm has a complexity bound of $\tilde{O}(\|A\|^6_F \|A\|^2 / \sigma^8\varepsilon^2)$, which improves in $\varepsilon$ over \cite{chepurko2022quantum}, but suffers in $\|A\|^6_F$.
In \cite{gilyen2022improved}, the authors recognize the roughly $\|A\|^2_F / (\sigma^4 \varepsilon^2)$ improvement on their work, when $\lambda = 0$ and $Ax^* = b$ exactly (without any regression parameter). The authors note the similarity between this method and the ones introduced in \cite{shao2022faster}; the randomized Kaczmarz update is a variant of SGD, and by replacing their stochastic gradient with the Kaczmarz update (and with accommodating minor adjustments to their sketch), a similar running time to \cite{shao2022faster} is achieved for the no regression setting.

Finally, Bakshi and Tang \cite{bakshi2023improved} achieve a running time of $\tilde{O}(\|A\|^4_F / \sigma^{11} \varepsilon^2)$ with their method and note that this is comparable with, or improves upon several, prior works over the matrix inversion and linear systems tasks. When contrasted with \cite{chepurko2022quantum}, their work exhibits better $\varepsilon$ dependence and equivalent $\|A\|_F$ dependence. Notably a superior $\sigma$ dependence in low-error settings is achieved, although in situations where a higher error is permissible, the approach in \cite{chepurko2022quantum} may be more efficient. Relative to \cite{shao2022faster}, the authors match the dependence in $\varepsilon$, excels in $\sigma$-dependence, but falls short in Frobenius norm dependence. Additionally, the authors note their approach is more general than in \cite{shao2022faster}, which requires $b$ to be in the column space of $A$.

\subsubsection{Support Vector Machine}

The seminal quantum support vector machine (SVM) algorithm presented in \cite{rebentrost2014quantum} finds a solution to a set of linear equations --- the Least Squares SVM (LS-SVM); a variant of the SVM optimization problem. LS-SVM attempts to label points in $\mathbb{R}^m$ as +1 or -1. Given input data points $x_j$ for $j = 1, ..., m \in \mathbb{R}^n$ and their corresponding labels $y$ in ${\pm 1}^m$, the goal is to find a hyperplane, specified by $w \in \mathbb{R}^n$ and $b \in \mathbb{R}$, that separates these points. Since it may not be possible to perfectly separate all points, a slack vector $e \in \mathbb{R}^m$ is introduced, where $e(j) \geq 0$ for all $j \in [m]$. The objective is to minimize the squared norm of the residuals, which is a combination of the squared norm of $w$ and $e$, given by:

\begin{align}
    & \min_{\textbf{w}, b} \frac{\|w\|^2}{2} + \frac{\gamma}{2} \|e\|^2 \\
    & \text{s.t.} \quad y_i(w^\intercal x_i + b) = 1 - e_i, \forall i \in [m].
\end{align}

In the dual problem, LS-SVM tries to find a classification for the data points based on their projection on the optimal hyperplane. This is achieved by solving for the $\alpha$ values in the equation:
\begin{align}
    (X^\intercal X + \gamma^{\negexp{1}} I)\alpha = y,
\end{align}
where $\alpha = u_k y_k$; this term quantifies the weight of each data point in defining the classification hyperplane. The equation for the hyperplane thus becomes $x^\intercal X\alpha = 0$. For new data points $x$, the LS-SVM performs the classification by evaluating $\text{sgn}(x^\intercal X\alpha)$.

In the quantum approach, labeled data vectors $x_j$ for $j = 1, ..., m$ are transformed into quantum vectors $\ket{x_j} = \frac{1}{\|x_j\|} \sum_k (x_j)_k \ket{k}$ via QRAM. The kernel matrix is then assembled by leveraging quantum inner product estimations. The solution is obtained by solving a system of linear equations related to the quadratic programming problem of the SVM, facilitated by the HHL algorithm. A dequantized LS-SVM algorithm was presented in \cite{ding2021quantum}. Here, the data vectors are stored in classical QRAM \cite{kerenidis2016quantum}, allowing for $SQ(x)$. Although the L2 norm sub-sampling methods from \cite{gilyen2018quantum, chia2018quantum} could be used to inverse $(X^\intercal X + \gamma^{\negexp{1}} I)$, they would not be able to produce the inverse in the reported logarithmic time, since directly computing $X^\intercal X$ is done in polynomial time. Instead, the authors devise an indirect sampling technique to perform the inversion, given only $SQ(X)$. This process involves two primary stages: column and row sub-sampling from the matrix $X$, forming matrices $X'$ and $X''$ and their squares $A' = X'^\intercal X'$ and $A'' = X''^\intercal X''$ respectively. Here, it is the spectral decomposition of $A'' = V''\Sigma^2V''^\intercal$ that yields the approximate eigenvalues and eigenvectors of $X$. The elements of the vector $\tilde{\lambda}_l = \tilde{V}^\intercal_l y$ are estimated, using the approximated eigenvectors and the data labels. Then, a vector $u$ is derived as a weighted sum of the columns from the spectral decomposition. The algorithm constructs $\tilde{\alpha}$, an approximation of the solution to the LS-SVM problem obtained by finding query access of $\tilde{\alpha} = \tilde{R}^\intercal u$, where $\tilde{R}$ represents a sampled part of matrix $R =  X'^\intercal X$. The final classification of any new data point $x$ is determined by calculating its projection onto the classification hyperplane, given by $x^\intercal X\tilde{\alpha}$, with the sign of this projection determining the class of $x$.

\subsubsection{Semidefinite Programming}

Semidefinite programming (SDP) is a subfield of convex optimization concerned with the optimization of a linear objective function over the intersection of the cone of positive semidefinite matrices with an affine space. While traditional, classical SDP solvers assume entry-wise access to matrices, quantum SDP solvers make use of oracle access to specially constructed data structures, allowing for sub-linear access of matrix elements in superposition. Such quantum SDP algorithms have been explored in the sparse setting to achieve exponential speeds. With respect to this paradigm, \cite{chia2019quantum} propose a dequantized SDP solver on the back of Tang's methodology \cite{tang2019quantum}, since such access can be analogously described classically via the sample and query framework.





In \cite{chia2019quantum} the normalized SDP feasibility problem variant of SDP is explored, which involves finding, for a given $\varepsilon > 0$, a positive semidefinite matrix $X$ with trace 1 that satisfies $\text{Tr}[A_iX] \leq a_i + \varepsilon$ for a set of given Hermitian matrices $A_i$ and real numbers $a_i$, or determining that no such matrix exists. By leveraging binary search, the $\varepsilon$-approximation of the SDP can be simplified into a sequence of feasibility tests, as per the above definition. The Matrix Multiplicative Weight (MMW) method is used as the solver, which is a game-theoretic, iterative process that seeks an approximate feasible solution. In each iteration, the first player attempts to find a feasible solution $X \in S_\varepsilon$, which is updated based on any violation $j \in [m]$ by any proposed $X$ of the constraints found by the second player, i.e., $\text{Tr}[A_jX] > a_j + \varepsilon$. Each round $t$ has the update $X_{t+1} \leftarrow \exp[-(A_{ji} + \ldots + A_{jt})]$ for the next round. The solution is arrived at in the equilibrium of this process.

The approach here is to sample from an approximate description of the matrix exponentiations $\exp[-(A_{ji} + \ldots + A_{jt})]$, using the approximate spectral decomposition $A := \sum_{r=1}^t A_{j\tau}$ as $A \approx \hat{V}\hat{D}\hat{V}^\dagger$ to achieve $e^\negexp{A} \approx \hat{V}e^\negexp{D}\hat{V}^\dagger$, where $\hat{V} \in \mathbb{C}^{n \times r}$ and $\hat{D} \in \mathbb{R}^{n \times n}$ are diagonal. This work devises two methods to address a few encountered challenges. First, the dynamic nature of the matrix $A$ throughout the MMW method makes the assumption $SQ(A)$ infeasible. This problem is circumvented by devising a weighted sampling procedure that provides a succinct description of a low-rank approximation of $A$ by sampling each individual $A_{j\tau}$: $A := \sum_{r=1}^t A_{j\tau}$. Secondly, standard sampling procedures yield an approximation $V D^2 V^{\dagger} \approx A^{\dagger}A$ instead of a spectral decomposition $\hat{V}\hat{D}\hat{V}^{\dagger} \approx A$, even if $A$ is Hermitian. This discrepancy is problematic for matrix exponentiation, as the singular values disregard the signs of the eigenvalues, leading to significant errors when approximating $e^{-A}$ by naively exponentiating the SVD. The resolution is a novel approximation procedure called symmetric approximation, which can calculate and diagonalize the small matrix $(\hat{V}^{\dagger}A\hat{V})$, yielding approximate eigenvalues of $A$ and a desired description of its spectral decomposition in logarithmic time in dimension. Evaluating the solution via the constraints can be done with the dequantized protocols, e.g. $\text{Tr}[A_jX]$ via Protocol 1.


\subsubsection{Quantum Singular Value Transform}\label{sec:deq_qsvt}
Recent research by Gilyén et al. \cite{gilyen2019quantum} has proposed that a broad array of quantum algorithms may be manifestations of a singular underlying principle, encapsulated by the Quantum Singular Value Transformation (QSVT). The QSVT can be understood as a process that, given a matrix $A \in \mathbb{C}^{m \times n}$ and a function $f: [0, \infty) \rightarrow \mathbb{R}$, implements a unitary $U$ that represents the polynomial transformation of $A$ to the matrix $f(A)$, defined via its singular value decomposition. The QSVT is uniquely characterized by its parametric nature, defined by phase angles ${\phi(k)}$ which can be
efficiently and stably computed in a classical manner given
a desired polynomial transformation \cite{martyn2021grand}. The flexibility of this parameterization imbues the QSVT with remarkable power and adaptability; many quantum algorithms can be phrased
under the QSVT framework, so long as there can be found
a specific parameterization for the task. As such, the QSVT
has been touted as a “unifying framework” for quantum
algorithms \cite{martyn2021grand}; its power lies in its capacity to encapsulate a broad range of quantum algorithms within a coherent theoretical structure.

The QSVT extends upon Quantum Signal Processing and Qubitization \cite{low2019hamiltonian}, applicable to an entire vector space. It can apply polynomial transformations to all the eigenvalues of a given Hamiltonian $H$ that has been block-encoded into a larger unitary matrix $U$ -- the Quantum Eigenvalue Transform (QET). This can be generalized for non-square matrices to its singular values -- the QSVT. Given the singular value decomposition of a matrix $A$:
\begin{align}
    A = U_\Sigma \Sigma V^\dagger_\Sigma = \sum_{k=1}^r \sigma_k\ket{u_k}\bra{v_k},
\end{align}
where $U_\Sigma , V_\Sigma$ are unitaries and $\Sigma$ is a diagonal matrix with non-negative, real singular values $\sigma_1, \ldots, \sigma_k$ along the diagonal, up to the $k$-th ($r =$ rank($A$)) singular value. $U_\Sigma$ and $V_\Sigma$ span orthonormal bases, denoted by $\{|u_k|\}$ (the left singular vector space) and $\{|v_k|\}$ (the right singular vector space) respectively. These spaces can be defined by the projectors $\tilde{\Pi} := \sum_k \ket{u_k}\bra{u_k}$ and $\Pi := \sum_k \ket{v_k}\bra{v_k}$. The block-encoding of $A$ within a unitary $U$ can thus be given by:

\newcommand\scalemath[2]{\scalebox{#1}{\mbox{\ensuremath{\displaystyle #2}}}}

\NiceMatrixOptions{cell-space-limits = 1pt}
\begin{align}
    U = \
    \begin{bNiceMatrix}[first-row,first-col,columns-width = 0.5cm]
        & \scalemath{0.75}{\Pi} &  \\
        \scalemath{0.75}{\tilde{\Pi}} & A & \cdot \\
         & \cdot & \cdot
    \end{bNiceMatrix}.
\end{align}
$\tilde{\Pi}$ and $\Pi$ locate $A$ within $U$, such that they form the projected unitary encoding $A := \tilde{\Pi} U \Pi$. The QSVT is then realized with the projector-controlled phase-shift operations of $\tilde{\Pi}$ and $\Pi$ to approximate the degree-$d$ polynomial transformations:
\NiceMatrixOptions{cell-space-limits = 1pt}
\begin{align}
    U_{\overrightarrow{\phi}} = \
    \begin{bNiceMatrix}[first-row,first-col,columns-width = 0.5cm]
        & \scalemath{0.75}{\Pi} &  \\
        \scalemath{0.75}{\tilde{\Pi}} & f^{\text{(SV)}}(A) & \cdot \\
         & \cdot & \cdot
    \end{bNiceMatrix},
\end{align}
where $f^{\text{(SV)}}(A)$ is defined for an odd polynomial as:
\begin{align}
    f^{\text{(SV)}}(A) := \sum_{k=1}^r f^{\text{(SV)}}(\sigma_k)\ket{u_k}\bra{v_k},
\end{align}
which applies the polynomial transformation to the singular values of $A$. A similar result is obtained for even polynomial transformations, where only the right singular vector projections are involved. 
Of importance is the compositional properties of block-encoded matrices. Many quantum algorithms and QML applications can be reformulated via this framework, given a satisfactory approximation can be achieved by carefully considered polynomial transformations.

\textbf{Dequantizing the QSVT:} A natural question now is whether the QSVT can be successfully dequantized. Several independent efforts have been made on this front. Chia et al. \cite{chia2022sampling} note that algorithms framed under QSVT fall roughly into two categories based on their assumptions on input data $A$: those with sparsity assumptions and those with low stable rank and under QRAM assumptions (which inform how the block-encodings in \cite{gilyen2019quantum} can be obtained). These settings allow for efficient block-encodings, outside of which they are generally not efficiently computable. It may not be possible to dequantize algorithms under sparsity assumptions, since this would imply dequantizing the HHL algorithm, which is recoverable from the QSVT framework; HHL is known to be BQP-complete for sparse matrices, even for constant precision \cite{harrow2009quantum}. For block-encoding arithmetic operations over algorithms using QRAM-based data structures, this category shares efficient operations with classical algorithms that have \textit{oversampling and query} access available. From this insight, the authors go on to implement a dequantized QSVT framework applicable to the prior tasks examined under the dequantized lens, showing improvements in query and time complexity over several established methods. These results suggest that dequantization techniques can be generalized to most, if not all QRAM-based QSVT models, and give strong evidence that these models admit no exponential speedup in the bounded Frobenius norm regime. In particular, the framework extends upon the notion of SQ access for vectors $SQ(v)$ and matrices $SQ(A)$, and formalizes $\phi$-oversampling and query access to vectors and matrices $SQ_\phi (v)$ and $SQ_\phi (A)$ respectively. $\phi$-oversampling and query access to a vector $v$ is available if we have $Q(v)$ and sample and query access $SQ(\tilde{v})$ to a vector $\tilde{v}$, which serves as an “element-wise upper bound” of $v$ and complies with the following properties: $\|\tilde{v}\|^2 = \phi\|v\|^2$ and $|\tilde{v}_i|\geq|v_i|$ for every index $i$. The definition can be extended similarly for a matrix. The factor $\phi$ can be interpreted as a kind of computational surplus incurred in the execution of algorithms: Via rejection sampling, $SQ(\tilde{v})$ is capable of performing approximations of all the queries of $SQ(v)$, albeit with an additional cost denoted by the factor $\phi$. This cost roughly corresponds to overheads in post-selection in the quantum setting. The oversampling and query input model has closure properties very similar to that of block-encodings, that allow for the chained composition of complex arithmetic operations, and can be achieved whenever quantum states and generic block-encodings can be prepared efficiently.

Specifically, these closure properties allow for accessing the approximation of $f(A^\dagger A)$ for a smooth, bounded Lipschitz-continuous function $f$ in time independent of dimension, given a generic matrix $A$. $f(A^\dagger A)$ can be shown to, via various matrix sketching techniques and the aforementioned closure properties, approximate an RUR decomposition $R^\dagger \bar{f}(CC^\dagger) R$, where  $\bar{f}(x := f(x)/x)$, and $R$ and $C$ contain the normalized columns of $A$ and $R$ respectively. This expresses a desired matrix as a linear combination of $r$ outer products of rows of the input matrix, for which oversampling and query access is available, since having $SQ(A)$ implies $SQ_\phi (R^\dagger U R)$, where $U := \bar{f}(CC^\dagger)$. We can then achieve an approximation for $f^\text{(SV)}(A)$ up to some error, from which we can sample from some solution vector $v := \ket{f^\text{(SV)} (A)b}$ such that $\|f(A)b - v\| \leq \varepsilon$ in $\tilde{O}(d^{22} \|A\|^6_F / \varepsilon^6)$ time; the dequantized generic singular value transform for degree-$d$ polynomials. Significantly, the authors argue for the generality of their framework, which allows for approximately low rank matrices, rather than strictly low rank matrices, as input, which has been stipulated in prior works. This framework is then applied to re-derive several existing dequantized algorithms, showing improvements in complexity bounds in most cases, and some relaxation of input constraints in many, as seen in Section \ref{sec:deq_complexitycomparisons}. Further, the framework was applied to two tasks that had not yet been explored under the dequantization framework: Hamiltonian Simulation and Discriminant Analysis.

In a parallel work, Jethwani et al. \cite{jethwani2019quantum} also proposed their version of the dequantized QSVT. The authors similarly define the function $f$ as smooth, Lipschitz-continuous, such that: $f^\text{(SV)}(A^\dagger) = h(A^\dagger A)A^\dagger$ for $h(x) = f(\sqrt{x})/\sqrt{x}$. The authors provide a sketch of $A$ down to a smaller sub-matrix $W$ from successive samples of rows and columns of $A$. Then a similar result to \cite{chia2022sampling} is obtained by computing the SVD of $W$ to find $U := W^+ f(W^\dagger) W^+ (W^\dagger)^+$. The solution is then sampled from $R^\dagger U R A^\dagger b \approx f(A^\dagger)b$. The complexity achieved is $\tilde{O}(\|A\|_F^6\kappa^{20}(d^2 + \kappa)/\varepsilon^6)$, where $\kappa$ is the condition number of $A$. The authors require that $A$ be a well-conditioned matrix, such that all $\sigma_k$ of $A$ lie within $\left[1/\kappa, 1\right]$. The conditional number dependence makes direct comparison with the results from \cite{chia2022sampling} difficult, however, \cite{chia2022sampling} surmises that for the singular value transform in \cite{jethwani2019quantum}, the typical cases of $f$ require $A$ to be strictly low-rank.

Gharibian and Le Gall \cite{gharibian2022dequantizing} explored dequantizing QSVT for sparse matrices, noted to be a difficult task. This is the primary setting relevant to quantum chemistry and quantum PCP applications. They derive a dequantized QSVT algorithm for $s$-sparse Hamiltonians, where $s$ denotes the number of non-zero entries per row or column. This algorithm is then considered for the task of Guided Local Hamiltonian estimation (GLH): estimating the ground state energy of a local Hamiltonian when given a state sufficiently close to the ground state. Two key results follow: (1) showing that QSVT in sparse settings provided with SQ access can be efficiently dequantized and computed classically given constant-degree polynomials and with constant precision, and; (2) this problem becomes BQP-complete for input representations in a “semi-classical state”, which are unit-norm vectors represented as a uniform superposition of basis states within a polynomial size subset, with polynomial precision. This is distinguished from other efforts in dequantizing the QSVT, which only consider QSVT circuits in the QRAM setting. This work highlights the idea that dequantizing quantum algorithms is not a uniquely ML-driven focus; other fields may benefit from efforts in dequantizing algorithms for their respective applications.


Finally, the most recent entry into this line of work is presented by Bakshi and Tang \cite{bakshi2023improved}, who focused on improving the complexity bounds of the dequantized QSVT algorithms presented earlier. The authors suggested that a significant cost of performing these algorithms is due to the computation of the singular value decomposition of matrix $A$, which directly incurs the $\|A\|^6_F / \varepsilon^6$ cost in \cite{chia2022sampling}. A key insight is to additionally use iterative algorithms on top of the sketches of $A$ and $b$ to approximate matrix polynomials. Bakshi and Tang's work \cite{bakshi2023improved} combines sketches of matrices $A$ and $b$ with Clenshaw recurrence, an iterative algorithm, to approximate matrix polynomials more efficiently than previous approaches. They introduce the Bilinear Entry Sampling Transform (BEST) for matrix sparsification, which optimizes $\varepsilon$-dependence without requiring full spectral norm bounds. This technique streamlines the error analysis and achieves a notable reduction in complexity bounds for dequantized QSVT. Consequently, their method presents substantial improvements in terms of time and resource consumption; computing $v$ only requires $\tilde{O}(d^{11} \|A\|^4_F / {\varepsilon^2})$ time, and without the condition number dependence in \cite{jethwani2019quantum}. This complexity reduction brings the time required within the scope suggested as indicative of quantum advantage \cite{babbush2021focus}, thus providing robust evidence countering the notion of quantum supremacy. The framework was then used to find improvements in complexity bounds for the tasks of recommendations systems, linear regression, and Hamiltonian simulation.

\subsubsection{Discriminant Analysis}

This subsection leads on from the work of \cite{chia2022sampling}. The dequantized QSVT framework is used to dequantize the quantum Fisher's Linear Discriminant Analysis (LDA) algorithm, presented by Cong and Duan \cite{cong2016quantum}. The LDA problem aims to project classified data onto a subspace that maximizes between-class variance while minimizing within-class variance. Given $M$ input data points $\{x_i \in \mathbb{R}^N : 1 \leq i \leq M\}$ belonging to $k$ classes, we define between-class scatter matrix $S_B$ and within-class scatter matrix $S_W$. The original goal is to solve the generalized eigenvalue problem $S_Bv_i = \lambda_iS_W v_i$, but this may not be feasible in cases where $S_W$ is not full-rank. Consequently, Cong and Duan consider a relaxation where small eigenvalues of $S_W$ and $S_B$ are ignored, leading to an approximation using inexact eigenvalues, which can be applied to the quantum context.

Dequantizing this involves finding an approximate isometry $U$ and a diagonal matrix $D$, such that $S^{\frac{1}{2}}B S^{-1}_W S^{\frac{1}{2}}B U \approx U D$, which finds the approximate eigenvalues and eigenvectors of $S^+_W S_B$. Given $SQ(B,W) \in \mathbb{C}^{m \times n}$, with $S_W \approx W^\dagger W$ and $S_B \approx B^\dagger B$, the Even Singular Value Transform in \cite{chia2022sampling} approximates $\sqrt{W^{\dagger}W} \approx R^{\dagger}_W U_W R_W$ and $B^{-\dagger}B \approx R^{\dagger}_B U_B R_B$ through $RUR$ decompositions. Matrix sketching techniques then yield an approximate $RUR$ decomposition of the target matrix, $R^{\dagger}_W U R_W$, where $U = U_W R_W R^{\dagger}_B U_B R_B R^{\dagger}_W U_W$, from which the approximate eigenvalues and eigenvectors can be extracted.

Of course, here's the equation:

\subsubsection{Hamiltonian Simulation}

The Hamiltonian Simulation problem, rooted in the original motivation for quantum computers proposed by Feynman \cite{feynman2018simulating}, aims to simulate the dynamics of quantum systems. Given a Hamiltonian $H$, a quantum state $|\psi\rangle$, a desired error $\varepsilon > 0$, and time $t > 0$, the objective is to prepare a quantum state $|\psi_t\rangle$ such that $|||\psi_t\rangle - e^{iHt}|\psi\rangle|| \leq \varepsilon$. With wide applications in quantum physics and chemistry, the rich literature on quantum algorithms for Hamiltonian simulation includes optimal algorithms for simulating sparse Hamiltonians. In \cite{chia2022sampling} the dequantized QSVT framework is applied for Hamiltonian simulation that operate in different regimes, both for low-rank $H$ and for arbitrary $H$. Authors consider a Hermitian matrix $H \in \mathbb{C}^{n \times n}$, a unit vector $b \in \mathbb{C}^n$, and error parameters $\varepsilon, \delta > 0$. Given $SQ(H)$ and $SQ(b)$, the task is to output $SQ^\phi(\hat{b})$ with probability $\geq 1 - \delta$ for some $\hat{b} \in \mathbb{C}^n$ satisfying $||\hat{b} - e^{iH}b|| \leq \varepsilon$. Formulating the problem as a decomposition of a generic function $f(x)$ into sine and cosine functions results in the expression $e^{iH} b = f_{\text{cos}}(H^\dagger H)b + f_{\text{sinc}}(H^\dagger H)Hb$, where $\text{sinc}(x)=\sin(x) / x$, to which an RUR decomposition can be achieved. The authors obtain a running time of $\tilde{O}\left(\|H\|^6_F \|H\|^{16} / \varepsilon^6 \right)$ in the low-rank regime. The authors note that their algorithm for the Hamiltonian Simulation task is comparatively slower when compared with similar classical techniques in randomized linear algebra that consider sparsity in $H$ and $b$ \cite{rudi2020approximating}, and posit their framework exposes this trade-off between sparsity and speed.

In \cite{bakshi2023improved}, the authors improve upon the complexity in \cite{chia2022sampling} using their classical algorithm employing Clenshaw recurrence in the same setting. The authors here obtain a description that is 
$O(\varepsilon)$-close to $e^{iHt}$. Sampling from this can be done in
$\tilde{O}(\|H\|^4_F \|H\|^{9} / \varepsilon^2)$ time, an improvement over \cite{chia2022sampling} in all parameters.


\subsubsection{Complexity Comparisons}\label{sec:deq_complexitycomparisons}

A critical motivator in the dequantized algorithms space is in analyzing the nature of “quantum advantage” in QML settings, when presented with their classical counterparts. We adopt Chia et al.'s \cite{chia2022sampling} Figure 1 (Table \ref{tab:complexities} in this review), which presents the time complexities of the dequantized algorithms discussed, and the quantum algorithms they are based on. We extend this table with the complexities of subsequent works in the field. All complexities are given as poly-logarithmic in the input size. The dequantized algorithms are presented loosely in order of decreasing time complexity; the current gap between the QML and classical algorithms can be observed. Significant progress has been made with successive advancements, bringing us closer towards the QML benchmark, particularly seen in the matrix inversion and QSVT tasks. Targeted research efforts in these areas are justified due to the central role of matrix inversion as a fundamental subroutine in various linear algebra settings, and the generalization of numerous tasks under the QSVT framework. Further, the relaxation of parameters and data requirements can be observed in some successive works, showing an increase in generality of algorithms over time.


\subsubsection{Critical Views on the SQ Access Model}

Much of the work in dequantizing QML algorithms relies on the QRAM-like classical data structure introduced by Kerenidis and Prakash \cite{kerenidis2016quantum}. While this model has been successfully applied in the literature, there may exist alternative methods of representing the input data take can inform classical algorithms. In the quantum domain, input models that store data as entries in density matrices and simulating them as Hamiltonians is common, especially in QML applications \cite{lloyd2014quantum, rebentrost2014quantum}. Sparse representations on quantum circuits are another a form that has already been briefly discussed. 

Zhao et al. \cite{zhao2021smooth} address inefficiencies in the standard input access model for quantum machine learning due to the need for pre-computation and data storage. They introduce a flexible model enabling entry-wise data access, particularly useful when applying varying functions or requiring matrix row entries. The authors show that quantum amplitude encoding and classical L2-sampling can be conducted cost-effectively, even with moderately noisy input data, suggesting that quantum state preparation can be efficient despite initial state preparation challenges. 

Cotler et al. \cite{cotler2021revisiting} discuss the appropriateness of SQ access as a classical analog of quantum state inputs. SQ access, in its current form, allows classical algorithms to manipulate data that is exponentially difficult to extract from quantum states, thus artificially ascribing excessive power to dequantized algorithms when compared to their quantum counterparts. The authors suggest that the definition of SQ access needs to be revised. One plausible approach is to limit classical algorithms to accessing data obtained through measurements of quantum states inputs. This modification would preserve significant computational power for existing dequantized algorithms, while describing an oracle that is at most as powerful as inputs given to the quantum algorithms. Subsequent analysis notes that Quantum PCA under measurement data access retains its exponential quantum advantage \cite{huang2022quantum}. 

Alternative input models may thus need to be considered, and could provide a more accurate comparison between classical and quantum algorithms and better reflect the true potential of dequantized computation.

\setlength\tabcolsep{3pt} 
\begin{table*}[!ht]
    \centering
    \normalsize
    \caption{Comparison of time complexities for dequantized algorithms across specific tasks. In each row, the references given in the leftmost column correspond to each successive item from left to right.}
    \resizebox{1\textwidth}{!}{%
    \begin{tabular}{r|c|cccccc}
        & \textbf{Quantum Algorithm} & \multicolumn{6}{c}{\textbf{Dequantized Algorithms}} \\
        \hline

        \begin{tabular}[x]{@{}r@{}}\textbf{Rec.}\\[-1em]
        \begin{tabular}[x]{@{}r@{}}\textbf{Systems}\\[-1em]
        \begin{tabular}[x]{@{}r@{}}\cite{kerenidis2016quantum}, \cite{tang2019quantum}, \cite{chia2022sampling}, \\[-1em]\cite{chepurko2022quantum}, \cite{bakshi2023improved}\end{tabular}
        \end{tabular}
        \end{tabular}     
        & $\displaystyle \frac{\|A\|_F}{\sigma}$ & $\displaystyle \frac{\|A\|^{24}_F}{\sigma^{24} \varepsilon^{12}}$, & $\displaystyle \frac{\|A\|^{6}_F \|A\|^{10}}{\sigma^{16} \varepsilon^{6}}$, & $\displaystyle \frac{\|A\|^6_F}{\sigma^{6} \varepsilon^{6}}$, & $\displaystyle \frac{\|A\|^4_F}{\sigma^{9} \varepsilon^{2}}$ &  & \\

        \begin{tabular}[x]{@{}r@{}}\textbf{Supervised}\\[-1em]
        \begin{tabular}[x]{@{}r@{}}\textbf{Clustering}\\[-1em]\cite{lloyd2013quantum}, \cite{tang2021quantum}, \cite{chia2022sampling} \end{tabular}
        \end{tabular} 
        
        & $\displaystyle\frac{\|M\|^2_F \|w\|^2}{\varepsilon}$ & $\displaystyle\frac{\|M\|^4_F \|w\|^4}{\varepsilon^2}$, & $\displaystyle\frac{\|M\|^4_F \|w\|^4}{\varepsilon^2}$ &  &  &  & \\

        \begin{tabular}[x]{@{}r@{}}\textbf{PCA}\\[-1em]\cite{lloyd2014quantum}, \cite{tang2021quantum}, \cite{chia2022sampling} \end{tabular} & $\displaystyle \frac{\|X\|_F \|X\|}{\lambda_k \varepsilon}$ & $\displaystyle \frac{\|X\|^{36}_F}{\|X\|^{12} \lambda^{12}_k \eta^6 \varepsilon^{12}}$, & $\displaystyle \frac{\|X\|^{6}_F}{\|X\|^{2} \lambda^{2}_k \eta^6 \varepsilon^{6}}$ & &  &  &  \\

        \begin{tabular}[x]{@{}r@{}}\textbf{Matrix}\\[-1em]
        \begin{tabular}[x]{@{}r@{}}\textbf{Inversion}\\[-1em]
        \begin{tabular}[x]{@{}r@{}} \cite{gilyen2019quantum}, \cite{gilyen2018quantum}, \cite{chia2022sampling},\\[-1em]
        \begin{tabular}[x]{@{}r@{}}\cite{gilyen2022improved}, \cite{chepurko2022quantum}, \cite{shao2022faster}\\[-1em]
        \cite{bakshi2023improved}
        \end{tabular}
        \end{tabular}
        \end{tabular} 
        \end{tabular} 
        
        & $\displaystyle \frac{\|A\|_F}{\sigma}$ & $\displaystyle \frac{k^6 \|A\|^6_F \|A\|^{16}}{\sigma^{22}\varepsilon^6}$, & $\displaystyle \frac{\|A\|^6_F \|A\|^{22}}{\sigma^{28}\varepsilon^6}$,  & $\displaystyle \frac{\|A\|^6_F \|A\|^6}{\sigma^{12}\varepsilon^4}$, & $\displaystyle \frac{\|A\|^4_F \log(c)}{\sigma^8 \varepsilon^4}$, & $\displaystyle \frac{\|A\|^6_F \|A\|^2}{\sigma^8\varepsilon^2}$, & $\displaystyle \frac{\|A\|^4_F}{\sigma^{11} \varepsilon^2}$ \\
        

        \begin{tabular}[x]{@{}r@{}}\textbf{SVM}\\[-1em]\cite{rebentrost2014quantum}, \cite{ding2021quantum}, \cite{chia2022sampling}
        \end{tabular}
        & $\displaystyle \frac{1}{\lambda^3 \varepsilon^3}$ & $\displaystyle \text{poly}\left( 
        \frac{1}{\lambda}, \frac{1}{\varepsilon} \right)$, & $\displaystyle \frac{1}{\lambda^{28} \varepsilon^6}$ & & & & \\

        \begin{tabular}[x]{@{}r@{}}\textbf{SDP}\\[-1em]\cite{van2018improvements}, \cite{chia2019quantum}, \cite{chia2022sampling}\end{tabular}
        
        & $\displaystyle \frac{\|A^{(\cdot)}\|^7_F}{\varepsilon^{7.5}} + \frac{\sqrt{m} \|A^{(\cdot)}\|^2_F}{\varepsilon^4}$  & $\displaystyle \frac{mk^{57}}{\varepsilon^{92}}$, & $\displaystyle \frac{\|A^{(\cdot)}\|^{22}_F}{\varepsilon^{46}} + \frac{\sqrt{m} \|A^{(\cdot)}\|^{14}_F}{\varepsilon^{28}}$ & & & & \\

        \begin{tabular}[x]{@{}r@{}}\textbf{QSVT}\\[-1em]
        \begin{tabular}[x]{@{}r@{}}\cite{gilyen2019quantum}, \cite{chia2022sampling}, \cite{jethwani2019quantum},\\[-1em] \cite{bakshi2023improved}
        \end{tabular}
        \end{tabular} 
        & $\displaystyle \frac{d \|A\|_F \|b\|}{p^{(QV)} (A)b}$ & $\displaystyle \frac{d^{22} \|A\|^6_F}{\varepsilon^6}$, & $\displaystyle \frac{\|A\|_F^6\kappa^{20}(d^2 + \kappa)}{\varepsilon^6}$, & $\displaystyle \frac{d^{11} \|A\|^4_F}{\varepsilon^2}$ &  &  & \\

        \begin{tabular}[x]{@{}r@{}}\textbf{HS}\\[-1em] \cite{gilyen2019quantum}, \cite{chia2022sampling}, \cite{bakshi2023improved} \end{tabular} & $\displaystyle \|H\|_F$ & $\displaystyle \frac{\|H\|^6_F \|H\|^{16}}{\varepsilon^6}$, & $\displaystyle \frac{\|H\|^4_F \|H\|^{9}}{\varepsilon^2}$ & & & & \\

        \begin{tabular}[x]{@{}r@{}}\textbf{DA}\\[-1em] \cite{cong2016quantum}, \cite{chia2022sampling}
        \end{tabular} & $\displaystyle \frac{\|B\|^7_F}{\varepsilon^3 \sigma^7} + \frac{\|W\|^7_F}{\varepsilon^3 \sigma^7}$ & $\displaystyle \frac{\|B\|^6_F \|B\|^4}{\varepsilon^6 \sigma^{10}} + \frac{\|W\|^7_F \|W\|^{10}}{\varepsilon^6 \sigma^{16}}$ &  &  &  & & \\
        \hline
    \end{tabular}
    }
    \label{tab:complexities}
\end{table*}
\setlength\tabcolsep{6pt}

\subsection{Tensor Networks}\label{sec:tensor_networks}

A large focus in QiML research in the current day has been on the use of tensor networks (TNs) as a machine learning models. QiML-based tensor network research benefits from a rich body of prior knowledge\footnote{It is worth noting that TNs have been widely discussed and analyzed in the literature outside of the machine learning context, with a multitude of informative, introductory materials available on the topic \cite{orus2014practical, montangero2018introduction, biamonte2017tensor, bridgeman2017hand, wang2023tensor, rieser2023tensor}.} in the classical context, both in and out of the machine learning field; theory and practical applications show a useful degree of transferability into the quantum domain \cite{huggins2019towards, araz2022classical}. A key motivator in the use of TNs is their ability to classically simulate the many-body quantum wavefunction efficiently, which has been shown to be consistent with higher-order tensor representations \cite{biamonte2017tensor}. Hamiltonians of many physically realistic systems tend to exhibit strong locality; the interactions between constituent particles are limited to next or nearest neighbors \cite{eisert2010colloquium}. In the case of gapped, local Hamiltonians, the exponentially large and intractable Hilbert space is constrained to low energy states bounded by the entanglement area-law, i.e., these states cannot be highly entangled \cite{eisert2010colloquium}. This constricts the exploration space to only a relevant fraction of the Hilbert space. When modeled by tensor networks, approximating this subset can be done in polynomial time \cite{orus2014practical}. 

Further, tensor networks form a bridge between classical neural network methods and quantum computing. Several works have identified the natural corollary of tensor networks to quantum circuits, where many tensor networks have a direct quantum circuit translation \cite{biamonte2017tensor, huggins2019towards}. Complex tensors under tensor network decomposition are represented as unitary gates; the bond dimension connecting two nodes of the tensor network is determined by the number of qubits connecting two sequential unitaries in the circuit. It is shown that qubit-efficient tree tensor network (TTN) and matrix product state (MPS) models can be devised, with logarithmic scaling in the number of physical ($O(1)$, independent of the input data size) and ancilla ($O(\log_2 D)$ in bond dimension $D$) qubits required \cite{wall2021generative}. This allows tensor networks to express higher order feature spaces in classical settings, and act as classical simulators for quantum circuits. Several works have since successfully implemented tensor networks as parameterized quantum circuits on small, near-term quantum devices for machine learning tasks, with many of the advancements in the classical setting being readily transferred to the quantum domain \cite{grant2018hierarchical, wall2021tree, wall2021generative, dborin2022matrix}. 

The many-body quantum system of \(N\) particles can be described as follows:

\begin{equation}
\label{eqn:intro_mbqs}
|\Psi\rangle=\sum_{s_1 s_2 \cdots s_N} \Psi^{s_1 s_2 \cdots s_N}\left|s_1 s_2 \cdots s_N\right\rangle
\end{equation}
where, in quantum computing, the quantum state of \(N\) qubits \(|\Psi\rangle\) with amplitude \(\Psi^{s_1 s_2 s_3 \cdots s_N}\) is a composition of the \(s_N\) single-qubit basis states. This can be effectively considered as one, large tensor. TNs aim to find alternative representations of this computationally-inefficient description by reducing the complexity of the system. This is achieved by decomposing the large tensor into a network of many smaller tensors of smaller rank. The total number of parameters in the final representation scales sub-exponentially with the number of composite tensors and the bond dimension (the dimension of the largest contracted index within the tensor network) between them, which allows for the classical computation of expectation values \cite{orus2014practical}. For example, a tensor with $N$ indices, each of dimension $d$, must generally be specified by $d^N$ parameters. In contrast, the MPS representation of such a tensor with bond dimension $m$ only requires $Ndm^2$ parameters, which now scales linearly with $N$ \cite{biamonte2017tensor}. Various such methods of tensor network decomposition exist, which depend on the properties of the original tensor and the desired resulting network. In further sections, we discuss relevant tensor network methods and their application as QiML techniques.


\begin{figure}[!ht]
    \centering
    \centerline{\includegraphics[width=0.5\textwidth]{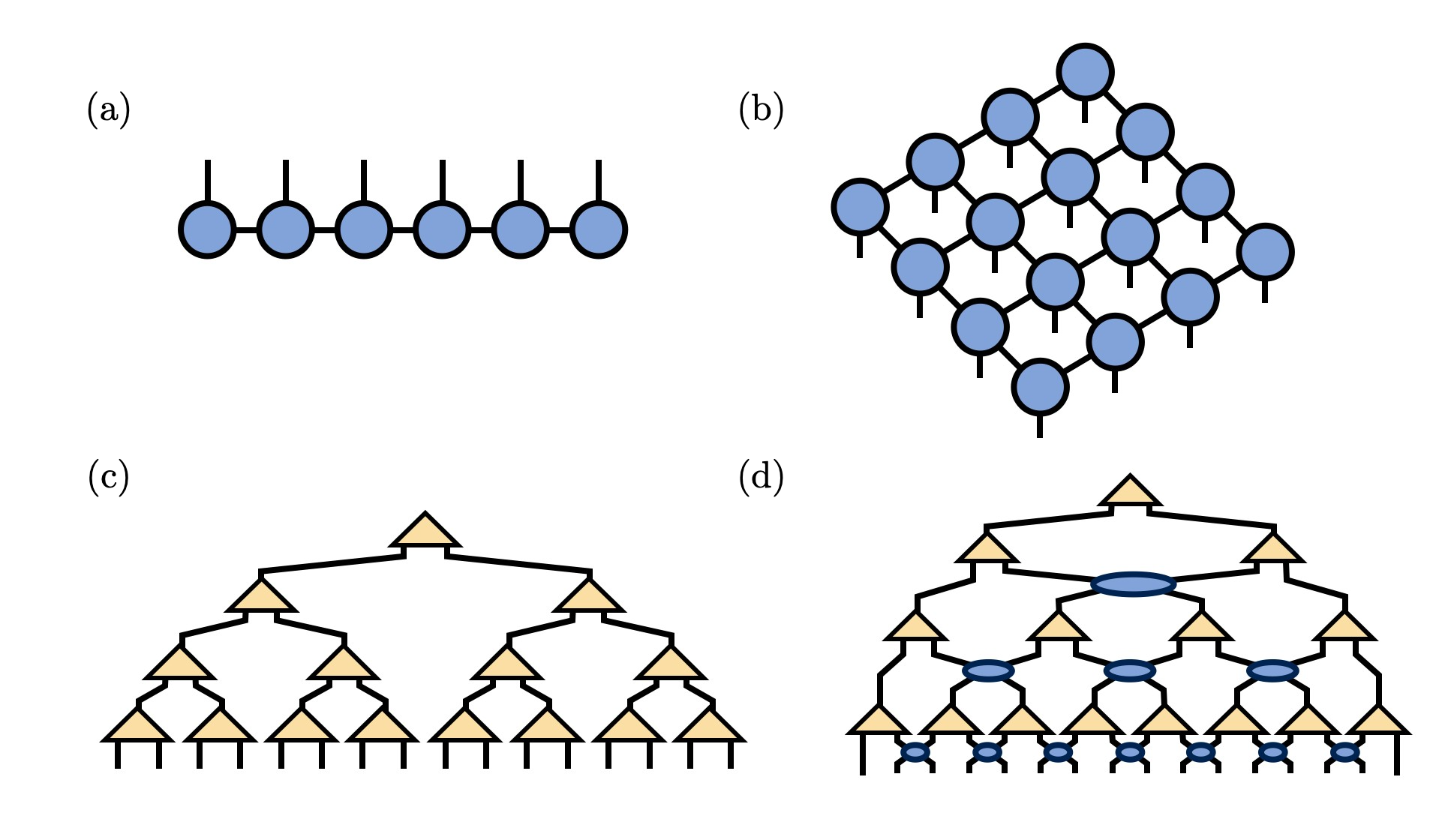}}
    \caption{Common Tensor Network Decompositions. (a): Matrix Product State (MPS), (b): Project Entangled-Pair States (PEPS), (c): Tree Tensor Network (TTN), (d): Multi-scale Entanglement Renormalization Ansatz (MERA) \cite{stoudenmire2018learning}}
    \label{fig:tn_decomps}
\end{figure}
 

\subsubsection{Matrix Product States}

Matrix Product States (MPS), or Tensor Train Networks, are widely used for the efficient representation of 1D gapped quantum systems with low energy states \cite{perez2006matrix}. In fact, any such quantum state can be exactly represented by the MPS structure efficiently \cite{hastings2007area}. The MPS is achieved through the decomposition of the quantum state into a product of smaller matrices which represent the constituent tensors. A general process involves successively partitioning away individual tensors from the rest of the system via the singular value decomposition (SVD) \cite{bridgeman2017hand}; several other related methods have been also been proposed for performing this decomposition in a practical manner \cite{paeckel2019time}. By considering only relevant states with bounded entanglement, a compressed form is obtainable by truncating the least relevant singular values. This low rank approximation is what drives the efficient representation of the quantum state by the MPS, which would otherwise be described exponentially with an \(n\)-qubit state.

\textbf{Supervised Tensor Network Modeling:} The first notable instance of tensor networks being used as machine learning models utilize the MPS decomposition. Stoudenmire and Schwab \cite{stoudenmire2017supervised} demonstrated their capabilities in parameterized supervised learning. They note that handling high-dimensional vectors is also a critical challenge in non-linear kernel learning. Traditional approaches often rely on the “kernel trick” \cite{scholkopf2018learning} to solve the dual representation; instead the authors advocated for applying tensor network decompositions. In their proposed model, the kernel learning problem is expressed as:

\begin{align}
\label{eqn:tn_kernel}
    f^l(x) = W^l \cdot \Phi(x).
\end{align}
Here, vectors $x$ are mapped to a high-dimensional feature space through a non-linear mapping $\Phi(x)$, with $W^l$ as the weight matrix; $l$ is a tensor index for the function $f^l(x)$ that maps every $x$ to the space of classification labels. Both $W$ and $\Phi(x)$ could be considerably large, which presents large computational bottleneck scaling exponentially with their size; i.e. the resulting Hilbert space of $\Phi(x)$ is $2^N$, and necessitates a compatibly sized $W$. An MPS tensor decomposition is thus leveraged to reduce this computational complexity. By representing $W$ via an MPS and optimizing it directly, the model scales linearly with the training set size. An MPS decomposition can derived from $W^l$:

\begin{align}
\label{eqn:tn_mps_weight_decomp}
    W^l_{s_1 s_2 \dots s_N} = \sum_{\{a\}} A^{\alpha_1}_{s_1} A^{\alpha_1 \alpha_2}_{s_2} \cdots A^{l: \alpha_j \alpha_{j+1}}_{s_j} \cdots A^{\alpha_{N-1}}_{s_N},
\end{align}
where index $\alpha_j$ is the bond dimension between sites. $\Phi(x)$ is obtained via an embedding scheme that converts classical input data into a linear combination of quantum states in an orthogonal basis. In \cite{stoudenmire2017supervised}, the quantum mapping function $\phi (x_j)$ is used to embed every $j$-th gray-scale pixel value

\begin{align}
    \label{eqn:tn_trigonometric_basis}
    \phi (x_j) = \left[\cos\left(\frac{\pi}{2}x_j\right), \sin\left(\frac{\pi}{2}x_j\right)\right]
\end{align}
into the L2-normalized trigonometric basis. A full image is thus the tensor product of $\phi(x_j)$ individual local embeddings over all $N$ pixels:

\begin{align}
    \label{eqn:tn_feature_mapping}
    \Phi(\mathbf{x})=\phi \left(x_1\right) \otimes \phi \left(x_2\right) \otimes \cdots \otimes \phi \left(x_N\right).
\end{align}
Words and sentences within natural language documents can be associated with quantum systems by building their vector and tensor space representations \cite{zhang2018quantumqmwf, zhang2019generalized, miller2021tensor}. Each word $w_i$ is mapped to an $m$-dimensional vector space, with each dimension representing a different semantic meaning. A word can thus be represented as a linear combination of these orthogonal semantic bases:

\begin{align}
    \label{eqn:tn_semantic_basis}
    w_i = \sum_{i=1}^{m} \alpha_{i} \ket{e_i}, 
\end{align}
where $\alpha_{i}$ is the coefficient of the $i$-th base vector $e_i$. A sentence $s = (w_i, \dots, w_n)$ or length $n$ can then be modeled as a tensor product of the word vectors:


\begin{equation}
\begin{split}
    s & = w_1 \otimes \cdots \otimes w_n 
    \\ & = \sum_{i,\ldots,n=1}^{m} A_{i,\ldots,n} (e_{i} \otimes \cdots \otimes e_{n}),
\end{split}
\end{equation}\label{tn:semantic_qmbs}
with $A$ being the $m^n$-dimensional coefficient tensor of basis states. Trainable word embeddings, similar to those used in recurrent neural networks (RNN) that treat the word embeddings as variational parameters, have also been proposed; this approach has shown strong predictive performance \cite{tangpanitanon2022explainable}.

A few other local embedding methods have been discussed in the literature, including the polynomial basis \cite{novikov2018exponential}:

\begin{align}
    \label{eqn:tn_polynomial_basis}
    \phi (x_j) = \left[1, x_j\right],
\end{align}
which enables a transformed feature space capturing interactions within categorical data, and simplifying the interpretation of the resulting model as a high-degree polynomial. Other kernels that are the product of some $N$ local kernels could also potentially be used \cite{stoudenmire2017supervised}.

A quadratic cost function:

\begin{align}
    L = \frac{1}{2} \sum^{N_T}_{n=1} \sum_l (y_n^\ell - f^\ell (\mathbf{x}_n))^2,
\end{align}
is optimized via a “sweeping” algorithm, inspired by the density matrix renormalization group (DMRG) algorithm successfully used in physics applications \cite{schollwock2011density}. This process essentially involves "sweeping" across the MPS (Matrix Product State), where only two adjacent MPS tensors are varied at a time. The tensors at sites $j$ and $j+1$ are combined into a single bond tensor $B^\ell$, followed by the calculation of the derivative of the cost function with respect to the bond tensor for a gradient descent step. The gradient update to the tensor $B^\ell$ can be computed as:

\begin{align}
    \Delta B^{\ell}=-\frac{\partial L}{\partial B^{\ell}}=\sum_{n=1}^{N_T}\left(y_n^{\ell}-f^{\ell}\left(\mathbf{x}_n\right)\right) \tilde{\Phi}_n ,
\end{align}
where $\tilde{\Phi}_n$ is the projection of the input via the contraction of the “outer ends” of the MPS that do not include $B^\ell$. $B^\ell$ is then replaced by the updated bond tensor: $B^{'\ell} = B^\ell + \alpha \Delta B^\ell$, where $\alpha$ is a scalar value that controls convergence. This can then be decomposed back into separate MPS tensors using a singular value decomposition (SVD), which assists in adapting the MPS bond dimension. The singular value matrix $S$ can then be absorbed into the right singular vector matrix, resulting in the updated sites $A'_{s_j} = U_{s_j}$ and $A'^{\ell}_{s_{j+1}} = SV^{\ell}_{s_{j+1}}$. The MPS form is restored, with the $\ell$ indexing moving to the $j+1$ site. The process then iteratively continues to the next $j+1$ and $j+2$ tensors and returning through the opposite direction when an end node is reached for a predetermined number of “sweeps”. A major advantage to this is that the resulting bond dimension can be chosen adaptively based on number of large singular values. This flexibility allows the MPS form of $W$ to undergo maximum possible compression; the degree of compression can vary for each bond, while still ensuring an optimal decision function. Inference is performed by successively contracting the network until the value $f^l(x)$ is obtained, where the classification $l$ is determined by the largest $|f^l(x)|$.

\begin{figure}[!ht]
    \centering
    \centerline{\includegraphics[width=0.5\textwidth]{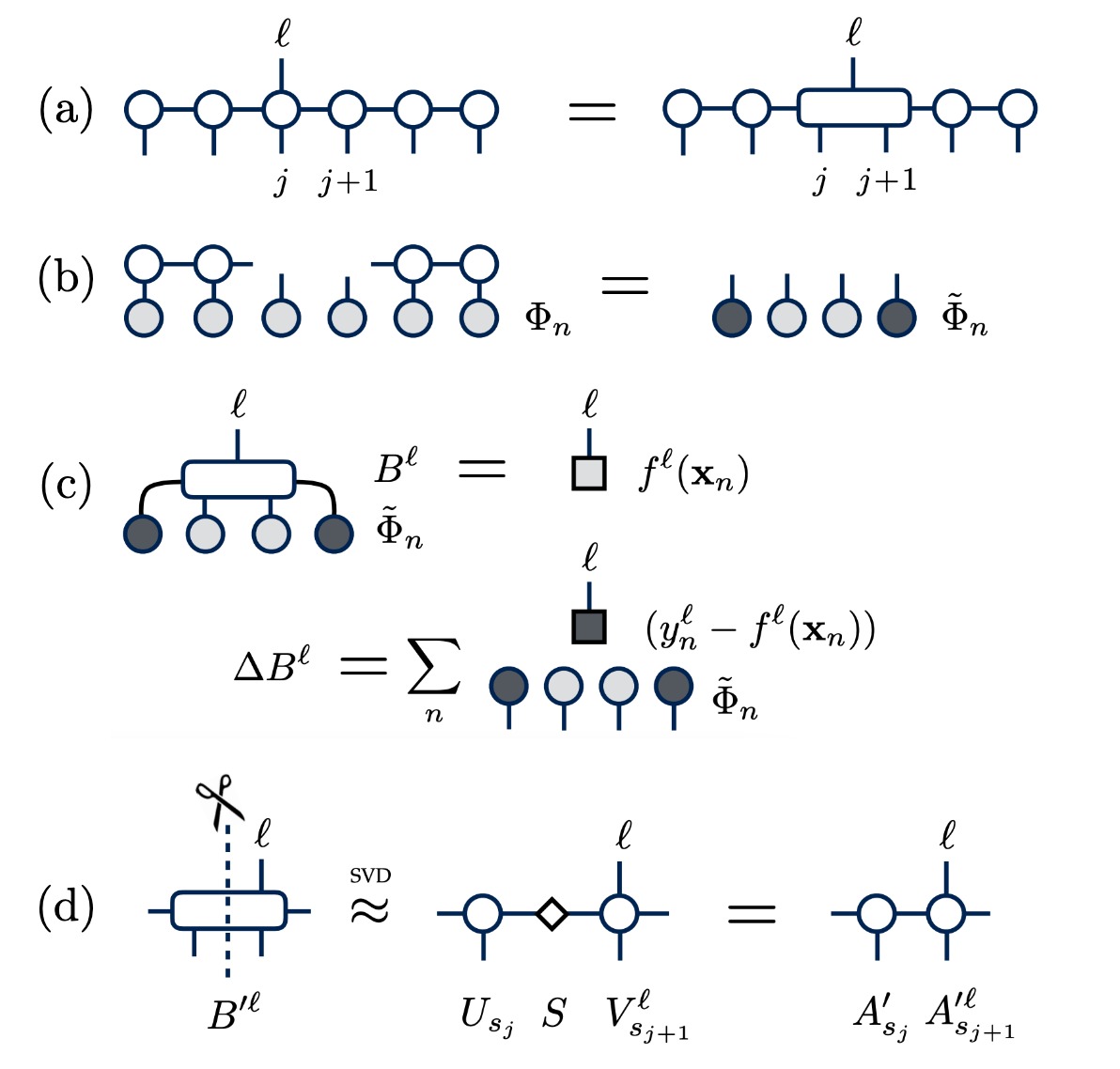}}
    \caption{DMRG Sweeping Algorithm \cite{stoudenmire2017supervised}: (a) Formation of the bond tensor $B^\ell$ at bond $j$; (b) Projection of a training input into the “MPS basis” $\tilde{\Phi}_n$ at bond $j$; (c) Computation of the decision function and the gradient update $\Delta B^\ell$; (d) Restoration of the MPS form by the SVD.}
    \label{fig:tn_dmrg}
\end{figure}

While the DMRG sweeping method has seen success in several works, gradient descent based methods can be used to directly optimize tensor networks \cite{novikov2018exponential, efthymiou2019tensornetwork, sun2020generative, glasser2020probabilistic}. This was first noted by Novikov et al. \cite{novikov2018exponential} who developed parameterizing supervised learning models with MPS in parallel with the work in \cite{stoudenmire2017supervised}. Gradient descent based approaches for minimizing the cost function incurs an increased bond dimension after each iteration, which \cite{stoudenmire2017supervised} handles via successive SVD operations to reduce the rank. To avoid the bond dimension growth, a stochastic Riemannian optimization procedure is used instead, with the addition of a bond dimension regularization term in the cost function. The polynomial basis (Equation \ref{eqn:tn_polynomial_basis}) was used as the local feature mapping.

Efthymiou et al. \cite{efthymiou2019tensornetwork} similarly provide an implementation using automatic gradients for optimization, rather than DMRG. The results show strong performance over benchmark image classification tasks.

Araz and Spannowsky \cite{araz2021quantum} perform experiments comparing the effectiveness of DMRG and SGD approaches in high-energy physics applications, as well as investigating training methods that combined both SGD and DMRG techniques, previously noted to be speculatively compatible \cite{stoudenmire2017supervised}. At each epoch, the first batch has the DMRG applied for a certain number of sweeps, after-which SGD takes over for the remaining batches. This method showed comparable performance with MPS training solely on DMRG or SGD, however the authors note the conflicting aims of the algorithms; DMRG attempts to reduce degrees of freedom of node, while the SGD tries to increase it.

\textbf{Unsupervised Tensor Network Modeling:} The second prominent machine learning formalism for tensor networks is in probabilistic generative modeling. Generative modeling approaches learn the underlying probability distribution that describes a set of data, from which new data instances can be either generated or inferred from the distribution. Quantum states inherently possess a probabilistic interpretation, in which the squared norm of a quantum state's amplitudes gives rise to the probabilities of different outcomes. This connection can be traced back to Born's rule in quantum mechanics \cite{zettili2009quantum}. Thus, Han et al. \cite{han2018unsupervised} first proposed the use of tensor networks for generative modeling. Their model, dubbed by contemporaries as the “Tensor Network Born Machine” (TNBM) \cite{alcazar2020classical}, produces random samples by first encoding probability distributions into quantum states, which are represented by a tensor network. A given dataset $x$ is modelled using a quantum state described by a real-valued wavefunction $\Psi(x)$, which could be some quantum state embedding kernel such as in Equation \ref{eqn:tn_kernel}. This in turn forms the model probability distribution:

\begin{align} \label{eqn:tn_born_machine}
    P(x) = \frac{|\Psi(x)|^2}{Z},
\end{align}
normalized by a partition function $Z = \sum |\Psi (x)|^2$; $|\Psi (x)|^2$ is the energy function of $x$. This form allows for the representation of $\Psi(x)$ via a tensor network. Similar to the supervised setting, both DMRG-like, and gradient descent algorithms can be used for optimization \cite{han2018unsupervised}. The loss function used is the negative log-likelihood (NLL): 
\begin{align}
    L = -\frac{1}{N_T} \sum^{N_T}_{n=1} \ln{P(x_n)}.
\end{align}
While the TNBM formalism has been predominately used in unsupervised contexts, it has also been adapted for supervised learning \cite{sun2020generative}, where classification involves finding the maximum fidelity between a given test example and the learned quantum states $\text{argmax}_c |v^\dagger \Psi_c|$, for $c \in {1, \dots, K}$, where $K$ is the total number of classes. The work in \cite{sun2020generative} demonstrates that the mapping of images into Hilbert space produces natural clustering patterns with and between classes, which was only previously assumed. This explicitly shows advantages in solving machine learning tasks in the many-body Hilbert space as opposed to data-driven feature spaces.

Bit-string classification over parity datasets have suggested that the highly expressive probability distributions encoded by generative MPS networks lead to strong predictive outcomes \cite{stokes2019probabilistic}. Further, MPS decompositions have been shown to handle tasks that deep learning methods cannot. Bradley et al. \cite{bradley2020modeling} note that standard models like Restricted Boltzmann Machines (RBMs) often struggle while learning high-length parity datasets, despite their categorization as universal approximators, theoretically endowed with the necessary expressive capabilities. Results show that MPS models can excel at such tasks, lending further weight to the unique inductive bias provided by tensor networks, echoing findings from other studies such as \cite{sun2020generative}.


\subsubsection{Tree-Tensor Networks/MERA}

Tree Tensor Networks (TTN) (also called the Hierarchical Tucker decomposition \cite{wang2023tensor}) are a tensor network structure where the tensors are arranged hierarchically, often forming a binary tree structure \cite{shi2006classical}. TTNs exhibit advantages in computational efficiency compared with other tensor network forms, which stems from the tree structure that avoids loops in the network, enabling efficient and exact contraction. The Multi-scale Entanglement Renormalization Ansatz (MERA) is a particular type of TTN, which is specifically designed to efficiently represent quantum states with long-range correlations \cite{vidal2007entanglement} due to its incorporation of disentanglers and isomorphisms in its network structure, accounting for and managing entanglement at different length scales. This makes it particularly suitable for representing ground states of critical systems or systems with power-law decay of correlations. While not as general as PEPS or MPS, TTNs and MERA offer advantages in specific contexts where their structure aligns with the physical system. 

The mathematical form of the TTN varies based on the compositional structure, dependent on the number of layers, the number of tensors, and the contraction of those tensors within those layers. The contractions are often defined recursively. An example is given in Equation \ref{eqn:ttn_weight_decomp} for the construction of a binary TTN, with each tensor node connected to two children:

\begin{align}
\label{eqn:ttn_weight_decomp}
W = \sum_{\{\alpha\}} A^{[1]}_{\alpha_2, \alpha_3} \prod^{N}_{n=2} A^{[n]}_{\alpha_{n}, \alpha_{2n}, \alpha_{2n+1}},
\end{align}
where tensor $A^{[1]}$ is the root tensor, with subsequent $A^{[n]}$ children over $N$ total layers \cite{cheng2019tree}. Similar to the MPS, the TTN can be trained via DMRG and gradient-based optimization. Predicting an output is then given by the contraction:

\begin{align}
    \ket{\tilde{p_n}} = W^\dagger \cdot \Phi(x_n).
\end{align}\label{eqn:ttn_prediction}
In addition to gradient-descent based optimization used in the MPS, a MERA-like training process has also been proposed \cite{liu2019machine}. The cost function to be minimized is chosen as:

\begin{align}
    f = - \sum^{N}_{n=1} \braket{p_n|\tilde{p_n}},
\end{align}
with $N$ as the total number of samples. This cost can be reduced by imposing unitary constraints on all tensors $A$ of the TTN such that $A^\dagger A = I$, inducing the whole transformation as a unitary: $W^\dagger W = I$. Thus the simplified cost function becomes:

\begin{equation}
    \begin{split}
        f &= \sum^{N}_{n=1} \left( \bra{\Phi(x_n)}WW^\dagger\ket{\Phi(x_n)} -2\bra{\Phi(x_n)}W\ket{p_n} + 1 \right) \\  &= \sum^{N}_{n=1} \bra{\Phi(x_n)}W\ket{p_n},
    \end{split}
\end{equation}
and is shown to reduce the complexity of optimization. Over the task of Modified National Institute of Standards and Technology (MNIST) \cite{lecun2010mnist} image classification in \cite{liu2019machine}, this learning method exhibited relatively small entanglement between classification states, meaning that the TTN efficiently represents the MNIST dataset; a conjecture that may extend over classical images in general.

2D hierarchical structures have also been proposed for generative modeling as a direct extension of the MPS version in \cite{han2018unsupervised}, where the modeling of 2D images can be directly achieved \cite{cheng2019tree}. This was shown to overcome the issue of exponential decay of correlations in MPS, making it more effective in capturing long-range correlations and perform better for large-size images. 

TTNs have been used as a means of coarse-grained unsupervised feature extraction in \cite{stoudenmire2018learning}. The model prepares quantum states from classical input and feeds them into a 1D TTN. Optimal weights $W$ are computed from its left singular vectors $W_s = \sum_n \beta_n U_s^{\dagger n}$. The basis $U^\dagger$ diagonalizes the feature space covariance matrix $\rho^{s'}_s = \sum_n U_n^{s'} P_n U_s^{\dagger n}$. Since direct diagonalization of $\rho$ is not feasible, the DMRG-like algorithm is employed to iteratively produce and diagonalize reduced density matrices. This procedure is repeated $log(N)$ times to produce a suitable approximation, leading to the diagonalizing $U$ of isometry tensors, approximated as a layered TTN. To produce a classification output, the reduced feature set is subsequently used in a supervised context, where the layers of $U$ are fixed and the top tensor is replaced with an MPS decomposition, optimized via DMRG. The authors note that the method is akin to a direct computation of kernel PCA in feature space.

\subsubsection{Projected Entangled-Pair States}

Projected Entangled-Pair States (PEPS) are a natural extension of MPS to higher-dimensional systems \cite{verstraete2004renormalization}. Just as MPS provide efficient descriptions of 1D quantum systems, PEPS have been shown to efficiently represent ground states of gapped 2D local Hamiltonians \cite{orus2014practical}. The tensors in PEPS are arranged in a grid-like structure, allowing the PEPS to effectively capture long-range quantum correlations \cite{cirac2021matrix}. The PEPS decomposition has a polynomial correlation decay with respect to the separation distance between parts of the network, whereas the MPS decomposition shows an exponential decay \cite{wang2023tensor}. Like MPS, the bond dimension of PEPS determines the maximum entanglement entropy across any cut of the tensor network. However, contracting the PEPS networks is computationally more challenging than MPS due to the increased tensor connectivity. As such, exact calculations for PEPS are practically infeasible for larger systems, and approximation methods are usually employed \cite{orus2014practical}.

The weight matrix $W$ can be modelled as PEPS decomposition of tensors on an $L \times L$ grid, mapped to some $x \in \mathbb{R}^{L \times L}$ input feature tensor:

\begin{align}
    W^l_{s_1 s_2 \dots s_N}  &= \sum_{\{a\}} A_{s_1}^{\alpha_1,\alpha_2} A_{s_2}^{\alpha_3,\alpha_4,\alpha_5} \cdots \\  &\quad A_{s_j}^{l:\alpha_k,\alpha_{k+1},\alpha_{k+2},\alpha_{k+3}} \cdots A_{s_N}^{\alpha_{K-1},\alpha_K},
\end{align}
where $K$ is the number of bonds in the lattice, and each tensor has a "physical" index $s_j$ connected to the input vector, along with "virtual" indices $\alpha_k$ for contraction with adjacent tensors. A special tensor in the center also has a "label" index $l$ to generate the output vector  \cite{cheng2021supervised}. 

In unsupervised generative modeling, the PEPS model can capture the probability distribution $P(x)$ as a decomposed sum of individual distributions:
\begin{align}
    P(\mathbf{x})=\frac{N_1}{N} P_1(\mathbf{x})+\ldots+\frac{N_m}{N} P_m(\mathbf{x}),
\end{align}
each representing different labels or categories over data, in which the total wavefunction seen as a superposition of these distributions with smaller entanglement \cite{vieijra2022generative}. Each $P_i$ is weighted by the fraction of its category within the total training set $N_i / N$, with the categorisation determined by labels, if present, allowing for supervised generative modeling, or some agnostic clustering algorithm.

Efficient contractions are possible when both image size $L$ and bond dimension $D$ are small; for larger representations, an approximate contraction method is used, which treats the bottom row tensors as an MPS and the rest of row tensors as the operators applied on the MPS, over which the DMRG-like truncating algorithm can operate \cite{cheng2021supervised}. 

The PEPS decomposition has seen success in image modeling. Cheng et al. \cite{cheng2021supervised} note that the method better captures structural and spatial information in images when compared to MPS and TTNs \cite{cheng2021supervised}. Each image pixel is mapped to each PEPS tensor without the need for flattening, such as with MPS decompositions. The grid structure of the PEPS allowed for the exploration of additional feature maps; two are explored, the trigonometric feature map (Equation \ref{eqn:tn_trigonometric_basis}), and an adaptive feature map that leverages convolutional kernels, allowing the PEPS to accept a feature tensor as input. The use of convolutional feature extractors proved more effective in experiments over MNIST and Fashion-MNIST \cite{xiao2017fashion} datasets. Vieijra et al. \cite{vieijra2022generative} explore PEPS for unsupervised generative modeling, which showed greater performance then that of existing MPS and TTN models. The results suggest the enhanced capability of multi-dimensional tensor network structures in unsupervised generative modeling for image classification, with \cite{vieijra2022generative} stating that tensor networks perform better when the network mimics the local structure of the data.

\subsubsection{Matrix Product Operators}\label{sec:mpo}

The Matrix Product Operator (MPO) is an extension of the MPS concept to describe quantum operators, especially in the context of 1D quantum systems \cite{hubig2017generic}. An MPO is expressed as a collection of matrices arranged in a chain, with each matrix indexed by physical indices that account for the ingoing and outgoing states of the operator. This structure allows for a compact and efficient representation of complex quantum operators that would otherwise require an exponentially large amount of information. Further, MPOs can be compressed to approximate a given operator with smaller matrix dimensions. Similar to the MPS, techniques such as SVD can be applied, reducing the bond dimensions while maintaining a controlled approximation error.

\begin{figure}[!ht]
    \centering
    \centerline{\includegraphics[width=0.4\textwidth]{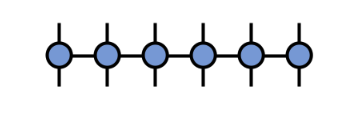}}
    \caption{Diagrammatic Notation of the Matrix Product Operator (MPO) with six Sites \cite{tensornetwork.org}}
    \label{fig:tn_mpo}
\end{figure}

An MPO can be expressed in the form:


\begin{align}
    M^{s_1,\ldots,s_N}_{s'_1,\ldots,s'_N}  &= \sum_{\{a\}} A^{s_1 \alpha_1}_{s'_1} A^{s_2 \alpha_2}_{\alpha_1 s'_2} \cdots A^{s_{j+1} \alpha_{j+1}}_{\alpha_j s'_{j+1}} \cdots A^{s_N}_{\alpha_{N-1} s'_N},
\end{align}
in traditional mathematical notation, or diagrammatically as shown in Figure \ref{fig:tn_mpo} \cite{tensornetwork.org}.

The ability for compressed representation has allowed for the use of MPOs in classical neural network structures. Research has shown that replacing one, or several dense hidden layers in the neural network with a parameterized MPO can allow for significantly decreases the number of parameters involved, effectively compressing the network \cite{novikov2015tensorizing, gao2020compressing}. This transformation maintains the network's ability to express complex relationships and functions, thus preserving its overall performance while also improving the training efficiency and memory
consumption. In \cite{gao2020compressing}, a slight improvement is seen in test accuracies in state-of-the-art models that incorporate the MPO compression, compared to their standard parameterization. The authors suggest that by reducing in the number of parameters, local correlations in input signals are emphasized, and the risk of overfitting is reduced by constraining the linear transformation matrix, which helps avoid trapping the training data in local minima. Patel et al. \cite{patel2022quantum} extend this approach for solving differentiable equations involving in portfolio optimization, to similar results.

Wang et al. \cite{wang2020anomaly} explore the use of MPOs as sparse representations of operators in parameterizing a linear transformation over an exponentially large space with respect to the number of features. The authors highlight the suitability of the model for anomaly detection, as linear models utilizing MPOs can provide control and manage behavior over the entire input space, even when it's vastly imbalanced in terms of inliers and outliers. In the anomaly detection model, data is first embedded into a high-dimensional vector space $V$ using a fixed feature map $\Phi$. “Normal” instances undergoing the transformation would be projected close to the surface of the hypersphere $W$, whereas anomalous instances are mapped close to its center.

\subsubsection{Other Tensor Network Methods}

Several other tensor network decompositions have been suggested for use in probabilistic modeling, particularly over natural language modeling tasks. The Canonical Polyadic (CP) and the Tucker decompositions, similar to decompositions previously mentioned, have been used for their capacity for producing low-rank approximations of high dimensional tensors. Several authors have demonstrated their suitability in modelling the sequential and polysemic natural of words. Hierarchial MPS models have also been proposed, providing a structured way to compute exact normalized probabilities and perform unbiased direct sampling, allowing for efficient training through gradient-based procedures and demonstrating competitive performance on tasks such as image classification with reduced computational resources. Finally, we touch on the use of tensor networks embedded within neural network structures, in which tensor networks have demonstrated their capabilities in model compression.

\textbf{Canonical Polyadic Decomposition:} In Zhang et al. \cite{zhang2018quantumqmwf}, a weighted form of the CP decomposition is used, which factorizes a higher-order tensor into a sum of rank-one tensors, expressed as:
\begin{align}
    W = \sum_{r=1}^{R} \lambda_r \mathbf{a}_{r,1} \otimes \mathbf{a}_{r,2} \otimes \ldots \otimes \mathbf{a}_{r,N}
\end{align}
where $\lambda_r$ are scalar weights and $\mathbf{a}_{r,i}$ are unit column vectors of $M$-dimension: $(a_{r,i,1}, \ldots , a_{r,i,M})^\intercal$, where $M$ is the dimension of the corresponding mode of  $W$. $R$ denotes the minimum possible number of rank-one tensors. The authors propose an approach based on the CP decomposition to account for interaction among polysemic words, creating new compound meanings by combining the different basis vectors of the words. The global representation of all possible compound meanings is captured in the quantum many-body wavefunction of Equation \ref{eqn:tn_semantic_basis}. Solving the resulting high-dimensional tensor is made feasible through the use of tensor decomposition, which then enables the projection of a global semantic space onto a local one for a particular sequence of words. 


\textbf{Recurrent Methods:} Zhang et al. \cite{zhang2019generalized} propose a Recursive Tensor Decomposition, inspired by the MPS and Tucker decompositions. The decomposition of a tensor \( T \) is expressed as:
\begin{align}
    T &= \sum_{i=1}^{r} \lambda_i S_{(n),i} \otimes u_i, \\
    S_{(n),k} &= \sum_{i=1}^{r} W_{k,i} S_{(n-1),i} \otimes u_i
\end{align}
where \( S_{(1)} = 1 \in \mathbb{R}^r \). Here, an \( n \)-order tensor \( T \in \mathbb{R}^{m \times \ldots \times m} \) is decomposed into an \( n \)-order tensor \( S_{(n)} \in \mathbb{R}^{m \times \ldots \times r} \), a diagonal matrix \( \Lambda \in \mathbb{R}^{r \times r} \), and a matrix \( U \in \mathbb{R}^{r \times m} \): $\lambda_i$ and $u_i$ are the $i$-th singular values and left singular vectors resulting from each successive factorization. The parameter \( r \) (with \( r \leq m \)) denotes the rank of the tensor decomposition, and the decomposition can be viewed as a matrix SVD after tensor matricization or flattening by one mode. The method reduces the parameters from \( O(m^n) \) to \( O(m \times r) \), effectively capturing the main features of the tensor \( T \) while significantly reducing the complexity. The decomposition can be used to calculate the conditional  probability \( p(w_t|w_{t-1}^1) \) of sequential words via the softmax function over the inner product of two tensors \( T \) and \( A \), where $A_{t-1}$ is the input of $(t-1)$ words $\alpha_1, \ldots , \alpha_{(t-1)}$. Intermediate variables \( h_t \) are recursively calculated using:
   \[
   h_t = W h_{t-1} \circ U \alpha_t
   \]
   where \( \circ \) denotes element-wise multiplication, and \( W \) and \( U \) are matrices decomposed from the tensor \( T \).


Miller et al. \cite{miller2021tensor} propose the uniform Matrix Product State (u-MPS) model as a type of recurrent tensor network used for processing sequential data. In the u-MPS, all cores of the MPS are identical tensors \( A \) with shape \( (D, d, D) \). The model's recurrent nature allows it to generate \( n \)-th order tensors \( T_n \in R^{d^n} \) for any natural number \( n \), enabling its application to sequential data. For a sequence of arbitrary length \( n \) over an alphabet of size \( d \), a u-MPS can map the sequence to the index of an \( n \)-th order tensor \( T_n \), defining a scalar-valued function \( f_A(s) = \alpha^\intercal A(s) \omega \), where \( A(s) = A(s_1)A(s_2)\ldots A(s_n) \) represents the compositional matrix product of the sequence, and $\alpha$ and $\omega$ are $D$-dimensional vectors that serve as boundary conditions terminating the initial and final bond dimensions of the network. The application of a unitary MPS (u-MPS) model has shown unique generative properties and successful extrapolation of non-local correlations, indicating potential scalability to real-world sequence modeling tasks.


\textbf{Generalized Tensor Networks:} Glasser et al. \cite{glasser2020probabilistic} explored the connection between tensor networks and probabilistic graphical models, from which “generalized tensor networks” are proposed. These allow for input tensors to be copied and reused in other parts of the network, allowing for greater computational efficiency while also involving fewer parameters, and greater expressivity over new types of variational wavefunctions when compared with regular tensor networks. Several generalized tensor network structures are proposed, the discriminative string-bond states (SBS) and entangled plaquette state (EPS) models utilizing tensor copy operations, and are tested in supervised contexts in both image and environmental sound classification tasks. The method showed superior performance to other tensor network learning models; the authors note that these results are achieved over very small bond dimensions compared to previous works.


\textbf{Hierarchical MPS:} Hierarchical MPS models have emerged as a powerful approach for addressing complex machine learning tasks. They are characterized by a tiered structure that emphasizes the flexibility of representation and computational efficiency.

Liu, Zhang and Zhang \cite{liu2023tensor} noted the generally poor performance of prior works in generative tensor network models relative to standard methods. To address this, an autoregressive MPS (AMPS) is proposed, where the joint probability distribution $P(x)$ is not represented by a tensor network directly, but the factorization of it as a product of conditional probabilities; an idea which stems from autoregressive modeling in ML. The model constructs a 2D hierarchical tensor network representation using separate MPS decompositions for individual conditional probabilities. These individual MPS tensor elements can be trained and parameterized through a gradient-based NLL minimization process, demonstrating a significant, theoretical expressive power that exceeds previous tensor network models, backed by empirical investigation.

Selvan and Dam \cite{selvan2020tensor} introduce the Locally Orderless Tensor Network (LoTeNet) model using the theory of locally orderless images that allows it to handle larger images without sacrificing global structure, a deviation from prior models that flatten entire 2D images \cite{stoudenmire2017supervised, han2018unsupervised, efthymiou2019tensornetwork}. LoTeNet begins with a squeeze operation, rearranging local \( k \times k \) image patches and stacking them along the feature dimension, with the stride of the kernel \( k \) determining the reduction in spatial dimensions. The squeezed images with an inflated feature dimension of \( C \cdot k^2 \) are then flattened from 2D to 1D and processed through MPS blocks, embedded into the joint feature space, and contracted to output a vector with dimension \( \nu \). The output vectors from all MPS blocks are reshaped back into 2D space and passed through subsequent layers of the model. After \( L \) layers, the final MPS block performs a decision contraction, resulting in predictions for the \( M \) classes. In evaluation, this model required fewer hyperparameters and significantly less GPU memory than standard deep learning models. This is primarily due to its design of successively contracting input into smaller tensors, thereby avoiding escalating memory consumption with larger images and batch sizes. 

\textbf{Tensor Networks in Neural Network Architectures}. In \ref{sec:mpo}, the benefits of using MPO tensor networks as neural network layers as methods of neural network compression is discussed. Other decompositions have also been suggested for this purpose, such as in \cite{konar20233}, who integrate the Tucker decomposition into a neural network architecture for performing voxel-wise processing for fully automated semantic segmentation of brain and liver volume scans. Using a tensor representation for the high-dimensional weight vector is shown to enhance network operations and extracts critical semantic information, while also facilitating faster convergence and improving precision. 

Tensor networks have also been used as feature extractors for neural networks, which is a natural extension given their kernel-based feature map representations. The resulting low-rank tensors in \cite{zhang2018quantumqmwf} are used as trainable kernel weights fed into a convolutional layer in a classical CNN model.


\subsubsection{Differences in Optimization Methods}

Tensor networks are most commonly optimized using either DMRG sweeping or gradient descent, particularly SGD. A comparison of these methods reveals various strengths and weaknesses:

\textbf{Performance:} As evidenced in Tables \ref{tab:tn_sup_performance} and \ref{tab:tn_unsup_performance}, gradient descent-based optimization generally outperforms DMRG on both supervised Fashion-MNIST and unsupervised binary MNIST. An exception can be found in the method used in \cite{stoudenmire2017supervised} on the supervised MNIST task, which is the second-best performing model. However, the study by \cite{araz2021quantum} observed that SGD and DMRG produced similar classification results over various settings. These findings do not entirely reflect the performances in Tables \ref{tab:tn_sup_performance} and \ref{tab:tn_unsup_performance}, however this may be due to other factors pertaining to the models in question. Understanding the nature of these observed inconsistencies may call for additional examination or targeted studies.

\textbf{Efficiency and Usability:} Gradient descent optimization is often more efficient than DMRG, especially when stochastic gradients are considered \cite{stoudenmire2018learning, araz2021quantum}. DMRG is known to exhibit large computational complexity, especially for decompositions outside of boundary MPS \cite{cheng2021supervised}. This may limit the applicability of the algorithm; we note that DMRG is mostly used in MPS and TTN settings from Tables \ref{tab:tn_sup_performance} and \ref{tab:tn_unsup_performance}, or on small-sized PEPS \cite{cheng2021supervised}. Additionally, the widespread use of GD in machine learning, as well as its suitability for parallelization \cite{efthymiou2019tensornetwork} makes it readily adoptable by ML practitioners.

\textbf{Simplicity and Interpretability:} DMRG provides a simpler learning structure and can adapt the network architecture based on the complexity of the problem, reducing the need for hyperparameter optimization \cite{efthymiou2019tensornetwork}. Furthermore, it provides more interpretable results than SGD, emphasizing inherent data structures for a better understanding of the learned representations \cite{araz2021quantum}.

\textbf{Information Capture:} In image recognition, SGD and DMRG produce similar classification results, but SGD is shown to capture more information due to a larger entropy, potentially outperforming DMRG in scenarios with large data fluctuations. DMRG focuses more on the image's central part and maintains a similar entropy distribution per site \cite{araz2021quantum}. 

\subsubsection{Enhancing Tensor Network Methods} 

Image recognition tasks have served as an experimental ground for enhancing tensor network methods, especially in optimizing computational efficiency, understanding the exploitation of Hilbert feature spaces, and optimizing model selection based on entanglement scaling analysis.

Liu et al. \cite{liu2018entanglement} proposed using quantum entanglement information to guide the learning of MPS architectures for image recognition tasks. By converting images into frequency space using a direct cosine transformation, natural local correlations in 1D space are enhanced, and entanglement structures are shown to prioritize low-frequency data. These factors are captured in lower entanglement entropy across the system. An MPS is used to model these correlations. A MERA-based learning method is used, which minimizes the bipartite entanglement entropy (BEE) by rearranging the MPS contraction path to align the single-site entanglement entropy values (SEE) in descending order. Low SEE sites are discarded. This approach has demonstrated solid accuracy with relatively small bond dimensions on the MNIST dataset.


A few works also tackle the question of selecting the optimal tensor network decomposition for a given task. Convy et al. \cite{convy2022mutual} applied entanglement scaling analysis from quantum physics to classical ML data. For a quantum system represented as a tensor network, the entanglement entropy of the system is bounded by its maximum bond dimension $m$ and $n$ connecting indices. Assuming a fixed $m$ (typically a hyperparameter), the entanglement scaling differs between tensor networks due to $n$, which depends on the network geometry. Consequently, the foremost objective when deploying tensor network ansatze is to align the entanglement scaling dictated by the data, or the quantum state, with the inherent entanglement scaling of the network. The authors proposed the Mutual Information (MI) score to analyze entanglement scaling on classical data. They investigated MI scaling patterns in the MNIST and grayscale Tiny Images \cite{torralba200880} datasets. Results suggested that Tiny Images' MI follows a boundary law, while the findings were less conclusive for MNIST. These insights could guide the selection of tensor networks, with 2D geometries like PEPS being suitable for datasets obeying a boundary law. 

Hashemizadeh et al. \cite{hashemizadeh2020adaptive} proposed a greedy algorithm for tensor network structure learning, aimed at efficiently traversing the space of tensor network structures for common tasks like decomposition, completion, and model compression. They introduced a novel tensor optimization problem that seeks to minimize a loss across diverse tensor network structures with a parameter count constraint. This bi-level optimization problem uniquely involves discrete optimization over tensor network structures at the upper level and continuous optimization of a specific loss function at the lower level. Their approach entails a greedy algorithm to address the upper-level problem, which is then combined with continuous optimization techniques to solve the lower-level problem. Starting with a rank one initialization, the algorithm successively identifies the most promising edge in a tensor network for a rank increment. This allows for the adaptive identification of the optimal tensor network structure for a given task, directly from the data, evidenced by experimental results over image reconstruction.

\subsubsection{Understanding Neural Networks via Tensor Networks}\label{sec:tn_understanding}

Aside from the use of tensor networks for learning tasks, the application of these methods to machine learning can help us understand the machinery of neural networks in new ways. 


Cohen et al. \cite{cohen2016expressive} provides comprehensive examination of the representational capabilities deep learning structures by finding equivalences between neural networks and tensor-network architectures in the context of probabilistic modeling, including structures like non-negative matrix product states and Born machines. This research underscored the fundamental advantage of deep networks over their shallow counterparts, demonstrating that functions that could be efficiently represented by a deep CNN of polynomial size would necessitate an exponential size for approximation by a shallow CNN, which was previously a conjectural theorem. Moreover, the study highlighted the extraordinary power of neural network depth in exponentially reducing the need for breadth with each additional layer, although the exact class of expressible functions remains a subject of ongoing exploration. Subsequent work further extends this analysis to recurrent neural networks, showcasing their natural correlation with MPS and validating Cohen's theorem in this context \cite{khrulkov2017expressive}.

Glasser et al. \cite{glasser2019expressive} analyze the expressive power of tensor network formalisms, highlighting unbounded disparities in resource requirements for modeling certain distributions. This is reinforced by studies indicating that 1D tensor networks are inefficient for text, as mutual information scales with an exponent close to a volume law, while 2D tensor networks like PEPS may be suitable for images due to an area-law scaling \cite{lu2021tensor}. Tangpanitanon et al., \cite{tangpanitanon2022explainable}, however, demonstrate that MPS variational ansatze may in fact be suitable for sequence modeling tasks such as NLP, particularly for sentiment analysis of movie reviews. The suitability arises not from entanglement entropy but from the high-dimensional word vector embedding, which allow for strong predictive accuracies in sentiment analysis. The authors also reported a phenomenon of entanglement entropy saturation with implications on machine learning model selection; as the model size grows, word embedding becomes key to increasing model expressiveness, putting forward the MPS structure as a viable approach in NLP. This work stands in contrast to arguments against MPS's usefulness in NLP \cite{stokes2019probabilistic, glasser2019expressive, lu2021tensor}, highlighting instead the significance of high-dimensional word embedding.


Gao and Duan \cite{gao2017efficient} show that efficient Deep Boltzmann Machine (DBM) representations can be efficiently constructed from any tensor network state, including PEPS and MERA, provided a deep enough neural network. In essence, assuming quantum state presentation is a P/poly class problem, deep neural networks possess the capability of representing the majority of physical states efficiently, including the ground states originating from many-body Hamiltonians and states that emerge from quantum dynamics. Chen et al. \cite{chen2018equivalence} show that under certain conditions, the reverse is also true --- TN states can be converted into RBMs if they describe a non-entangled quantum system. Levine et al. \cite{levine2019quantum} furthers the work in \cite{gao2017efficient} and extends the proof to CNNs and RNNs --- when considered under TN representations, these neural networks can efficiently represent entangled quantum systems.

\subsection{Quantum Variational Algorithm Simulation (QVAS)}\label{sec:quantum_variational_algorithm_simulation}

In view of the categorization of QiML within the “classical-classical” (CC) mode of QML, as indicated in Figure \ref{fig:intro_intersections}, one could interpret QiML as the application of classical data to quantum circuits, all simulated on classical hardware, to process machine learning tasks. This interpretation then overlaps QiML with the “classical-quantum” (CQ) aspect of QML, which encompasses concepts such as variational quantum circuits where optimization of quantum parameters is offloaded to classical computation methods. This research area delves into the potential capabilities of quantum computing in anticipation of the realization of quantum hardware. It is worth noting that despite the common disjunction between references to QiML and classical-quantum based ML in the literature, the common denominator remains that classically simulated quantum circuits are indeed run on classical hardware. Considering this perspective, the task becomes discerning those studies that not only investigate practical machine learning applications but also develop techniques particularly intended for classical hardware simulation. These constitute a minor subset within the vast body of QML literature, where the CQ paradigm takes center stage. Moreover, using naive keyword search terms such as "quantum simulation" may not yield desired results, as this term already denotes an established research domain \cite{georgescu2014quantum, daley2022practical}. 

Our survey into this area is thus informed by both our own keyword searches, and the recent reviews of other authors exposing the interested subset, such as \cite{gujju2023quantum}. We target works that use quantum simulation and attempt to further machine learning tasks of interest by some metric (performance, speed, resource consumption, etc.), over purely classical implementations.

In the following section, we highlight recent advancements in simulating quantum computing, present a concise overview of QML learning frameworks, and discuss recent practical applications in this field. For a more thorough understanding, we direct interested readers towards comprehensive resources such as \cite{zhang2020recent, cerezo2021variational, schuld2021machine, alchieri2021introduction, de2022survey, garcia2022systematic, gujju2023quantum}.

\subsubsection{Frontiers of Classical Simulation}

The successful emulation of quantum computations on classical hardware hinges on the capability to simulate qubits and their potential for exponential information storage effectively. Determining the boundary between classical and quantum computation elicits a discussion on quantum supremacy \cite{preskill2012quantum}; finding the exact crossover point, beyond which a quantum system becomes infeasible for classical computer simulation, remains a complex issue \cite{zhou2020limits}. Classical resources needed for such simulations scale exponentially with the number of qubits and the depth of the quantum circuit \cite{zhou2020limits}, marking an exponential cost in their classical parameterization. 

Despite these challenges, researchers have pushed the boundaries of classical hardware capabilities. Strategies such as data compression \cite{wu2019full}, optimized circuit partitioning \cite{chen201864}, and large-scale batching methods \cite{pan2022simulation} have enabled the simulation of many tens of qubits. However, these frontiers are largely restricted to supercomputing, or high-performance platforms. For users without access to such architectures, the possibilities are considerably more limited; a PC equipped with 16GB of GPU memory can simulate approximately 30 qubits \cite{xu2023herculean}. Given this limitation, many practical applications of QML operate within this smaller qubit range. As such, we will focus on studies that have achieved promising results on these smaller-scale, more accessible devices.

For a comprehensive exploration of the challenges involved in the practical simulation of quantum computers, we refer the interested reader to the work by Xu et al. \cite{xu2023herculean}.

\subsubsection{Encoding Classical Data}

Classical data must be processed through an encoding mechanism for quantum settings, wherein an $m$-dimensional classical dataset is mapped onto a quantum state vector within Hilbert space. This procedure permits us to leverage the vast feature space in quantum systems, thereby offering superior representational power in comparison to classical feature spaces. We outline common encoding schemes below:

\begin{enumerate}
\item \textbf{Basis Encoding:} Also known as computational basis encoding, it is the simplest way of encoding classical data. Given a classical vector $x = (x_1, x_2, ..., x_n)$, where $x_i \in {0,1}$, each classical bit $x_i$ is encoded onto the state of the $i$-th qubit. An $n$-bit classical string is directly encoded into a quantum state of $n$ qubits:

\begin{equation}
|x\rangle = |x_1, x_2, ..., x_n\rangle.
\end{equation}

\item \textbf{Amplitude Encoding:} This method allows efficient encoding of classical data, taking advantage of the exponentially large Hilbert space. In amplitude encoding, an $n$-dimensional normalized real vector $x = (x_1, x_2, ..., x_n)$ such that $\sum_{i=1}^{n} |x_i|^2$ is encoded into the amplitudes of a quantum state. This requires at least $\log_2(n)$ qubits, where $n$ is the dimension of the classical vector. The encoded state is:

\begin{equation}
    \ket{x} = \sum_{i=1}^{n} x_i \ket{i}.
\end{equation}

\item \textbf{Angle Encoding:} In angle encoding, data is encoded into the angles of rotational gates. Given a classical vector $x = (x_1, x_2, ..., x_n)$, each value $x_i$ is used as a parameter in a rotation gate applied to the $i$-th qubit. For example, with $R_y$ rotations, the encoded state is:

\begin{equation}
\ket{x} = R_y(x_1) \otimes R_y(x_2) \otimes ... \otimes R_y(x_n) \ket{0}^{\otimes n}.
\end{equation}

Note the similarities between this and Equations \ref{eqn:tn_kernel} and \ref{eqn:tn_trigonometric_basis}; a possible $R_y$ gate rotation could produce the embedding: 

\begin{equation}
    \begin{split}
        \ket{x} & = \bigotimes\limits^{n-1}_{i=0} R_y(x_i) \ket{0}^{\otimes n} \\ & = \bigotimes\limits^{n-1}_{i=0} \cos\left(\frac{\pi}{2}x_j\right)\ket{0} + \sin\left(\frac{\pi}{2}x_j\right)\ket{1}.
    \end{split}
\end{equation}\label{eqn:angle_encoding_trig}

\end{enumerate}
In each of these methods, classical data is encoded into the quantum state space, and these encoded states are then used as inputs to quantum circuits. Different encoding methods can lead to different computational advantages, and the choice of encoding is often problem-specific.

Once classical data is embedded into the quantum space, it can be manipulated via various QML algorithms. These algorithms are diverse and span a broad range of types, stemming from various mathematical bases \cite{garcia2022systematic, gujju2023quantum}. Some algorithms, such as Quantum Boltzmann Machines, are known to be BQP complete, implying they cannot be effectively simulated on classical computers \cite{amin2018quantum}. Conversely, there are QML algorithms that remain within the realm of classical simulation, up to classical computing limits. These typically include quantum kernel and quantum variational methods \cite{gujju2023quantum}.

\subsubsection{Quantum Kernel Methods}

Quantum kernel methods utilize quantum devices to compute kernel functions, thus capturing the similarity between data points in a feature space. Notably, these methods have the potential to provide exponential speedups for specific problems while still being classically tractable. A frequently employed quantum kernel method is the Quantum Kernel Estimator (QKE) \cite{havlivcek2019supervised}. This method involves defining a quantum feature map, $\phi(x)$, that transforms classical data, $x$, into quantum states via unitary operations on a quantum circuit $U_{\phi}(x)$. This is performed by applying the circuit to an initial state, commonly chosen as $\ket{0}^{\otimes n}$:
    
\begin{align}
    \label{eqn:encoding_circuit}
    \ket{\Phi(x)} = U_{\phi}(x)\ket{0}^{\otimes n},
\end{align}
where $U_{\phi}(x)$ can be considered as a unitary that produces quantum states $\ket{\Phi(x)}$ based on a chosen feature mapping $\phi$, and $n$ is the number of qubits. For any two data points, $x_i$ and $x_j$ in the dataset $D$, their corresponding encoded states are $\Phi(x_i)$ and $\Phi(x_j)$. The kernel entry between $x_i$ and $x_j$ is given by:

\begin{align}
    \kappa(x_i, x_j) &= |\braket{\Phi(x_j)|\Phi(x_i)}|^2  \\ &= |\bra{0}^{\otimes n} U^\dagger_{\phi}(x) U_{\phi}(x)\ket{0}^{\otimes n}|^2, 
\end{align}
representing the inner product of the two feature vectors in the quantum state space. The computation is achieved by approximation in computing the overlap of quantum states $\Phi(x_i)$ and $\Phi(x_j)$. The quantum kernel can then be used to construct a Quantum Support Vector Machine (QSVM) that integrates the quantum kernel (constructed via QKE) with a classical SVM, replacing the kernel $K(\textbf{x}_i, \textbf{x}_j)$ in Equation \ref{eqn:bg_svm} \cite{havlivcek2019supervised}. This method essentially processes data classically and uses the quantum state space as feature space, enabling the use of high-dimensional, non-linear feature mappings that are difficult to compute classically.

The optimal choice of $U_{\phi}(x)$ is largely an unsolved research problem, especially in the context of classical simulation. A common kernel is the $U_{\Phi(\mathbf{x})} = Z_{\Phi(\mathbf{x})} H^{\otimes n} Z_{\Phi(\mathbf{x})} H^{\otimes n}$, where $H$ is the Hadamard gate, and $Z$ is a diagonal unitary in the Pauli-Z basis \cite{havlivcek2019supervised}. However, this kernel has been known to be hard to implement classically. There may be opportunity to explore quantum kernels that are more amenable to classical settings. 

The QSVM has been explored in multiple simulated application scenarios. Sim{\~o}es et al. \cite{simoes2023experimental} demonstrated its superior accuracy over classical SVMs on small datasets like Iris and Rain. Other applications in cybersecurity corroborate these results, although high computational costs and prolonged execution times are often noted \cite{payares2021quantum, masum2022quantum}.


\subsubsection{Variational Quantum Circuits}

Variational quantum circuits (VQC) use a hybrid quantum-classical approach to solve complex problems. A classical optimizer adjusts the parameters of a parameterized quantum circuit (PQC), so that the output of the quantum circuit approaches an optimal solution. VQCs serve as quantum analogues of neural networks, with the capability to encode classical data into quantum states and harness the power of quantum computing to minimize cost functions. These cost functions often represent the expectation value of some operator, such as a Hamiltonian in quantum physics or a measurement operator in machine learning tasks. 

VQCs have been devised in response to current limitations in implementing quantum algorithms on true quantum computers. By employing a hybrid quantum-classical framework, VQCs utilize classical optimization techniques to fine-tune the parameters of quantum circuits \cite{benedetti2019parameterized}. The classical optimizer guides the training process, iteratively updating the quantum circuit's parameters based on the outcomes of quantum measurements. This approach leverages the power of quantum computing while accommodating the constraints of near-term quantum devices, such as error rates and limited coherence times, thereby facilitating the development of quantum applications that would be currently infeasible with solely quantum-based methods.


\begin{figure*}[!ht]
    \centering
    \centerline{\includegraphics[width=0.85\textwidth]{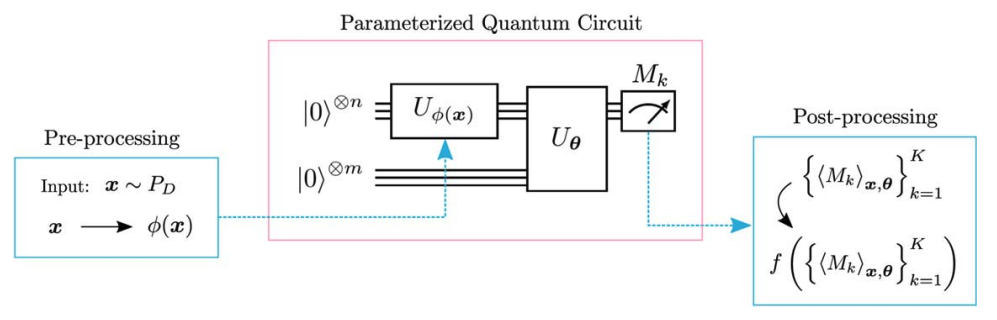}}
    \caption{Overview of the Variational Quantum Classifier (VQC) \cite{benedetti2019parameterized}. The state preparation circuit \( U_{\Phi(x)} \) encodes classical data into a quantum state using the function \( \phi(x) \). The variational circuit \( U_\theta \), parameterized by \( \theta \), acts on this prepared state and possibly on additional ancilla qubits. The framework outputs observable quantities \( \{M_k\}^K_{k=1} \), which are measured and mapped to the output space through a classical post-processing function \( f \) to predict a result. Classical methods optimize the parameters \( \theta \).}
    \label{fig:vqc}
\end{figure*}

A VQC workflow typically comprises three steps \cite{gujju2023quantum}:

\begin{itemize}
    \item \textbf{Quantum Feature Map}: Classical data $x$, is encoded into the quantum state space using a non-linear feature map ($\phi$). The encoding circuit defined in Equation \ref{eqn:encoding_circuit} can be used, where $\phi$ may be some chosen encoding method. This process can be repeated or interleaved with the variational circuit, depending on the problem at hand. 

    

    \item \textbf{Variational Circuit}: A short-depth parameterized quantum circuit, $U(\theta)$, is applied to the quantum state obtained from the feature map. This circuit consists of layers of quantum gates parameterized by $\theta$. Learning the parameters $\theta$ can be seen as an objective minimization task over some loss function $L(\theta)$ with respect to the circuit expectation values $M_k$, similar to classical machine learning routines. Thus the use of classical routines such as gradient-descent have seen success in application \cite{mitarai2018quantum, benedetti2019parameterized, schuld2020circuit}. Offloading the computation to classical machines reduces the number of quantum resources required, allowing for feasibility on NISQ hardware implementation. Analogous to neural networks, variational circuits have been shown to approximate any target function, up to arbitrary error \cite{benedetti2019parameterized}, making them viable learning models. The choice of ansatz, or circuit design, in this stage is critical and can significantly influence the performance of the VQC.

    \item \textbf{Measurement}: A measurement is performed on the final quantum state, resulting in a bit string $z \in \{0, 1\}^n$, which is then mapped to a label. By running this circuit multiple times, the probability of observing $z$ can be estimated:
    
    \begin{align}
        f(x, \theta) = \bra{\Phi(x)}U(x, \theta)^{\dagger}(\theta)M_k U(x, \theta)\ket{\Phi(x)}
    \end{align}

    This quantity is computed for each of the different classes $k \in \{1, \ldots, K\}$ using the measurement operator $M_k$ and can be interpreted as the prediction of $x$ by the circuit.
\end{itemize}
Early works have demonstrated the predictive capabilities of the VQC in classical simulation. Schuld et al. \cite{schuld2020circuit} propose a VQC that is both low-depth and highly expressive. The proposed circuit geometry uses systematically entangled gates, allowing for circuits that scale poly-logarithmically with the dataset size and representative principle components. However, the authors note that this approach limits the set of applicable datasets to those that allow for $(k, \delta)$ reductions, where $k$ scales polynomially with the dataset dimensions and $\delta$ measures classification uncertainty. Simulated tests on handwritten digit and tabular classification tasks revealed its edge over classical models like MLPs and SVMs, with fewer parameters. Comparative research for VQCs extends across domains like cybersecurity \cite{islam2022hybrid}, finance \cite{emmanoulopoulos2022quantum}, and physics \cite{terashi2021event, gianelle2022quantum}. These studies underscore the advantages of VQCs over traditional classical methods, for tasks with specific requirements, such as low qubit count and small datasets. Yet, quantum simulation resource needs grow exponentially with increased variables \cite{terashi2021event, emmanoulopoulos2022quantum}.


Blance and Spannowsky \cite{blance2021quantum} proposed a combination of quantum and classical gradient descent methods for parameter optimization. The optimization process employs a forward pass to calculate the mean squared error (MSE) loss function, followed by a backpropagation procedure to update the trainable parameters. Quantum gradient descent, based on the Fubini-Study metric \cite{stokes2020quantum}, is employed to optimize the quantum weight parameters $w$ providing an advantage over the Euclidean-based gradient descent by taking into account the geometry of the parameter space of quantum states. Vanilla gradient descent is used to optimize the classical bias term $b$. The combined optimization algorithm is given by:

\begin{equation}
\begin{split}
    \boldsymbol{\theta}_{t+1}^w & = \boldsymbol{\theta}_t^w - \eta g^+ \nabla^w L(\boldsymbol{\theta}_t),  \\
    \boldsymbol{\theta}_{t+1}^b & = \boldsymbol{\theta}_t^b - \eta \nabla^b L(\boldsymbol{\theta}_t),
\end{split}
\end{equation}

where $g^+$ is the pseudoinverse of the Fubini-Study metric. The benefits of the quantum method are seen in simulation results, where faster learning and convergence rates are observed in comparison with classical neural networks using standard steepest descent.

Various extensions to the VQC formulation have also been proposed, which aim to mirror classical neural network frameworks. These VQCs are typically characterized by their circuit architecture, choice of quantum gates, and overall compositional structure. The pipeline of feature mapping, circuit construction, and measurement persists within all these algorithms. We briefly discuss such methods in the sections below.



\subsubsection{Quantum Convolutional Neural Networks}

The quantum convolutional neural network (QCNN) employs VQCs to perform convolutional operations that mimic the functionality of their classical counterparts but in the quantum domain. The QCNN architecture consists of quantum convolution layers represented by quasilocal unitary operations, pooling layers achieved by measuring qubits and applying subsequent unitary operations, and fully connected layers, implemented by specific unitary transformations. 

Various frameworks for the QCNN have been offered, exhibiting differing approaches and techniques. In Cong et al. \cite{cong2019quantum}, the quantum convolutional layer is characterized by a single quasilocal unitary operation (denoted by \( U_i \)). Pooling is carried out by measuring some qubits and using the results to define unitary operations (denoted by \( V_j \)) on nearby qubits. When the remaining qubit count is manageable, a unitary operation \( F \) acts as a fully connected layer before the final measurement. During the training phase, the unitaries are optimized with \( O(\log(N)) \) parameters, where \( N \) is the input qubit count. Li et al. \cite{li2020quantum} introduced the Quantum Deep Convolutional Neural Network (QDCNN) that uses a layered sequence of VQCs as the convolutional filters, with a final VQC producing the classification result. The implementation, however, relies on efficient QRAM for input state preparation. Henderson et al. \cite{henderson2020quanvolutional} proposed the Quanvolutional Neural Network, a model that integrates random quantum circuits as filter layers within a traditional CNN structure for feature extraction in image classification tasks. These quanvolutional filters can be stacked by inserting a classical pooling layer in between. In particular,  Riaz et al. \cite{riaz2023accurate} empirically showed that increasing the number of quanvolutional layers enhances performance. Additionally, the use of strongly entangled quantum circuits instead of random quantum circuits as transformation layers further improved performance. Numerical simulation of these methods performed over benchmark image datasets have showed superior performance to a classical CNNs with similar structure, often with faster convergence \cite{cong2019quantum, li2020quantum, henderson2020quanvolutional, hur2022quantum}.

\subsubsection{Quantum Autoencoders}

The quantum autoencoder (QAE) aims to compress quantum data into a smaller dimension, preserving essential information while reducing the number of qubits required to describe the data. It consists of two main parts: the encoder and the decoder. The encoder maps the original data into a compressed space by applying a VQC, reducing the dimensionality of the data. The decoder reverses this process, attempting to reconstruct the original data from the compressed representation. The aim is to minimize the difference between the input and the reconstructed states. This is typically achieved by optimizing the parameters of the encoding and decoding circuits using a cost function, often the fidelity between the original and reconstructed states. Several different methods have been explored. A single unitary can be used that acts as both encoder and decoder. The unitary evolves an input state $\ket{\phi}$ to a latent state $\ket{\chi}$ using an encoder circuit $U(\theta)$, and learns to reconstruct this state using its Hermitian conjugate $\ket{\phi} = U^\dagger(\theta)\ket{\chi}$ \cite{romero2017quantum}. Alternatively, two unitaries can be learned with individual parameterizations, each acting as the encoder and decoder respectively \cite{srikumar2021clustering}. The size of the latent dimension can be fixed by discarding intermediate qubits that would feed into the decoder circuit.

\subsubsection{Quantum Generative Adversarial Networks}

Quantum Generative Adversarial Networks (QGANs) \cite{lloyd2018quantum, dallaire2018quantum} are composed of two main quantum circuits, the generator and the discriminator. The generator is a VQC controlled by a set of parameters \(\Theta_G\), and is responsible for transforming random quantum noise into quantum states that resemble the real data distribution. The discriminator is another VQC controlled by parameters \(\Theta_D\) tasked with distinguishing between real quantum states from the target distribution and the fake ones generated by the generator. In the hybrid quantum-classical setting, the discriminator is a classical neural network \cite{zeng2019learning}. During the training process, the objective is to simultaneously train the generator to produce indistinguishable states from the real data and the discriminator to efficiently differentiate between the real and generated states. The loss function is typically designed as a two-player min-max game, and the parameters are iteratively updated through gradient-based optimization methods, aiming to find the equilibrium of this adversarial game. 

\subsubsection{Quantum Circuit Born Machines}

The Quantum Circuit Born Machine (QCBM) is a type of generative model that approximates a target discrete probability distribution $q$ in the wavefunction of a quantum system \cite{benedetti2019generative}. The QCBM consists of three main components: a VQC, an objective function, and a classical optimizer. An initial \( n \)-qubit state \( \ket{0}^{\otimes n} \) is passed through the VQC \( U(\theta) \) to generate a quantum state. The Born rule is then applied to compute the probability \( p_{\theta}(x) \) of sampling a computational basis state \( x \). This is mathematically represented as \( p_{\theta}(x) = \text{tr}(\Pi_x U(\theta) (\ket{0}\bra{0})^{\otimes n} U(\theta)^{\dagger}) \), where \( \Pi_x = \ket{x}\bra{x} \) is the projection operator. The KL-divergence \( D_{KL}[q||p_{\theta}] \) measures the discrepancy between \( q \) and \( p_{\theta} \), and the Particle Swarm Optimizer (PSO) algorithm adjusts the parameters \( \theta \) to minimize this divergence \cite{tian2023recent}. Other objective functions, such as the maximum mean discrepancy (MMD) and the Sinkhorn divergence (SHD), have also been used. Studies have suggested that MMD can be more efficiently applied in large-scale systems, as KL divergence is often inaccessible. Coyle et al. \cite{coyle2020born} reported that SHD is superior to MMD in terms of accuracy and convergence rate, supported by numerical simulations. Besides PSO, various other optimization algorithms have been implemented. Gradient-free methods, such as Covariance Matrix Adaptation Evolution Strategy (CMA-ES) and Bayesian Optimization, are prevalent for optimizing non-linear and non-convex objective functions. In contrast, for large-scale parameter optimization, gradient-based algorithms like the Adam optimizer have shown successful application \cite{tian2023recent}. 

Since direct access to the quantum wavefunction is not available, the QCBM is a implicit generative model, where sampling from $q$ is easy, but characterizing it is difficult. Due to this, large QCBM circuits may be computationally intractable. Coyle et al. \cite{coyle2020born} prove a hardness result for their Quantum Circuit Ising Born Machine (QCIBM) that utilizes a Hamiltonian-informed ansatz; many circuits trained in the QCIBM model are proven to resist efficient classical simulation up to multiplicative error.

\subsubsection{Tensor Network-based VQCs}

As initially discussed in Section \ref{sec:tensor_networks}, tensor networks serve as effective ansatz for the variational circuits in VQCs. Such implementations are particularly advantageous for near-term quantum hardware, as they often require a reduced number of physical qubits, scaling logarithmically or even remaining constant with data size \cite{gujju2023quantum}. Huggins et al. \cite{huggins2019towards} outline the construction of TTN and MPS decompositions using qubit lines connected in a tree-like structure. In these models, tensor nodes correspond to multi-qubit gates, with incoming and outgoing qubits representing the bonds. Isometric tensor network nodes are converted into unitary quantum gates, and bond dimensions are defined by the number of transferred qubits between such gates \cite{rieser2023tensor}. VQCs based on MERA architectures have also been proposed \cite{cong2019quantum, huang2021variational, araz2022classical}. In classical simulation, left-over outgoing qubits are discarded or traced out after the application of each unitary. A qubit-efficient approach has been proposed that instead reinitializes the discarded qubits to the $\ket{0}$ state and are then used as inputs into subsequent unitaries \cite{huggins2019towards}. This allows for a reduced total qubit count, which is now constant for MPS, determined by the input and bond dimension, and logarithmic in the input size for TTN. Circuit cutting techniques also allow for fewer qubit usage, which partition large quantum circuits into smaller segments executable on limited-qubit hardware \cite{guala2023practical}. These smaller segments are then classically post-processed to combine their results. This approach facilitates not only the execution of large tensor-network quantum circuits on resource-constrained quantum devices, but also simplifies the classical simulation of a broader spectrum of tensor-network-based VQCs. Numerical simulations have confirmed the effectiveness of circuit cutting techniques in achieving scalable and efficient simulations for MPS-based VQCs. \cite{guala2023practical}. Empirical results have also suggested that tensor network-based VQCs outperform common VQC models and classical generative models such as GANs in terms of expressivity for certain generative tasks \cite{haghshenas2022variational, gili2022evaluating}. Additionally, they may require less training data and computational resources to achieve similar performance to classical TN models \cite{araz2022classical}.

Tensor networks have also been used as feature extractors to prepare inputs for VQCs. MPS and TTNs have been used to produce low dimensional feature vectors from input data, which can subsequently be fed into a VQC for classification \cite{chen2020hybrid, araz2022classical}. The parameters of the feature extracting TN can also be learned alongside the VQC in an end-to-end fashion. In Chen et al. \cite{chen2020hybrid}, the MPS showed stronger representational power then other dimensionality reduction methods such as PCA. In Araz and Spannowsky \cite{araz2022classical}, the hybrid architecture showed better predictive performance than standalone TN classifiers.

\subsubsection{Quantum Natural Language Modeling}

Quantum natural language modeling has scarcely left the theoretical realm, with descriptions of prospective frameworks being offered in the literature. Research has focused on the distributional-compositional-categorical model of meaning (DisCoCat) \cite{coecke2010mathematical}, which combines linguistic meaning and structure into a single model via via tensor product composition. Subsequent works show that when semantic representations are modeled in this way can be interpreted in terms of quantum processes, over which quantum computation can readily handle the resulting high dimensional tensor product spaces \cite{zeng2016quantum, coecke2020foundations, meichanetzidis2020quantum}.

Various ansatze and encoding schemes have been suggested for NLP computation over quantum hardware; a few works have implemented these ideas on classical simulators. Kartsaklis et al. \cite{kartsaklis2021lambeq} developed the lambeq Python library that allows for the conversion of sentences into quantum circuits, providing the tools for implementing experimental quantum NLP pipelines, following the methodology in \cite{meichanetzidis2020quantum} who first present a quantum pipeline for the DisCoCat methodology. In brief, the pipeline initiates with the generation of a syntax tree from a sentence using a statistical Combinatory Categorial Grammar (CCG) parser, delineating the sentence's grammatical structure. This tree is then translated into a string diagram, refined using rewriting rules to streamline the computation process, potentially omitting redundant word interactions. Finally, the adjusted diagram is transformed into a concrete quantum circuit or tensor network ansatz, trained via standard ML optimization backends such as PyTorch and JAX. Experiments using the Qiskit Aer cloud quantum simulator were performed using a simple binary meaning classification dataset of 130 sentences created using a simple context-free grammar. When compared against a classical pipeline, where the sentences are encoded as tensor networks, similar testing accuracies were achiever, albeit with fluctuation and instability at the early stages of training. These results were corroborated by Lorenz et al. \cite{lorenz2023qnlp}, where practical simulations of the DisCoCat compositional model were compared against quantum-friendly versions of the word-sequence and bag-of-words models, the latter methods being represented by simple tensor compositions of semantic bases. The compositional model showed superior results on both classical and quantum hardware.


Li et al. \cite{li2022quantum} propose a Quantum Self-Attention Neural Network (QSANN) for text classification, noting that the DisCoCat compositional model requires heavy syntactic preprocessing and a syntax-dependent network architecture, limiting its scalability to larger datasets. 
The self-attention mechanism that has seen large success in classical NLP is introduced into the quantum setting; the key component vectors of classical self-attention: queries, keys and values are modeled and trained using quantum ansatze, with an additional projection onto 1D space and Gaussian function applied to handle long distance correlations induced by inner-product
self-attention. Numerical results against both classical self-attention neural networks and the DisCoCat model \cite{meichanetzidis2020quantum}, and showed superior results to both in terms of predictive performance while requiring fewer parameters.

\subsection{Other QiML Methods}\label{sec:other}

Works in this section do not necessarily fall into the aforementioned categories, but are more intrinsic to early definitions of QiML --- methods that incorporate and adapt quantum phenomena in classical settings.

\subsubsection{Quantum-Inspired Nearest Mean Classifiers}

A line of work in QiML research, commenced by Sergioli et al. \cite{sergioli2018quantum} has explored a so-called “new approach” to QiML \cite{sergioli2020quantum} which explores supervised binary classification using quantum concepts. In \cite{sergioli2018quantum}, the authors developed a quantum-inspired version of the nearest mean classifier (QNMC). The NMC problem finds an average ‘centroid’ $u_i$ for each class $C_i$ in the training dataset:

\begin{align}
    u_i = \frac{1}{n_i} \sum_{x_i \in C_i} x_i,
\end{align}
where $n_i$ is the number of data points belonging to the class $C_i$. New instances are assigned labels based on proximity to these centroids. The normalized trace is used as the distance metric due to its ability to preserve the order of distances between arbitrary density patterns. The introduction of the quantum-inspired NMC (QNMC) stems from the observation that any real, two-feature pattern $x$ corresponds exactly with some quantum pure density operator $\rho$, obtained by the stereographic projection of $x$ onto the Bloch Sphere representation. From these encodings, the quantum centroid is defined as 

\begin{align}
    \rho_{QC} = \frac{1}{n} \sum^n_{i=1} \rho_i.
\end{align}
The authors show that the QNMC potentially outperforms the classical NMC on synthetic, non-linear data, particularly on datasets with high data dispersion or mixed class distributions. Improvements to this model have come from subsequent works. In \cite{sergioli2017quantum}, the correspondence between real and quantum objects is extended to an arbitrary $n$ feature patterns, allowing for experimentation on more complex, real-world datasets. The QNMC again shows considerably enhanced performance over classical NMC. In \cite{sergioli2019new}, Helstrom's distance is used instead, producing a model coined as the Helstrom Quantum Classifier (HQC). This allowed for the use of multiple copies of quantum states, bolstering their informational content, leading to empirically enhanced performance over benchmark tasks. This model was extended to the multi-class context in \cite{giuntini2023quantum} by leveraging the \textit{pretty-good measurement} (PGM) measurement technique from quantum state discrimination; 
a minimum-error discrimination that discerns between multiple unknown quantum states with high success probabilities. These models have also seen success in practical implementation over biomedical contexts \cite{sergioli2018quantumbiomedical, sergioli2021quantum}. Leporini and Pastorello \cite{leporini2022efficient} consider a geometric construction of the classifier, and discusses a method to encode real feature vectors into the amplitudes of pure quantum states using Bloch vectors. Bloch vectors represent the density operators, and the centroids of data classes are directly calculated based on these vectors. The obtained Bloch vector is rescaled into a real sphere to identify the centroid as a proper density operator, since the mean of a set of Bloch vectors is not typically a Bloch vector. This representation allows for data compression by eliminating null and repeated components, allows for the implementation of feature maps while saving space and time resources without compromising performance. This method showed similar, and sometimes improved accuracies over benchmark datasets, when compared with similar proposed classifiers. Bertini et al. \cite{bertini2023quantum} later propose a KNN version of the Bloch vector-based classifier by executing the method over a local neighborhood of training data points.

\subsubsection{Density Matrix-Based Feature Representation}

A central source of inspiration for QiML is derived from the probabilistic interpretation of quantum mechanics, known as the quantum probabilistic framework. Unlike classical probability, quantum probability encompasses complex-valued probability amplitudes and allows for phenomena such as superposition and entanglement. 


The quantum probabilistic framework introduces mathematical structures that can be applied to classical machine learning, specifically through the use of density matrices and Hilbert spaces. A density matrix, represented as \(\rho\), describes the statistical state of a quantum system, allowing for a mixture of pure states, and can be expressed as: 

\begin{align}
    \rho = \sum_{i} p_i \ket{\psi_i}\bra{\psi_i},
\end{align}
where \(\ket{\psi_i}\) are the pure states of the system, and \(p_i\) are the classical probabilities for each state.

In the context of machine learning, this framework can be used to represent complex relationships and dependencies within data. The density matrix can reflect the covariance among different embedding dimensions, representing how scattered words are in the embedded space \cite{zhang2018end}. Quantum entanglement can also be leveraged to describe intricate correlations between features, providing a more expressive model \cite{li2021quantumvideo}. 

The use of density matrices is prevalent in quantum-inspired NLP methods, used to capture probabilities and semantic sub-spaces of the individual words in the sentence \cite{zhang2018quantum, zhang2019quantum}. Once classical data is encoded into quantum states, the density matrix representation of these states can be computed via several methods. One method is to compute $\rho$ directly, where a sentence or document corresponds to a mixed state represented by a density matrix of individual semantic spaces \cite{zhang2020quantum, li2021quantumvideo}:

\begin{align}
    \rho = \sum_i p_i \ket{w_i}\bra{w_i},
\end{align}
where $p_i$ is the relative importance of word $w_i$ within the sentence, satisfying $\sum_i p_i = 1$. The assigned $p_i$ can be captured in various ways, such as uniformly \cite{zhang2019quantum}, by number of occurrences \cite{zhang2020quantum}, or by a softmax function \cite{li2019cnm}. This representation extends to multi-modal settings by considering non-textual media (images, videos) as “textual” features. For instance, the SIFT algorithm \cite{lowe2004distinctive} can be used to detect and describe local features in images as vectors, which can be clustered, with cluster centers considered as “visual word” vectors \cite{zhang2018quantum} and used directly in the calculation of the density matrix to produce a compositional Hilbert Space \cite{li2021quantumvideo, li2021quantumemotion}:


\begin{align}
    \rho &= \sum_i p_i  (\ket{w^m_i} \otimes \cdots \otimes \ket{w^M_i}) (\bra{w^m_i} \otimes \cdots \otimes \bra{w^M_i}) \\
    &= \sum_i p_i (\rho^m_i \otimes \cdots \otimes \rho^M_i),
\end{align}
This method assumes the importances $p_i$ are accurately known or can be reliably computed. An alternative method is to construct projectors in semantic space from word features, from which $\rho$ can be learned \cite{zhang2018quantum, zhang2019quantum}. Each projector can be described as:

\begin{align}
    \label{eqn:other_semantic_projector}
    \Pi_i = \ket{w_i}\bra{w_i},
\end{align}
where $\Pi_i$ describes the semantic space of a normalized word vector $\ket{w_i}$. Such vectors may be obtained via various real-valued embedding schemes, such as by GloVe \cite{pennington2014glove} or BERT \cite{devlin2018bert} word embeddings. A document is then considered as a sequence of projectors, $P = \{ \Pi_1, \Pi_2, \ldots, \Pi_n\}$, where $n$ is the number of terms in the document. A randomly initialized density matrix is trained and iteratively updated based on this sequence of projectors via a globally convergent Maximum Likelihood Estimation algorithm until a convergence threshold is reached, using the objective function:

\begin{align}
    F(\rho) & \equiv \max_{\rho} \sum_{i} \log \left( \mathrm{tr} \left( \Pi_{i} \rho \right) \right), \\
    \text{s.t.} \quad \mathrm{tr}(\rho) & = 1, \\
    \rho & \geq 0.
\end{align}
The final density matrix encapsulates the semantic dependencies and distributional information of the terms in the document. In multi-modal settings, image and video feature vectors are extracted, which which projectors can be constructed via Equation \ref{eqn:other_semantic_projector} and included in the set $P$. A multi-modal fusion method inspired by quantum interference has been proposed \cite{zhang2018quantum, zhang2020quantum}. Each mode has its own classifier that takes in only input density matrices for that mode. Then, the individual inferences on a document with $M$ modes produces $M$ predictions, thus the overall sentiment of the document could be uncertain. This sentiment can be modeled as a quantum wavefunction, analogized as a combination of the modal sentiment components. For example, the combination of the sentiment of a document containing text and the image components is described by:


\begin{align}
    \psi(x_i) = \alpha \psi(x_i^{\text{text}}) + \beta \psi(x_i^{\text{image}}).
\end{align}
Thus the overall sentiment score can be represented by a probability distribution, measured as:
\begin{align}
    P(x_i) &= \alpha^2 P_t + \beta^2 P_i + I \\
    I &= 2 \alpha \beta \sqrt{P_t P_i} \cos \theta
\end{align}
such that $P(x_i) = |\psi(x_i)|^2$ describes the probability of the document's sentiment score. $P_t = |\psi(x_i^{\text{text}})|^2$ and $P_i = |\psi(x_i^{\text{image}})|^2$ are the probabilities governing the sentiment scores of the text and image respectively. $\alpha$, $\beta$ and $\theta$ are learnable parameters. The interference term $I$ reflects the degree of conflict in local decisions;  if both modalities agree in sentiment, there is a constructive interference, resulting in a strongly positive or negative sentiment. 

In question and answering tasks, the joint representation for a question-answer pair can be viewed as their multiplicative interaction \cite{zhang2018end}:

\begin{align}
    \rho_{QA} = \rho_Q \rho_A.
\end{align}
The spectral decomposition of $\rho_{QA}$ exposes the joint eigenspaces, over which the trace inner product reveals the similarity between the question and answer. From this, \cite{zhang2018end} proposed two methods of constructing the feature set for each document. The first uses the trace and diagonal elements of $\rho_{QA}$: $[\text{tr}(\rho_Q \rho_A); \text{diag}(\rho_Q \rho_A)]$, where the former captures the semantic overlaps between the question-answer pair, and the latter accounts for the varying degrees of importance for similarity measurement. The second uses 2D convolutions to scan the density matrices, resulting in feature maps which are processed using row-wise and column-wise max-pooling to generate feature vectors, aiming to capture a more nuanced and complex understanding of similarity between question and answer pairs. 

\textbf{Learning and Classification:} Several classical ML techniques have been used to learn a performant parameterization over the density matrix features, including SVMs and Random Forests \cite{zhang2019quantum}, as well as deep learning structures such as RNNs \cite{zhang2020quantum, shi2021two}, and CNNs \cite{zhang2018end}. Back propagation is often used as the optimization method.


In addition, quantum measurement-based procedures can also be used to extract predictions. Once the set of states \(\{ \rho_c^t \}\) is obtained that represent the data, a global observable \( O \) is introduced, uniquely represented by a set of eigenvalues \(\lambda\) and corresponding eigenstates \(\ket{e}\), expressed as \( O = \sum_i \lambda_i \ket{e_i}\bra{e_i} \). These eigenvalues and eigenstates correspond to some outcome representation of interest, such as sentiment-related aspects \cite{li2021quantumvideo} or emotional states \cite{li2021quantumemotion}. The measurement process leads to a collapse of the state onto one of the eigenstates, and a probability distribution over the eigenstates is calculated as \( p_i = \bra{e_i} \rho_c^t \ket{e_i} \), where \(\rho_c^t\) is the state at time \( t \). This probability distribution can then be used to perform predictions over the data.

\textbf{Complex-Valued Density Matrix Features:} Recently, researchers have explored the effect of using complex-valued word embeddings in tandem with considering local word correlations, noting in using only real-valued vectors, the full probabilistic properties of a density matrices and their complex formulations are ignored \cite{li2019cnm}.

The formulation of words via Equation \ref{eqn:tn_semantic_basis} is preserved, except with the inclusion of complex components. This complex-valued approach is inspired by the representation of words as a superposition of semantic units, where each word \( |w\rangle \) is defined as a unit-length vector on \( H \):

\begin{align}
    \label{eqn:tn_semantic_basis_complex}
    |w\rangle = \sum_{j=1}^{n} r_je^{i\phi_j}|e_j\rangle,
\end{align}
where $i$ is the imaginary unit, and the $r_j$ and $\phi_j$ are non-negative real-valued amplitudes and corresponding complex phases, satisfying $\sum_{j=1}^{n} r_j^2=1$, and $\phi_j \in [-\pi, \pi]$ respectively. The inclusion of complex components in the word embeddings allows for a richer representation of semantic information by leveraging both the magnitude and phase of complex numbers. The magnitudes $r_j$ describe the importance of each semantic base in the composition of the word, while the phases $\phi_j$ capture the subtle interrelations between different semantic units. This leads to a more expressive and nuanced modeling of word semantics \cite{li2019cnm}.

Complex-valued representations have also been used over multi-modal tasks. In emotion recognition, Li et al. \cite{li2021quantumemotion} consider each utterance is as a mixture of unimodal states whose features are recast as pure states creating a multimodal mixed state representation. Then, a procedure inspired by quantum evolution is employed to track the dynamics of emotional states in a conversation. A quantum-like recurrent neural network tracks the evolving emotional states during a conversation, considering the uncertainties in the conversational context and efficiently memorizing the context information due to unitary transformation, which ensures zero information loss. The "measurement and collapse" phase introduces a global observable to measure the emotional state of each utterance, calculating a probability distribution that corresponds to the likelihood of the state collapsing onto specific eigenstates. The result is then mapped to emotion labels using a neural network with a single hidden layer.

Shi et al. \cite{shi2021two} similarly propose complex-valued word embeddings which likens words to quantum particles existing in multiple states, representing polysemy. The method corresponds a word with multiple meanings to a quantum particle that can exist in several states. Additionally, sentences are likened to quantum systems where these particles (or words) interact or interfere with each other, just as quantum particles can interact in a quantum system. Complex-valued word embeddings can be formed from amplitude word vectors and phase vectors, capturing rich semantic and positional information with greater alignment with quantum concepts. These embeddings are used in text classification models utilizing gated recurrent units (GRUs) and self-attentive layers to extract more semantic features. An extended model is also presented that applies a convolutional layer on the projected word embeddings matrix to capture local textual features. This is inspired by the quantum theory concept of ‘entanglement’, where the state of one particle is connected to the state of another, no matter the distance between them. In a similar vein, the convolutional layer captures dependencies between different parts of the text, or ‘local features’, that might otherwise be missed.

This quantum probabilistic formulation of density matrices not only serves as an effective representation for sentences or documents in NLP tasks but also offers a versatile framework that may extend to other contexts where data can be captured as probabilistic events.

\subsubsection{Quantum Formalisms Applied to Neural Networks}

Exploring neural network representations through the lens of quantum mechanics has been a long explored topic. Such methods aim to improve the robustness of classical neural networks by formulating quantum-based activation operators or utilizing quantum feature spaces \cite{altaisky2001quantum, zhou2006quantum, schuld2014quest}. In this subsection we present a few recent, practical works in this area.

Patel et al. \cite{patel2019novel} introduce the Quantum-inspired Fuzzy based Neural Network (Q-FNN), a three-layer neural network that employs Fuzzy c-Means (FCM) clustering to fine-tune connection weights and decide the number of neurons in the hidden layer. The fuzziness parameter $m$, which manages the overlap among samples from different classes, takes on a qubit representation, which enlarges the search space for the selection of an appropriate fuzziness parameter. The final cluster centroids, found after numerous iterations of fuzzy clustering, serve as the final connection weights of the hidden layer. This model has been proven effective in dealing with two-class classification problems.

Sagheer et al. \cite{sagheer2019novel} replace the classical perceptron within the neural network model with a quantum-inspired version: the autonomous perceptron model (APM). In the APM, a feature vector $x \in \mathbb{R}^n$ is replaced by a quantum state vector $\ket{\psi} \in \mathbb{C}^n$ which can be represented as a complex linear combination of the basis vectors in the $n$-dimensional complex vector space. Instead of real-valued weights used in a classical perceptron, quantum weights are introduced as normalized complex numbers $\omega_i = \exp(i\theta_i)$ where $\theta_i$ is the phase of the $i$-th weight. The activation function is replaced by a measurement operation, which projects the state of the system onto one of the basis vectors. The output of the APM is the expectation value of this measurement, which can be calculated as $\langle\psi| M |\psi\rangle$, where $M$ is the Pauli-Z measurement operator in the computational basis. The measurement outcome is then compared with a threshold value to make binary decisions, akin to classical perceptrons. Results over the UC Irvine (UCI) Machine Learning Repository benchmark classification datasets \cite{UCI} showed the APM-based model outperformed 15 standard classifiers models when subjected to the same experimental conditions.

Konar et al. \cite{konar2020quantum} developed the quantum-inspired self-supervised network (QIS-Net) architecture for the automatic segmentation of brain magnetic resonance (MR) images. The model is composed of layers of classically implemented quantum neurons arranged. Each neuron is depicted as a qubit using matrix notation. The intra-connection weights among neurons within the same layer are set to $\pi/2$, to emulate a quantum state. The input layer deals with qubit information from connected neighborhood subsets of each seed neuron, which is then accumulated at the central neuron of the intermediate layer through interconnections. Image data by feeding image pixels to the input layer as quantum bits, which then propagate to the intermediate and output layers, which are updated through rotation gates determined by the relative quantum fuzzy measures of pixel intensity at the constituent quantum neurons between layers. A novel quantum-inspired multi-level sigmoidal (QMSig) activation function is integrated into the QIS-Net model to handle the complexity of multi-intensity grayscale values in brain MR images.

\begin{align}
    \text{QMSig} = \frac{1}{\lambda\omega + e^{-\nu(x-\eta)}}.
\end{align}
The function adjusts its activation based on qubits, improving the model's accuracy in segmenting complex images. When tested on Dynamic Susceptibility Contrast (DSC) brain MR images for tumor detection, QIS-Net demonstrated superior performance compared to classical self-supervised network models commonly used in MR image segmentation, whilst requiring less computational overhead with respect to time and resources.


Zhang et al. \cite{zhang2022quantum} introduce a method that leverages quantum entanglement to calculate joint probabilities between features and labels. The maximally entangled Bell state system of two qubits $\ket{\Phi^+}$ is defined, where one qubit is described as the feature and the other as the label. The observables and measurement operators of the entangled system are defined with specific spectral decompositions. The positive and negative measurement operators for the entangled system consists of the $n$-th attribute and the label, and are given by \(\mathcal{M}^{\pm}_{n}(\theta_n,\phi_n) = P^+_n(\theta_n,\phi_n) \otimes P^\pm_{l}\), where polar and azimuth angles $\theta_n$ and $\phi_n$, are the arbitrary real parameters. By applying these operators to the entangled system, probability values for both positive and negative examples are obtained: \(p^\pm_n(\theta_n,\phi_n) = \bra{\Phi^+} \mathcal{M}^\pm_n(\theta_n,\phi_n)\ket{\Phi^+}\). This formalism enables the calculation of quantum joint probabilities, and is integrated into a classical MLP by replacing hidden layer neurons in the MLP with the measurement process. Model optimization uses the cross-entropy loss function with the Adam optimizer to ensure smooth parameter changes. 

\subsubsection{Miscellaneous Quantum Mechanics-Based Classifiers} 

Various other classification methods have also been proposed that incorporate quantum phenomena into their classical learning model. In the following works, the main concepts of encoding classical data into some quantum state representation, and discrimination of those states via a measurement process are highlighted. 

Tiwari and Melucci \cite{tiwari2019towards} explored the prospect of classification inspired by quantum signal detection theory, which aims to decide between two different hypotheses --- the presence or absence of a signal. A codification process converts the signal into a particle state, which is then measured, much like the classical signal detection framework. When considered in a classification context, the two hypotheses subjected to decision become two class labels, represented by distinct density operators derived from data features, and characterize the system state associated with each class. Outcomes are decided via projections corresponding to the density operators. Using this knowledge, the authors devise a binary classifier over vectorized documents based on frequency of distinct elements. Density operators for each class are estimated using training samples:

\begin{align}
    \rho_c = \frac{\ket{v_c}\bra{v_c}}{\text{tr}(\ket{v_c}\bra{v_c})},
\end{align}
for each class $c \in \{0, 1\}$, from which an optimal projection operator can be calculated:

\begin{align}
    \Lambda = \sum_{l:e_l \geq 0} \ket{e_l}\bra{e_l}
\end{align}
from eigenstates $e_l$ of the operator $\rho_1 - \lambda \rho_0$. A test sample is then classified based on the result of a projection operation involving the sample's feature vector and the projection operator. If the result is greater or equal to 0.5, the sample is assigned to the class, otherwise, it is not. The model has been tested over image and textual datasets \cite{tiwari2018towards}, showing superior performance in recall, and comparable precision and F-measures across varying feature ranges compared to baseline models.

Zhang et al. \cite{zhang2021interactive} proposed a novel method for data classification using principles of quantum open system theory, termed the Interaction-based Quantum Classifier (IQC) that models the classification process as a quantum system's evolution. The interaction between the target system (qubit) and the environment (input data) is characterized by the Hamiltonian \( H_{\text{int}} = -\tilde{g}\sigma_Q \otimes \sigma_E \), where $\tilde{g}$ is the coupling constant whose magnitude reflects the strength of the interaction, leading to the unitary evolution \( U(\tau) = e^{i\sigma_Q \otimes \sigma_E(\tau)} \), where \( \sigma_Q = \sigma_x + \sigma_y + \sigma_z \). Both systems are initialized as equal probability superpositions, and the composite system evolves according to the unitary operator. Measurement of the evolved state determines probabilities used for classification. The two-category classification task is defined by a unitary operator involving input and weight vectors, \( U(\mathbf{x}_i) = e^{i\sigma_Q \otimes \sigma_E(\mathbf{x}_i)} \), and a gradient descent update rule for the weights, \( w_i = w_i - \eta(z_i - y_i)(1 - p_1^2)x_i \), where $p_1^2$ is the positive ground state probability, is applied to optimize the classification.


\section{QiML in Practice}\label{sec:discussion}

In this section, we delve into the practical applications of QiML, showcasing where these techniques have been employed and evaluated empirically. We present an exploration of several sectors, including medical, financial, physics, and more, detailing how QiML has been utilized, and discuss these in terms of the QiML methods presented (dequantized algorithms, TNs, QVAS, and others). To aid this discussion, we present a compilation of relevant works in Table \ref{tab:qiml_practical_applications}, offering a quick reference to the practical applications of QiML. Each perspective is treated as a subsection, allowing readers the flexibility to navigate the section that resonates with their domain-specific interest.





\begin{table*}[!ht]
    \centering
    \small
    \caption{Practical Application Domains of QiML: A summary of works that report experimental results and their availability of source code (* indicates source code availability).}
    \resizebox{0.93\textwidth}{!}{%
    \begin{tabular}{|l|c|c|c|c|}
        \hline
        &  \begin{tabular}[x]{@{}c@{}}\textbf{Dequantized}\\[-1em]\textbf{Algorithms}\end{tabular} & \textbf{TNs} & \textbf{QVAS} & \textbf{Other QiML Methods} \\
        \hline

        \multicolumn{5}{|l|}{\textbf{Image Modeling}} \\
        \hline
    
        Classification & - & \begin{tabular}[x]{@{}c@{}} \cite{stoudenmire2017supervised}*, \cite{liu2019machine}*, \cite{stoudenmire2018learning}, \cite{efthymiou2019tensornetwork}*, \\[-1em] \cite{sun2020generative}*, \cite{glasser2020probabilistic}, \cite{cheng2021supervised} \end{tabular} & 

        \begin{tabular}[x]{@{}c@{}}
        \cite{schuld2020circuit}, \cite{li2020quantum}, \cite{henderson2020quanvolutional}, \cite{chen2020hybrid}, \\[-1em] \cite{riaz2023accurate} \end{tabular}

        & \cite{tiwari2019towards}, \cite{tiwari2019binary} \\
        Generative Modeling & - & \cite{han2018unsupervised}, \cite{cheng2019tree}, \cite{vieijra2022generative}, \cite{liu2023tensor}* & \cite{rudolph2022generation},  \cite{tsang2022hybrid}*, \cite{zhou2023hybrid} & - \\

        \hline
        \multicolumn{5}{|l|}{\textbf{Natural Language Processing}} \\
        \hline
        Language Modeling & - & \cite{zhang2019generalized}, \cite{miller2021tensor}* & - & - \\
        Sentiment Analysis & - & - & - & \begin{tabular}[x]{@{}c@{}}\cite{zhang2018quantum}, \cite{zhang2019quantum}, \cite{li2021quantumvideo}, \cite{zhang2020quantum},\\[-1em]\end{tabular} \\
        Question-Answering & - & \cite{zhang2018quantumqmwf} & - & \cite{zhang2018end}, \cite{li2019cnm} \\
        Text Classification & - & - & \cite{kartsaklis2021lambeq}*, \cite{li2022quantum}, \cite{lorenz2023qnlp}* & \cite{tiwari2018towards}, \cite{shi2021two} \\
        Emotion Recognition  & - & - & - & \cite{li2021quantumemotion}* \\

        \hline
        \multicolumn{5}{|l|}{\textbf{Medical}} \\
        \hline
        Disease Classification & - & \cite{selvan2020tensor}*, \cite{selvan2020locally}* & \cite{pomarico2021proposal}, \cite{azevedo2022quantum}, \cite{esposito2022quantum} & \cite{sergioli2018quantumbiomedical}, \cite{sergioli2021quantum}* \\
        Image Segmentation & - & \cite{konar20233}* & - & \cite{konar2020quantum} \\
        \hline

        \hline
        \multicolumn{5}{|l|}{\textbf{Finance}} \\
        \hline
        Options Pricing & - & \cite{patel2022quantum} & \cite{sakuma2020application}*, \cite{ganguly2023implementing} & - \\
        Portfolio Optimization & \cite{arrazola2019quantum}* & \cite{mugel2022dynamic} & \cite{alcazar2020classical} & - \\
        Time-Series Forecasting & - & - & \cite{emmanoulopoulos2022quantum} & - \\
        Synthetic Data Generation & - & - & \cite{coyle2021quantum}, \cite{kondratyev2021non} & - \\

        \hline
        \multicolumn{5}{|l|}{\textbf{Physics}} \\
        \hline
        \begin{tabular}[x]{@{}l@{}}Event Classification\\[-1em] \& Reconstruction\end{tabular} & - & \cite{araz2021quantum}* & \begin{tabular}[x]{@{}c@{}} \cite{terashi2021event}, \cite{blance2021quantum}, \cite{wu2021applicationvqc}, \cite{wu2021applicationqsvm}, \\[-1em] \cite{gianelle2022quantum}, \cite{ngairangbam2022anomaly}, \cite{araz2022classical}*
        \end{tabular} & - \\

        \hline
        \multicolumn{5}{|l|}{\textbf{Chemistry}} \\
        \hline
        Molecule Discovery & - & \cite{moussa2023application} & - & - \\
        \hline

        \hline
        \multicolumn{5}{|l|}{\textbf{Cybersecurity}} \\
        \hline
        Attack Detection & - & - & \cite{islam2022hybrid}, \cite{suryotrisongko2022evaluating}, \cite{payares2021quantum}*, \cite{masum2022quantum} & - \\
        Intrusion Detection & - & - & \cite{gong2022network} & - \\
        Fraud Detection & - & - & \cite{herr2021anomaly} & - \\
        \begin{tabular}[x]{@{}l@{}}Automatic Speech\\[-1em]Recognition\end{tabular} & - & - & \cite{yang2021decentralizing}* & -  \\

        \hline
        \multicolumn{5}{|l|}{\textbf{Other Tasks and Applications}} \\
        \hline
        
        Generic Classification & \cite{arrazola2019quantum}*, \cite{ding2021quantum}*, \cite{chepurko2022quantum} & \cite{novikov2018exponential}* & \cite{schuld2020circuit}, \cite{simoes2023experimental} & \begin{tabular}[x]{@{}c@{}}  \cite{sergioli2018quantum}, \cite{sergioli2017quantum}, \cite{sergioli2019new}, \cite{giuntini2023quantum}, \\[-1em] \cite{leporini2022efficient}*,  \cite{bertini2023quantum}*, \cite{zhang2021interactive}, \cite{zhang2022quantum}, \\[-1em] \cite{patel2019novel}, \cite{sagheer2019novel} \end{tabular} \\
        
        Recommendation Systems & \cite{arrazola2019quantum}* \cite{chepurko2022quantum} & \cite{novikov2018exponential}* & - & - \\
        
        Bit-string Classification & - & \cite{stokes2019probabilistic}*, \cite{bradley2020modeling}*& - & - \\
        
        Generic PDE Solvers & - & \cite{garcia2021quantum}* & - & - \\
        Generic Anomaly Detection & - & \cite{wang2020anomaly} & - & - \\
        
        \hline

        \hline
    \end{tabular}
    }
    \label{tab:qiml_practical_applications}
\end{table*}

\subsection{Image Modeling}

Both image classification and generative image modeling have seen a wealth of implementation using tensor network methods. Particularly, the MNIST and Fashion-MNIST datasets have provided a relatively simple, low dimensional test bed for the development of TN methodologies. Tables \ref{tab:tn_sup_performance} and \ref{tab:tn_unsup_performance} outline these numerical results and presents key improvements in TN performance over the benchmark datasets. The current, classical, state-of-the-art benchmark for the given task is also included, indicated by the asterisk (*). The study conducted by Han et al. \cite{han2018unsupervised} is not featured in Table \ref{tab:tn_unsup_performance} due to the lack of experimental results from training on the full MNIST dataset. In general, TN learning models have shown competitive, but not superior performance when compared to classical benchmarks.

\begin{table}[!ht]
    \centering
    \small
    \caption{Supervised Tensor Network Performance, compared with Classical Benchmarks (the classical benchmark is indicated by *)}
    \begin{tabular}{l|c|c|c}
       \textbf{Task} & \textbf{Method} & \begin{tabular}[x]{@{}c@{}}\textbf{Test}\\[-1em]\textbf{Acc.}\end{tabular}& \textbf{Optimization} \\
       \hline
        MNIST & MPS \cite{stoudenmire2017supervised} & 99.03\% & DMRG \\
        & MPS + TTN \cite{stoudenmire2018learning} & 98.11\%  & DMRG \\
        & TTN \cite{liu2019machine} & 95\%  & MERA \\
        & MPS \cite{efthymiou2019tensornetwork} & 98\%  & SGD+Adam \\
        & GMPSC \cite{sun2020generative} & 98.2\% & SGD+Adam \\
        & PEPS \cite{cheng2021supervised} & 99.31\%  & SGD+Adam \\
        & \begin{tabular}[x]{@{}c@{}}CNN-\\[-1em]Snake-SBS\end{tabular} \cite{glasser2020probabilistic} & ~99\% & SGD \\
        & Ensemble CNN* \cite{an2020ensemble} & 99.91\% & - \\
        \hline
        
       \begin{tabular}[x]{@{}c@{}}Fashion-\\[-1em]MNIST\end{tabular} & MPS \cite{efthymiou2019tensornetwork} & 88\%  & GD \\
       & MPS + TTN \cite{stoudenmire2018learning} & 88.97\%  & DMRG \\
       & CNN-PEPS \cite{cheng2021supervised} & 91.2\% & SGD+Adam \\
       & \begin{tabular}[x]{@{}c@{}}CNN-\\[-1em]Snake-SBS\end{tabular} \cite{glasser2020probabilistic} & 92.3\% & SGD \\
       & \begin{tabular}[x]{@{}c@{}}Fine-Tuned\\[-1em]DARTS*\end{tabular} \cite{tanveer2021fine} & 96.91\%  & - \\
       \hline

    \end{tabular}
    \label{tab:tn_sup_performance}
\end{table}

\begin{table}[!ht]
    \centering
    \small
    \caption{Unsupervised Tensor Network Performance, compared with Classical Benchmarks (the current classical benchmark is indicated by *)}
    \begin{tabular}{l|c|c|c}
    \textbf{Task} & \textbf{Method} & \begin{tabular}[x]{@{}c@{}}\textbf{Test}\\[-1em]\textbf{NLL}\end{tabular}& \textbf{Optimization} \\
       \hline
       \begin{tabular}[x]{@{}c@{}}Binarized-\\[-1em]MNIST\end{tabular} & MPS \cite{cheng2019tree} & 101.5  & DMRG \\
       
       & TTN 1D \cite{cheng2019tree} & 96.9  & DMRG \\
       & TTN 2D \cite{cheng2019tree} & 94.3  & DMRG \\
       & PEPS (D = 4) \cite{vieijra2022generative} & 91.2  & SGD+Adam \\
       & AMPS \cite{liu2023tensor} & 84.1  & GD \\
       & Deep-AMPS \cite{liu2023tensor} & 81.8  & GD \\
       & CR-NVAE* \cite{sinha2021consistency} & 76.93 & - \\
       \hline
    \end{tabular}
    \label{tab:tn_unsup_performance}
\end{table}

Similarly, variational quantum algorithms have used image datasets to test various introduced methods \cite{schuld2020circuit, li2020quantum, henderson2020quanvolutional}, and explore their  capabilities \cite{riaz2023accurate}. In particular, QDCNN \cite{li2020quantum} and Quanvolutional Neural Networks \cite{henderson2020quanvolutional} have been introduced that incorporate quantum filters and operations that mirror classical convolution techniques, and have shown enhanced performance over classical methods with comparable architectures. Rudolph et al. \cite{rudolph2022generation} propose a QCBM to learn and sample the prior distribution of a classical GAN, extending its capabilities with quantum samples from multiple measurement bases. The method was shown to enhance the expressivity of the prior distribution, outperforming classical generative methods with as few as 8 simulated qubits. Zhou et al. \cite{zhou2023hybrid} utilize a QGANs with a quantum circuit generator and classical discriminator, and introduces a remapping method that aims to simplify the task of learning a multimodal distribution for image generation. Gray scale values of all pixels in the original image are sorted in ascending order and are then mapped back into their original pixel positions to create a new image with a unimodal distribution. This resulted in a reduction in the total number of required parameters, without sacrificing the quality of generations. Tsang et al. \cite{tsang2022hybrid} employed a “patching” strategy in their Wasserstein-QGAN implementation, which splits the output image generated into different patches, each generated by a separate quantum circuit. This allowed for a reduction in the quantum resources required.

Other methods in QiML have shown promise in image classification tasks \cite{tiwari2019towards, tiwari2019binary}. The extra flexibility provided by superposition is cited to contribute to better decision-making in these tasks \cite{tiwari2019towards}. Applications to real-world data is also seen in medical imaging \cite{sergioli2021quantum}. 

Further applications of such methods in this domain may be expected in future.

\subsection{Natural Language Processing}

Numerous natural language processing (NLP) tasks have been explored in QiML literature, including sentiment analysis, question-answering, and text classification. These methods typically extract word embeddings and project them into higher dimensional space, before producing either a tensorial representation by the summation of basis states describing individual semantic elements in the full vocabulary \cite{zhang2018quantumqmwf, zhang2019generalized} or a full density matrix feature representation \cite{zhang2018quantum, zhang2019quantum, tiwari2018towards}. Success is seen in performances over high dimensional word embeddings, with GloVe embeddings used at a 100-dimensional level to semantic sufficiency, although computation complexity concerns have been cited \cite{zhang2018quantum, zhang2019quantum, shi2021two}.
The dimensions of complex valued word embeddings has varied where used \cite{li2019cnm, li2021quantumvideo, li2021quantumemotion}, scaling with the vocabulary size. Where noted, the computational time of the QiML method is typically much longer than traditional methods \cite{zhang2018quantum, zhang2019quantum}, where the discrepancy is due to the matrix representation of samples. It is not clear which input preparation methods are best in representing natural language. Several works have utilized density matrices, or have offered physical interpretations for complex-valued embedding schemes, where different vector components are argued to encode low and high level semantic aspects \cite{li2019cnm}. Further research may inquire into the transferability or suitability of methods to different language tasks.


Concerning quantum variational methods, quantum NLP remains a largely theoretical field; advancements centered around the distributional-compositional-categorical model of meaning (DisCoCat) which integrates linguistic semantics and structure through tensor product composition \cite{coecke2020foundations}. Research indicates that when semantic interpretations are framed in this manner, quantum processes can manage the resulting high-dimensional tensor product spaces. Experimental results on this approach, using quantum simulators, align with classical tensor network outcomes, emphasizing the potential of quantum methods in NLP. Outside of the DisCoCat model, efforts to enhance classical NLP architectures using quantum components have seen the implementation of the Quantum Self-Attention Neural Network (QSANN) model \cite{li2022quantum}. In general, while quantum variational methods in NLP are still in nascent stages, there is a growing interest and understanding that they can offer computational benefits and efficiency, particularly in handling high-dimensional spaces and potentially reducing model parameters.

\subsection{Medical}\label{sec:practice_medical}

Quantum Nearest Mean Classifier models have shown some promising results \cite{sergioli2018quantumbiomedical, sergioli2021quantum} in the medical field. Models like the HQC \cite{sergioli2021quantum} have been successful, primarily due to their invariance to re-scaling, with the inclusion of the free re-scaling parameter appearing to be a key factor in their performance. In \cite{sergioli2018quantumbiomedical}, the authors emphasized the importance of incorporating qualitative features, often prevalent in biomedical contexts but neglected in their study, suggesting that more advanced modeling methods need to be investigated. The authors also underlined the criticality of identifying optimal encoding methods that can accurately represent the given dataset, a challenge that persists in both QiML and traditional ML fields.

Tensor networks have also been used in medical contexts, in binary classification over metastasis detection from histopathologic scans, detection of nodules in thoracic computed tomography (CT) scans \cite{selvan2020tensor, selvan2020locally}, and 3D magnetic resonance imaging (MRI) scans \cite{selvan2020locally}. The models showed strong area-under-the-curve (AUC) performance compared with classical baselines, while using only a fraction of the GPU memory. However when modeling 3D images, the approach seemed to require a high number of parameters when compared with CNN baselines. The same was not true for model in \cite{konar20233}, where tensor network compression was shown to give large savings in parameters. 

VQCs have been used in medical contexts over both image \cite{azevedo2022quantum} and audio-based \cite{esposito2022quantum} datasets. Azevedo et al. \cite{azevedo2022quantum} propose a transfer learning approach, where classical networks pretrained on ImageNet \cite{deng2009imagenet} are used to extract features for a quantum circuit, the DressedQuantumNet, which performs the final classification. This circuit, attached to the final linear layer of a pretrained model (Resnet18), takes 512 real values, and performs an angle encoding to construct quantum states, before being passed through to several variational layers. Notably, the quantum classifier is the only trainable part of the network, with weights updated during training via techniques like cross-entropy loss and the Adam optimizer. The model achieved an accuracy of 84\%, outperforming the classical standalone ResNet model, which achieved a maximum accuracy of 67\%.
Esposito et al. \cite{esposito2022quantum} applied the Quanvolutional Neural Network to detect COVID-19 from cough audio using DiCOVA \cite{muguli2021dicova} and COUGHVID \cite{orlandic2021coughvid} datasets. They integrated quantum circuits as quanvolutional layers into Recurrent and Convolutional Neural Networks (RNN, CNN) using the PennyLane library, with feature extraction via two- and four-qubit quantum circuits. Test accuracies for classical RNN and CNN were 79.4\% and 73.0\% respectively, while QNNs achieved 74.6\%-78.8\% with no noise. The results are thus comparable to classical methods, however the quantum simulations were observed to necessitate extended training duration. In \cite{pomarico2021proposal}, a breast cancer dataset was created from histological data for binary classification of metastatic diffusion to lymph nodes. The data was quantum-encoded using the trigonometric kernel mapping (Equation \ref{eqn:tn_trigonometric_basis}) and processed in a quantum circuit of nearest-neighbor two-qubit unitaries and Pauli rotations. CNOT gates managed qubit interactions. The quantum circuit was then converted to a tree tensor network for classical evaluation. Performance was comparable to the CancerMath prognosis tool across a variety of feature selection settings. The authors note the limitation of using the kernel, which severely limited the number of included prognostic factors.

\subsection{Finance}

Finance modeling has seen the implementation of QiML chiefly in portfolio optimization. The implementation in \cite{arrazola2019quantum} serves as a proof of concept for dequantized matrix inversion on large datasets with intrinsic large-scale matrix calculations, with the goal of identifying practical bottlenecks, rather than achieving high performance. Tensor network structures used here have shown effectiveness in optimization. In \cite{garcia2021quantum}, a classically simulated quantum register encoded via an MPS is proposed for multivariate calculus computation, capitalizing on the low entanglement between states for smooth functions with bounded derivatives. This representation enables the efficient storage of an exponential amount of weights and proves theoretically amenable to operations such as Fourier analysis, derivatives approximation, and interpolation methods. Mugel et al. \cite{mugel2022dynamic} implemented an MPS for dynamic portfolio optimization, which showed impressive performance when compared to D-Wave hybrid-quantum annealing and quantum variational circuits in terms of Sharpe ratios and ability to achieve global minimums reliably. The method, however, suffers greatly in computation time when compared with quantum methods. Patel et. al \cite{patel2022quantum} integrate MPO structures into neural network layers to reduce the number of model parameters. The authors show the consequently leading to the model size reduction and is also shown to lead to faster convergence in some cases. This network compression method was able to well approximate the original weight matrices with many fewer parameters, whilst exhibiting minimal loss in performance. However, in the presented works, the nature of the dataset used for experiments is either random, citing difficulties in scaling to large, real-world datasets \cite{mugel2022dynamic}, or not exposed or elucidated \cite{patel2022quantum}. 

Emmanoulopoulos and Dimoska \cite{emmanoulopoulos2022quantum} note that VQCs could match the performance of long short term memory (LSTM) models over time-series forecasting, even exhibiting slight superiority with high noise coefficients data due to the alignment of trigonometric functions in quantum circuits with the nature of time series signals. However, the authors acknowledged that the practical application of VQCs are currently constrained by their inability to handle large datasets.

QCBMs have also seen large application in finance, for portfolio optimization \cite{alcazar2020classical}, options pricing \cite{ganguly2023implementing}, and generating synthetic financial data \cite{coyle2021quantum, kondratyev2021non}. In many cases, numerical simulations reveals the enhanced expressivity of QCBMs over classical RBM models \cite{alcazar2020classical, coyle2021quantum, kondratyev2021non}.

\subsection{Physics}

QiML has seen wide use in high-energy physics (HEP) applications. A common task is discriminating between signal and background events in the context of the Standard Model of physics. 

Araz and Spannowsky \cite{araz2021quantum} utilized MPS tensor networks for top versus quantum chromodynamics (QCD) jet discrimination in physics modeling using a combined SGD and DMRG optimization method; applying DMRG in the first batch of each epoch and SGD thereafter. Despite slightly weaker performance compared to CNN models, the MPS method provided a higher degree of interpretability.


Variational quantum methods have also seen such use. Terashi et al. \cite{terashi2021event} applied variational quantum algorithms for signal event classification in HEP data analysis using supersymmetry. Two implementations were tested: one using RY and RZ gates with an Ising model Hamiltonian, and the other using Hadamard and RZ gates with Hadamard and CNOT for entanglement. The study compared these methods against traditional Boosted Decision Tree (BDT) and DNN algorithms, finding comparable discriminating power for small training sets (10,000 events or fewer). Simulations were run on Qulacs \cite{suzuki2021qulacs} and IBM Quantum QASM \cite{IBMQuantum} simulators. Resource demand for quantum simulation was high and increased exponentially with the number of variables used, making extended iterations impractical.


For data analysis of $t\bar{t}H$ (Higgs coupling to top quark pairs), both the quantum variational classifier \cite{wu2021applicationvqc} and the quantum kernel estimator \cite{wu2021applicationqsvm} methods were employed using IBM quantum simulators. Results on the quantum simulators using 10-20 qubits show that these quantum machine learning methods perform comparably to SVM and BDT classical algorithms, with both achieving reasonable AUC scores, indicating good classification performance. These results maintained over various quantum simulators, including Google Quantum \cite{google2023quantumai}, IBM Quantum \cite{IBMQuantum}, and Amazon Braket \cite{amazonbraket}.

In Gianelle et al. \cite{gianelle2022quantum}, VQCs were used for $\beta$-jet charge identification over Large Hadron Collider (LHC) data. Both amplitude and angle encoding schemes were assessed, alongside classical deep neural networks (DNNs). Results found DNNs to slightly outperform angle encoding VQCs, being compatible within a $2\sigma$ range, suggesting similar performance levels. Amplitude encoding VQCs consistently under-performed in comparison with angle encoding, but generally took less time to train due to being less complex in layer depth. The authors note that the number of layers is a parameter to be optimized, and show that increasing layer depth did not necessarily result in improved performance. The study also highlighted the resilience of quantum algorithms, with the angle embedding model maintaining efficacy with fewer training events, a potential advantage over classical ML methods. However, increases in model complexity and training time present challenges, with accuracy improvements saturating beyond five layers and longer training times for quantum models. The DNN also showed superior performance when a large number of features is employed.

In Ngairangbam et al. \cite{ngairangbam2022anomaly}, a quantum autoencoder (QAE) is used for the task of distinguishing signal events from background events, following an anomaly detection approach. Approximately 30,000 background and 15,000 signal events are generated. Anomaly detection then considers that the compression and subsequent reconstruction of data will work poorly on data with different characteristics to the background. Performance is compared against a classical autoencoder network (CAE); the QAE maintained higher classification performance than the CAE across a range of latent dimensions. Quantum gradient descent is used for faster convergence in optimization; the study finds that using this method allows the QAE to efficiently learn from as little as ten sample events, demonstrating the model is much less dependent on the number of training samples. This suggests that QAEs show better learning capabilities from small data samples compared to CAEs, particularly relevant to LHC searches where the background cross section is small. The authors hypothesize this could be due to the uncertainty of quantum measurements enhancing statistics and the relatively simple circuits employed in QAEs.


Preliminary findings suggest that quantum approaches can achieve comparable results to classical algorithms, especially for smaller training sets \cite{terashi2021event, ngairangbam2022anomaly}. Despite these promising outcomes, challenges such as increased training time and model complexity remain. Overall, quantum variational methods offer promising avenues for HEP data analysis, but their scalability and efficiency in comparison to classical techniques need further exploration.

\subsection{Chemistry}

Moussa et al. \cite{moussa2023application} utilized tensor network generative models for molecular discovery. Their work illuminates the data-specific effectiveness of generative models, showing that while GANs outperform TNs in some settings, the reverse is true in others. Specifically, GANs excelled on the QM9 molecule dataset \cite{ruddigkeit2012enumeration, ramakrishnan2014quantum} but were outperformed by TNs on an in-house antioxidants dataset. The study underscores the potential benefit of quantum-inspired and classical ensemble methods by showing that combining various generative models could result in more robust performance across different evaluation criteria. 

It should be noted that quantum-based methods have been used extensively in this domain, such as Variational Quantum Eigensolvers (VQE) which find the approximate ground state of a given Hamiltonian, often used in quantum chemistry and condensed matter physics problems \cite{li2019variational, bauer2020quantum}. Further, several quantum-inspired classical algorithms have been devised based on Gaussian boson sampling \cite{oh2022quantum, oh2023quantum} and unitary coupled cluster theory \cite{filip2020stochastic, chen2021quantum}. However, these methods do not fall under the purview of machine learning, as they primarily aim at solving specific physical problems through quantum simulation, rather than learning from data and generalizing to new instances, and as such are not discussed in this survey.

\subsection{Cybersecurity}

Quantum variational methods have been employed across a broad range of cybersecurity applications.


Payares and Martinez-Santos \cite{payares2021quantum} assessed QSVMs, VQCs, and an ensemble model for detecting DDoS attacks using the CIC-DDoS2019 dataset \cite{sharafaldin2019developing}, which closely mirrors real-world PCAP data. To accommodate quantum model computational limits, the dataset's 80 features were reduced to 2 using PCA. An angle-based strategy was applied for quantum embedding in all models. The ensemble model, drawing from quantum superposition, employed multiple parallel quantum classifiers. While all models showcased strong binary classification results, QSVM had high accuracy at 99.6\% but was computationally intensive. The ensemble model achieved 96.8\% accuracy efficiently, while the VQC excelled with 99.9\% accuracy and reasonable computational demand, albeit on a simplified dataset.

Masun et al. \cite{masum2022quantum} evaluated the performance of both QSVM and VQC for malware detection and source code vulnerability analysis, utilizing the ClaMP and Reveal datasets respectively. The dimensionality of the features was reduced using document vectorization and PCA to yield 16 explanatory variables. Both quantum methods under-performed compared to shallow classical neural networks and classical SVMs in malware detection, while achieving comparable performance in source code vulnerability analysis. Furthermore, both quantum methods exhibited significantly extended execution times.

Suryotrisongko and Musashi \cite{suryotrisongko2022evaluating} investigated the effect of adding a quantum circuit as a hidden layer in a classical neural network for domain generation algorithms (DGA)-based botnet detection. The classical model employs a standard deep learning architecture with 2 hidden layers (dense-dropout-dense), with the quantum layer inserted between the dense layers after dropout. Six combinations of ansatze were evaluated, using various embedding and entangling strategies made available by the PennyLane software framework. No single combination seemed to outperform all others across all settings, suggesting that quantum circuit architecture plays a significant role in determining the model's accuracy. The hybrid models performed slightly better than their classical counterparts under certain conditions. For instance, with the combination of Angle Embedding and Strongly Entangling Layers, the accuracy reached 94.7\% for 100 random samples. However, on average, the classical models outperformed the hybrid models.

Herr et al. \cite{herr2021anomaly} explored a variant of QGANs by adopting the AnoGan \cite{schlegl2017unsupervised} generative adversarial network structure, with the generative network portion replaced with a hybrid quantum-classical neural network. Specifically, a short state preparation layer encodes $N$ uniform latent variables as quantum states, which are fed into a parameterized quantum circuit of $N$ qubits. Measurement is performed over all qubits in the $Z$ basis, from which the now classical output is up-scaled via a classical dense network into a higher dimensional feature space. The intuition in using the VQC in the generator is in their ability to more efficiently sample from distributions that are hard to sample from classically. This is seen in experimental results over a credit card fraud dataset; the quantum AnoGAN method showed comparable F1 scores to variety of classical architectures and system sizes, whilst staying robust to changes in the dimension of the latent space.


Yang et al. \cite{yang2021decentralizing} proposed a novel decentralized feature extraction approach for speech recognition to address privacy-preservation issues. The framework is built upon a quantum convolutional neural network (QCNN), consisting of a quantum circuit encoder for feature extraction and a recurrent neural network (RNN) based end-to-end acoustic model (AM). This decentralized architecture enhances model parameter protection by first up-streaming an input speech to a quantum computing server to extract Mel-spectrogram feature vectors, encoding the corresponding convolutional features using a quantum circuit algorithm with random parameters, and then down-streaming the encoded features to the local RNN model for the final recognition. The authors test this approach on the Google Speech Commands dataset, attaining an accuracy of 95.12\%, showing competitive recognition results for spoken-term recognition when compared with classical DNN based AM models with the same convolutional kernel size. 


Other subfields within QiML have yet to see many works in the cybersecurity domain, with the anomaly detection work by Wang et al. \cite{wang2020anomaly} cited to have potential applications in fraud prevention and network security, among other applicable domains.


\subsection{Other Tasks and Applications}


Concerning dequantized algorithms, in general, while many works present the theoretical application of these methods to various ML tasks, few provide experimental analysis of the methods on data. Arrazola et al. \cite{arrazola2019quantum} both implemented, and performed analysis on the quantum-inspired algorithms for linear systems \cite{gilyen2019quantum} and recommendation systems \cite{tang2019quantum}. In implementation, the former was applied to portfolio optimization on stocks from the S\&P 500, and latter to a dataset of movies; significantly faster run-times were observed than what their complexity bounds would suggest. Analysis showed that when the rank and condition number are small, the dequantized algorithms provided good estimates in reasonable time, even for high-dimensional problems. However, outside this specification, the dequantized algorithms performed poorly in terms of both run time and estimation quality relative to direct computation of the solution on practical datasets. The authors note a threshold for improved relative performance when the matrix size is larger than $10^6$. Ding et al. \cite{ding2021quantum} test their quantum-inspired LS-SVM on low-rank, and low approximated rank synthetic data, and analyzed the performance against the classical counterpart LIBSVM. Their results show that their model outperformed LIBSVM by 5\% on average, and noted greater performance in low rank settings. However, the running times for both models are omitted. Chepurko et al. \cite{chepurko2022quantum} similarly provided analysis of their implemented algorithms. For the recommendation systems task, their algorithm operates in a similar setting to \cite{arrazola2019quantum}. The results indicated a notable six-fold speed increase, and demonstrated superior performance over direct computation methods. However, this enhancement was accompanied by a slight uptick in error. Comparable outcomes were found in their work on the ridge regression task. The observed improvements seem to stem from a more efficient implementation than \cite{arrazola2019quantum}, coupled with an algorithm possessing a superior asymptotic runtime. While the field has progressed far beyond these benchmarks, and given their inconclusive nature, the applicability of dequantized algorithms to practical data remains an open question, pending further investigation.

QiML has seen implementation over an assortment of ML tasks, typically over benchmark generic datasets, such as the UCI and PMLB \cite{Olson2017PMLB} datasets. Works involving the Quantum Nearest Mean Classifiers have used these datasets extensively to assess the model's capabilities, often as validation before moving to real-world contexts (Section \ref{sec:practice_medical}). In this setting, these works commonly cite improved performance over other baseline models. The models are also shown to be able to learn complex distribution, typically challenging for classical Nearest Mean Classifiers \cite{sergioli2017quantum}. However, a caveat of these methods is their much longer training and inference time compared with classical methods \cite{sergioli2020quantum}. Further, these benchmark datasets are typically small-sized. QNMCs have yet to see use over large-scale data. The HQC, despite it's prowess, admits a roadblock to this, as increasing the number of copies of samples introduces a non-negligible computation cost \cite{sergioli2019new}. The works inspired by quantum interference \cite{zhang2021interactive} and quantum correlation \cite{zhang2022quantum} show promise over these benchmark datasets, however also note computation costs to be a detriment to the applicability of these methods. These works defer improvements in training speed to the promise of realizable quantum computers.



In tensor network modeling, early works with MPS suggests the applicability of encoding features as polynomial functions \cite{novikov2018exponential}, showing success over generic classification tasks from the UCI dataset, recommendation systems, and synthetic data. Results show that inference time is competitive with baseline classical models, however training time suffers significantly, scaling with the chosen bond dimension of the MPS. In bit-string classification tasks, tensor network models have demonstrated superior learning capabilities compared to generative neural networks, especially for parity learning problems \cite{stokes2019probabilistic, evenbly2022number}. Establishing performance for challenging real-world tasks presents as potential future work.

\subsection{Limiting Factors on the Use of QiML in Practice}\label{sec:limiting}

\subsubsection{Dequantized Algorithms}\label{sec:limiting_deq}

A few key limitations to the applicability of these algorithms have been discussed in the literature. First, the requirements for the input matrices are often strict. The matrices must be of low stable rank \cite{tang2019quantum, gilyen2022improved}, have a small condition number \cite{jethwani2019quantum}, or be relatively sparse \cite{bakshi2023improved}. These requirements are generally not conducive to the needs of real-world datasets, though these conditions have progressively become more lax with advancements in the field. 



Secondly, the need for an input model that provides SQ access may not be readily amenable to current ML implementations. Performing the necessary preprocessing for adapting datasets to this structure may be reasonably assumed to be expensive and detrimental to computational efficiency, limiting the applicability of the these algorithms to existing systems \cite{tang2019overviewblog}.

Thirdly, many QML algorithms (and hence, often, the resulting dequantized algorithms) are tailored to solve tasks that deviate from what is conventionally addressed in the classical literature. For instance, while classical approaches to the recommendation systems problem typically employ low-rank matrix completion, the quantum algorithm instead executes sampling over a low-rank approximation of the input matrix \cite{kerenidis2016quantum}. In \cite{ding2021quantum}, a simplified version of the least squares SVM problem is considered by assuming data points are equally distributed across hyperplanes. Another example is the algorithm presented by \cite{chepurko2022quantum} which sees no classical counterpart. As such, many works do not remark on the performance of dequantized algorithms in comparison with other, more traditional classical algorithms for their examined tasks. 

Lastly, Chia et al. \cite{chia2022sampling} argues that dequantized algorithms operate under more restrictive and ostensibly weaker computation parameters compared to classical randomized numerical linear algebra algorithms. Dequantized algorithms assume the ability to efficiently measure quantum states related to the input data and aim to provide quick algorithms with dimension-independent runtime. However, this model is intrinsically weaker than its standard counterpart. In essence, a dequantized algorithm, with a runtime of $O(T)$, translates to a standard algorithm with a run time of $O(nnz(A) + T)$ dependent on both $T$ and the number of non-zero entries in the input matrix. Although this may lead to under-performance in typical sketching contexts, it broadens the range of problems where quantum speedup may not exponentially surpass classical solutions. Therefore, dequantized algorithms, in spite of their theoretical promise of exponentially improved runtime, may not perform as well as conventional sketching algorithms.

As such, the particular nature of the gap between classical and quantum ML algorithms remains an open question. In general, QML applications operate in either the low-rank or the high-rank data setting. The dequantization formalism suggests that most quantum linear algebra tasks over low-dimensional data can likely be dequantized into a classical variant, provided SQ access is available to that data. In contrast, evidence suggests that dequantizing high-dimensional problems incurs must greater difficultly. Several high-dimensional problems cannot be successfully dequantized despite SQ access. For example, the Fourier Sampling Problem is solved by randomized linear algebra techniques (i.e., SQ access) in exponential time, whereas the quantum version can find a solution in $O(1)$ \cite{aaronson2016complexity}. Furthermore, high-rank data frequently necessitates the use of the HHL algorithm or its variants, which have been discussed to be BQP-complete \cite{harrow2009quantum}. Another example is with quantum Boltzmann machine training, noted in \cite{tang2021quantum}, which cannot be dequantized in full unless BQP = BPP. This presents a significant impediment for classical algorithms trying to match the performance of their quantum counterparts. As such, QML algorithms can potentially extend their advantage in the high-rank setting by making assumptions such as taking sparse matrices as input or utilizing other high-rank quantum operations that cannot be efficiently implemented in the classical setting, such as the Quantum Fourier Transform \cite{chia2022sampling}. Additionally, dequantized algorithms hinge on cost-effective access to classically analogous QRAM. As of now, the quantum version of this input model has not been practically realized. However, should an efficient quantum input model be developed, one that eschews the need for expensive computations or classical interfacing, it could potentially prompt a reevaluation of the supposed advantages of dequantization methods.

In light of this, evidence has shown that there is a strong opportunity for classical algorithms to compete with quantum in the low-rank setting. In early works, for many tasks, the quantum algorithm still admitted a strong polynomial advantage. This restricted the applicability of several dequantized algorithms; even matrices with very low-rank could not be practically computed \cite{kerenidis2019q}. As noted by several authors, the cost of computation is dominated by the SVD computation that occurs after sampling down to the low-rank approximation \cite{tang2019quantum, chepurko2022quantum, bakshi2023improved}. New techniques have since been developed that bypass this computation, with Bakshi and Tang's method most recently demonstrating that low-degree QSVT circuits do not exhibit exponential advantage \cite{bakshi2023improved}. It should be noted that, at present, quantum algorithms polynomially surpass their dequantized counterparts. As both quantum and dequantized algorithms continue to improve their relative complexity bounds, it remains to be seen whether there is a limit to the performance of dequantized algorithms. An insurmountable threshold may exist that definitively ascribes quantum supremacy, or a classical regime might be discovered that denounces the advantage entirely.

\subsubsection{Tensor Networks}

Tensor networks learning models have scarcely stepped outside of a few, benchmark datasets, such as MNIST and Fashion-MNIST, due to difficulties in extending the tensor network model to larger inputs and higher dimensional feature spaces. In \cite{convy2022mutual}, the Tiny Image dataset was used, however the images were cropped to a 28 $\times$ 28 resolution and converted to gray scale, matching the context of MNIST images. For MPS structures, accommodating for images of larger resolution is difficult due to the inherent exponential loss of correlation across the network, which can also be exacerbated by common encoding methods. Images are typically flattened and encoded into high dimensional space \cite{stoudenmire2017supervised, efthymiou2019tensornetwork}; for small images, the pixel correlations are largely preserved, however they are lost when considering high resolution images \cite{selvan2020tensor}. This issue is less prevalent in higher order decompositions such as PEPS \cite{cheng2021supervised} and TTNs \cite{cheng2019tree}. However, several works have noted that the advantage of neural-network-based models over these tensor network methods lies in the better priors for images, made possible by the use of convolution. Tensor network methods that incorporate convolution are nearing the performance of traditional neural networks \cite{liu2023tensor}. The potential for discovering approaches within tensor network methods that could surpass neural networks is an area of interest and highlights a promising direction for future research.


Tensor network algorithms demand a high cost in both the bond dimension, a user-chosen free parameter, and the number of components contained after each local feature mapping, determined by choice of $\phi$ in Equation \ref{eqn:tn_feature_mapping}. Tensor network machine learning methods currently admit cubic, or even higher polynomial dependence on these parameters \cite{stoudenmire2017supervised, cheng2019tree}, despite only scaling linearly with the number of input components (e.g., the number of pixels in an input image). Thus there appears to be a a trade-off in the greater expressivity afforded by increasing these parameters and computation time.

The challenge of high dimensionality is especially prevalent in language modeling, as the semantic spaces of word vectors can be inherently large. This is in contrast to commonly used image modeling datasets, where pixel values can be represented in a low-dimensional format. The tensor products of such word vectors can become computationally challenging, maintaining high complexity even after tensor decompositions are applied \cite{zhang2019generalized}. This complexity may explain the limited research in the field of tensor networks for language modeling. In \cite{miller2021tensor}, tensor network evaluation was performed on a context-free language task using relatively simple, synthetic data. The field has yet to see robust, performant methods for complex language modeling tasks involving tensor networks.

These observations are supported in existing tensor network literature, stemming from the fact that classical tensor networks are only able to represent low-entangled, low-complexity states \cite{rieser2023tensor}. However this stipulation is less relevant outside of the quantum setting, where less complex classical input data is concerned which is inherently non-entangled. Further research may be necessary to understand what types and volumes of data become prohibitive in learning, and how to best utilize tensor networks for working with such data.

\subsubsection{Quantum Variational Algorithm Simulation}\label{sec:limiting_qvas}


The hybrid quantum-classical variational methods presented have seen success in performing computations over relatively small datasets, and using small-scale quantum circuits. Few methods have been presented outside this setting, due to the exponential limitation in simulating larger circuits with more qubits. For instance, methods using basis encoding require qubits that scale linearly with the number of representative features \cite{schuld2020circuit}. This restricts the number of features that can be used for learning. As such, when evaluating relative performance against classical counterparts, many works will apply constraints to classical methods in order to provide fair comparisons. This may involve severely condense the number of features \cite{wu2021applicationvqc, payares2021quantum}, or using a heavily reduced dataset size \cite{suryotrisongko2022evaluating}, for amenability with current-day quantum simulator architectures. Such limitations imply that comparisons between quantum simulation and classical methods in their fully-optimized, unrestricted settings may be inherently challenging, pending the development of more performant quantum algorithms.

The training time for simulating quantum algorithms is frequently noted to be more prolonged than their classical counterparts due to the inherent complexities of emulating quantum systems on traditional hardware \cite{esposito2022quantum, di2022dawn, gianelle2022quantum, esposito2022quantum, masum2022quantum}. In \cite{blance2021quantum}, the speedup over classical methods is owed to the use of quantum gradient descent used in the VQCs, allowing for faster convergence than classical neural networks using traditional gradient descent. Such optimizations could inform methods of decreasing training time in classical settings, for acceptable loss thresholds.

For classical simulation of QML in the noisy setting, such algorithms naturally inherit the limitations of QML, such as the need for robust gate error correction, and degradation of performance due to decoherence after prolonged training. Noiseless simulations are free from such issues, although they may not accurately represent the conditional settings of quantum hardware execution \cite{xu2023herculean}. Despite this advantage for classical implementation purposes, many of the works in this domain are forward-facing, developing methods and frameworks for true quantum computation, and assess their robustness with added noise and constraints. Very few studies have concentrated on the specific context where computation on classical machines is the primary objective of the proposed methodology.

\subsubsection{Other QiML Methods}

The range of QiML methods discussed demonstrate a diverse application of quantum theory to facilitate machine learning tasks. Generally, the introduction of quantum phenomena into these models is observed to enhance their expressive power compared to their classical counterparts, often resulting in improved performance \cite{zhang2018quantumqmwf, sergioli2020quantum, zhang2021interactive}. However, a commonly reported drawback among many of these methods is their high computational time complexity when compared with classical techniques. This issue predominantly affects models that rely heavily on quantum formalisms requiring computationally intensive operations, such as tensor products and complex number manipulations \cite{sergioli2020quantum, zhang2021interactive}. This is especially noted in methods that require matrix operations over density operators \cite{sergioli2017quantum, sergioli2018quantum, zhang2019quantum}, or require making tensor copies of quantum patterns in producing classification \cite{sergioli2019new, sergioli2021quantum, giuntini2023quantum}. In essence, there is an apparent trade-off between performance and computational speed, by way of simulating quantum operations via computationally heavy mathematical objects to incorporate greater expressivity in QiML models.

\section{Parallels Between QiML Models and Conventional Classical Models}

Efforts in QiML have applied quantum mechanics to enhance machine learning routines. Some of these efforts yield outcomes that bear resemblance to conventional classical ML models. In this subsection, we explore several of these parallels, shedding light on the relationships and distinctions between QiML techniques and their classical counterparts. By doing so, we aim to provide a familiar framework for understanding and interpreting QiML approaches. This approach could make the field more accessible to machine learning practitioners and researchers who are venturing into quantum information theory for the first time.



\subsection{Tensor Networks}

Previous studies have explored the relationship between tensor networks and neural network models \cite{cichocki2016low}. As mentioned in Section \ref{sec:tn_understanding}, several tensor network decompositions exhibit parallels between deep learning architectures, such as CNNs \cite{cohen2016expressive}, RNNs \cite{khrulkov2017expressive}, and RBMs \cite{gao2017efficient, chen2018equivalence}. Non-negative Matrix Product States (MPS) have been established as having a correspondence with Hidden Markov Models (HMM). Specifically, they can factorize probability mass functions of HMM into tensor network representations, capturing the essential stochastic relationships between hidden and observed variables \cite{glasser2019expressive}.

Tensor networks employ kernel learning approaches, similar to SVMs, in which samples are mapped to a higher dimensional feature space for improved separability \cite{sun2020generative}. 

The ability of tensor networks to model the joint distribution of variables, as seen in Tensor Network Born Machines (TNBMs), draws parallels with generative models like Variational Autoencoders (VAEs) and Generative Adversarial Networks (GANs) that parameterize conditional probabilities via deep neural networks \cite{liu2023tensor}. In fact, several works have shown that many-body quantum states can also be efficiently represented by neural network structures, such as DBMs \cite{carleo2018constructing} and shallow fully-connected neural networks \cite{cai2018approximating}, provided a simplified Hamiltonian ground state; similar conditions in which tensor networks see success.

\subsection{Quantum Variational Algorithm Simulation}

Variational quantum algorithms draw obvious parallels between classical machine learning methods. Quantum kernel methods, similar to SVMs, map data to a high-dimensional Hilbert space where they become linearly separable \cite{schuld2020circuit}. The computation of quantum kernel is effectively measuring the inner product in this Hilbert space, analogous to the operation of the kernel function in SVMs. VQCs combine classical optimization methods with a variational quantum circuit to learn a parameterized quantum state. This learning mechanism bears a strong resemblance to the operational principle of classical neural networks, which adjust weights and biases through an iterative optimization process to learn a function that can accurately classify data. Similarly, in a VQC, parameters of the quantum circuit are iteratively updated, effectively optimizing the quantum state to classify quantum data \cite{mitarai2018quantum}. As such, performance comparisons are often made between devised quantum algorithms and classical models of similar scale, i.e., by scaling down the number of available parameters in a classical neural network \cite{schuld2020circuit} 

The relationship between tensor networks and quantum circuits has also been explored in the literature, where tensor networks are seen to admit quantum circuits \cite{huggins2019towards}. As such, tensor network decompositions have inspired quantum variational methods. Hierarchical ansatz layouts mirroring tree-tensor and MERA networks have been shown to enable the classification of highly entangled states through greater expressive power \cite{grant2018hierarchical, lazzarin2022multi}.QCBMs utilize the probabilistic aspects of quantum wavefunctions for generative modeling, an approach that finds a parallel in Tensor Network Born Machines that use MPS for similar tasks \cite{liu2018differentiable, gong2022born}. While TNBMs are more aligned with explicit classical generative models, QCBMs show a greater affinity to implicit generative models, such as GANs.



\subsection{Dequantized Algorithms}

Regarding dequantized algorithms, extracting such correlations may initially seem an unassuming task. As noted in Section \ref{sec:limiting_deq}, the precise task that dequantized algorithms solve can be different to what is conventionally tackled. However, drawing face-value parallels between dequantized algorithms and classical machine learning methods seems intuitive and practical, since their underlying objectives are largely congruent. For instance, quantum-inspired SVMs aim to establish hyperplanes for classification, and quantum-inspired supervised clustering also engages in nearest centroid discrimination, mirroring their classical counterparts.

\subsection{Other QiML Methods}\label{sec:parallel_other}

Several QiML methods outside the aforementioned subsets incorporate quantum theory into existing classical structures. These include \cite{zhang2018quantum} and \cite{zhang2019quantum}, where density matrix projections are processed and feed into traditional classifiers such as CNNs, LSTMs and SVMs. In \cite{patel2019novel}, a traditional fully-connected neural network structure is realized using a fuzzy c-means-based learning process, where learning parameters are represented via quantum bits and quantum rotational gate operations. In \cite{shi2021two}, complex-valued word embedding are constructed via GRUs and scaled dot-product self-attention. Features are then extracted from projected density matrices via convolutional and max-pooling layers. 

Other works have produced methods that use quantum operations to mirror neural network behavior, such as in \cite{li2021quantumemotion} where a parameterized update function is used to evolve a “hidden” density matrix, which is updated at each time step based on the current input quantum state and the previous hidden density matrix, mirroring the key operational dynamics of an RNN.

Methods such as the QNMC and HQC \cite{sergioli2017quantum, sergioli2019new, leporini2022efficient} exhibit obvious similarities to classical nearest mean classification, as they rely on distance-based evaluation to a constructed centroid object. Methods involving quantum signal detection theory \cite{tiwari2018towards, tiwari2019binary, tiwari2019towards} share similarities to the Naive Bayes classification, where probabilities are calculated based on the frequencies of features. 

Some methods, like the Interactive Quantum Classifier (IQC) \cite{zhang2021interactive}, seem entirely quantum mechanical. The IQC interprets the classification process as an interaction between a physical target system and its environment, using quantum-inspired unitary transformations to adjust probability amplitudes and phases. Although the method involves common machine learning components, such as feature modeling and gradient-based updates, the core architecture of IQC is heavily rooted in quantum theory. 

As the field continues to evolve, we anticipate the emergence of other such models that deeply integrate quantum theory while still leveraging classical machine learning strategies to varying extents.


\subsection{Levels of Quantum Mechanics Integration}
Several sources of quantum inspiration have driven QiML learning methods. As briefly discussed in Section \ref{sec:parallel_other}, the various QiML methods vary in how much quantum mechanics they integrate, leading to both opportunities and challenges. On one hand, principles from quantum mechanics such as superposition and entanglement provide rich inspiration, allowing the development of innovative algorithms and encoding strategies beyond classical paradigms. On the other hand, implementing these quantum-infused methods in classical computing settings can be difficult. Quantum mechanics often exhibits complex correlations and dynamics that are hard to simulate classically, leading to potential exponential slowdowns or the need for approximations that may sacrifice quantum advantages. 

\begin{figure*}[!ht]
    \centering
    \centerline{\includegraphics[width=0.9\textwidth]{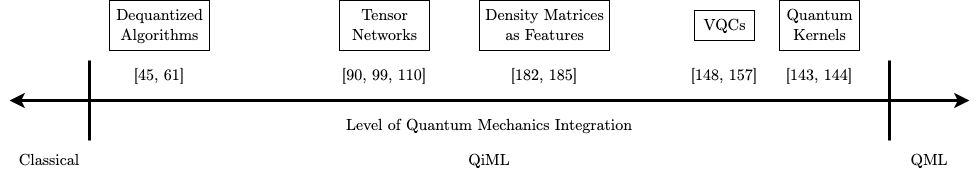}}
    \caption{QiML methods represented on a 1D spectrum, illustrating their relative positioning based on the level of quantum inspiration, as compared to purely classical and purely quantum machine learning. The positioning is not determined by a strict quantitative measure but rather reflects an informal assessment of the underlying principles and techniques. Citations given are representative works.}
    \label{fig:levels_of_inspiration}
\end{figure*}

Figure \ref{fig:levels_of_inspiration} illustrates a qualitative view of QiML methods based on the extent of quantum involvement. 
We surmise that methods incorporating more, or deeper levels of quantum mechanics tend to face greater challenges or limitations when attempting to simulate or implement them using classical computing resources. At the far right end of the spectrum, just before QML, lie variational quantum algorithms. Quantum circuit and quantum kernel learning methods heavily incorporate quantum aspects such as qubits, quantum gates, superposition, and entanglement. The classical-quantum hybridization of these techniques that rely on classical optimization allow for easier simulation on classical devices, however factors such as the circuit depth, width (i.e., number of qubits used), choice of gates, and encoding method can have a dramatic effect on this ease of simulation, as discussed in Section \ref{sec:limiting_qvas}. In other words, increasing the quantum-based complexity of these methods reduces the ability to classically simulate them.

QiML methods that project classical data into Hilbert feature spaces also exhibit varying levels of classical simulatability, typically tied to the nature of the projection and the dimensionality of the Hilbert space. Tensor network methods adapt techniques from quantum many-body physics, with network decompositions easing the computational burden by reducing the scope of the feature space. However, challenges still arise in their computation, as discussed in Section \ref{sec:limiting}. Additionally, higher-order tensor network methods such as PEPS often resort to approximation techniques instead of exact computation. In a machine learning context, these approximations may be sufficiently representative, as generalizations are usually more beneficial than precise exactitude. Furthermore, the need to operate over density matrices presents a common source of computational difficulty. Methods that rely on potentially large density matrices as features typically incur greater time complexity compared to classical ML methods using vector-based features, representing a trade-off for increased expressive power.

At the classical end of the spectrum, dequantized algorithms attempt to match QML methods using classical techniques. Although devoid of explicit quantum components, these methods can incur substantial computational costs in line with the matrix dimension and norms. Despite the intention to make them more classically amenable, dequantized algorithms may still be relatively slow, highlighting the intrinsic complexity and potential inefficiency of translating quantum-inspired techniques into classical paradigms. This mirrors the challenges found in more quantum-intensive techniques, revealing that the integration of quantum insights in classical contexts is a nuanced and demanding endeavor.

While the primary focus of this analysis has been on methods that trade computational capacity for quantum inspiration, it's worth noting a caveat: some QiML methods, including compression techniques, can actually experience speed-ups compared to their classical counterparts. This emphasizes the diverse potential of QiML, extending beyond mere computational trade-offs to offer tangible benefits and enhancements to classical ML.

\section{Available Resources for QiML Implementation}

We discuss the available resources allowing for the development and exploration of QiML methods.

\subsection{Research Implementations} 
While many of the explored works operate as standalone research models, not all perform empirical evaluations of their models using real data. Furthermore, only a subset of these works offer accessible code repositories that allow others to reproduce their results, with many authors stipulating that their code is available upon request. We highlight works that include accessible code in Table \ref{tab:qiml_practical_applications}, indicated by the asterisk (*). The ability to reproduce presented methods is an essential part of scientific inquiry. However, the quality and accessibility of these repositories can vary considerably, often due to varying levels of documentation and the specific requirements of certain computational environments and packages. This variation makes the replication process, and the adaptation of methods to wider domains and tasks challenging. Custom implementations of presented methods are less prevalent in tensor network research, thanks in part to the availability of well-developed toolboxes and libraries, as we discuss in Section \ref{disc:toolboxes}. 

We advocate for more uniformity in the way code is shared in the QiML research community. Greater transparency, along with increased adoption of best practices for documentation and repository organization, will enhance the accessibility of these implementations and enable more robust scientific discourse, especially in an emerging field.

\subsection{Toolboxes}\label{disc:toolboxes}

\begin{table*}[!ht]
    \centering
    \small
    \caption{QiML Toolboxes}
    \resizebox{0.95\textwidth}{!}{%
    \begin{tabular}{|l|p{1.4cm}|p{5.5cm}|p{1.8cm}|p{3.3cm}|}
        \hline
        \textbf{Toolbox} & \textbf{Mode} & \textbf{Functionality} & \textbf{Languages} & \textbf{Research Group} \\
        \hline
        LIQ\textit{Ui}$\ket{}$ (2014) \cite{liqui} & QVAS & comprehensive framework for quantum programming with three built-in classes of simulators & F\# & Microsoft Research \\
        NCON (2014) \cite{pfeifer2014ncon} & TN & functions that facilitate tensor network contractions, which are integral to several other tensor network toolboxes & MATLAB & Perimeter Institute for Theoretical Physics \\

        TT-Toolbox (2014) \cite{dolgov2014computation} & TN & basic tensor arithmetic, contractions, and routines involving MPS & MATLAB, Fortran, Python & Institute of Numerical Mathematics RAS \\
        Tensorly (2016) \cite{kossaifi2016tensorly} & TN & tensor methods and deep tensorized neural networks via several Python backends & Python & Imperial College London \\

        Qiskit (2017) \cite{Qiskit} & QVAS & software development framework for modeling circuits, algorithms, and hardware & Python & IBM  \\
        Cirq (2018) \cite{Cirq2023} & QVAS & quantum programming library for NISQ hardware control and simulation & Python & Google, Open-source\\

        PennyLane (2018) \cite{bergholm2018pennylane} & QVAS & open-source quantum software library for quantum machine learning tasks with support for various quantum computing platforms & Python & Xanadu AI \\

        Scikit-TT (2018) \cite{scikittt} & TN & MPS methods for representing and solving linear systems & Python & Freie Universität Berlin \\
        HQC (2019) \cite{sergioli2019new} & Other (QNMC) & probabilisitc classification with parallelization available & Python & University of Cagliari \\
        TensorNetwork (2019) \cite{roberts2019tensornetwork} & TN & defining and manipulating general tensor network models & Python & Alphabet (Google) X \\
        TorchMPS (2019) \cite{torchmps} & TN & MPS modeling with DMRG support via PyTorch backend & Python & Université de Montréal \\

        Intel Quantum Simulator (2020) \cite{guerreschi2020intel} & QVAS & quantum circuit simulator with high-performance computing capabilities & C++, Python & Intel Labs\\

        PastaQ (2020) \cite{pastaq} & TN, QVAS & various quantum circuit simulation methods using tensor-network representations & Julia & Flatiron Institute \\
        Qulacs (2020) \cite{suzuki2021qulacs} & QVAS & fast, low-scale quantum circuit simulator that provides a wide range of built-in quantum gates and operations & Python, C++ & QunaSys, Osaka University, NTT,  Fujitsu \\
        TensorFlow Quantum (2020) \cite{broughton2020tensorflow} & QVAS & provides tools and frameworks for building hybrid quantum-classical models & Python & Google \\

        lambeq (2021) \cite{kartsaklis2021lambeq} & QVAS & library for end-to-end quantum NLP pipeline development & Python & Cambridge Quantum Computing \\

        Qibo (2021) \cite{qibo_paper} & QVAS & builds and runs quantum circuits, supporting GPU, multi-GPU, and multi-threaded CPU. & Python & Quantum Research Center (QRC) \\

        ITensor (2022) \cite{fishman2022itensor} & TN & tensor arithmetic, contractions, and support for MPS and MPO decompositions & C++, Julia & Flatiron Institute \\
        tntorch (2022) \cite{UBS:22} & TN & supports tensor factorizations, including CP, Tucker, and MPS, and offers autodifferentiation optimization & Python & IE University, Madrid \\
        \hline
    \end{tabular}
    }
    \label{tab:qiml_toolboxes}
\end{table*}

Various toolboxes and libraries have been developed for tensor network operations and quantum circuit simulation. Table \ref{tab:qiml_toolboxes} outlines a few commonly used, available packages for QiML purposes.

Basic tensor operations are supported by implementations for various languages. Most prominently, Python has the TT-Toolbox \cite{dolgov2014computation}, TorchMPS \cite{torchmps} and Scikit-TT \cite{scikittt} frameworks, which all provide MPS solvers with support for DMRG optimization. tntorch \cite{UBS:22} facilitates auto differentiation-based optimization for MPS. Tensorly \cite{kossaifi2016tensorly} and TensorNetwork \cite{roberts2019tensornetwork} Python libraries offer more generalized functions, allowing for additional decomposition formats with support for various Python backends, such as PyTorch or JAX which provide the machine learning functionality. Non-specialised, generic Python libraries such as Numpy \cite{harris2020array}, which provides a base for many tensor network libraries, have also been successfully used on a standalone basis \cite{sun2020generative, liu2023tensor}. MATLAB, C++ and Julia also see a host of supporting tensor network libraries. Psarras et al. \cite{psarras2021landscape} provides a comprehensive survey of existing tensor network software. Wang et al. \cite{wang2023tensor} categorizes tensor network toolboxes based on their functionality and application areas. We collate the ones that have been used by researchers in the QiML context in Table \ref{tab:qiml_toolboxes}. For works that did not explicitly mention what packages were used, we discovered them by inspecting noted code repositories.

While these toolboxes have been used to much success, a few limitations present themselves, namely the lack of both predefined models for higher-order tensor decompositions (such as PEPS and TTN), and input embedding pipelines. This restricts the ability for users to freely produce and develop new models.

Several libraries have been developed to facilitate quantum circuit construction and simulation. PennyLane \cite{bergholm2018pennylane} and TensorFlow Quantum \cite{broughton2020tensorflow} are heavily focused on the integration of quantum computing with classical machine learning frameworks, such as PyTorch and TensorFlow (predominantly in the latter), with back-end support for various quantum simulation platforms. Qulacs \cite{suzuki2021qulacs} and Qibo \cite{qibo_paper} provide efficient and flexible standalone quantum simulators operable on personal computing devices. Intel Quantum Simulator \cite{guerreschi2020intel} offers similar support, with adaptations for high-performance computing environments. Qiskit \cite{Qiskit} provides general comprehensive quantum software development kits that supports a wide range of quantum computing workflows, including local simulation, circuit optimization, and execution on IBM's cloud-based quantum hardware. Cirq \cite{Cirq2023} allows user to build quantum circuits for near-term quantum hardware and NISQ (Noisy Intermediate-Scale Quantum) devices, giving fine-tuned control over quantum program execution.

\begin{table*}[!ht]
    \centering
    \caption{Various Cloud-based Quantum Computing Platforms}
    \begin{tabular}{|p{2.5cm}|l|p{2cm}|p{2cm}|l|p{2.8cm}|p{2cm}|}
        \hline
        \textbf{Platform} & \textbf{No. Sim. Qubits} & \textbf{Noisy/Noiseless Simulation} & \textbf{Languages} & \textbf{Integration with:} & \textbf{Quantum Hardware Accessibility} \\
        \hline
        Amazon Braket \cite{amazonbraket} & 34 & Yes/Yes & Python & Amazon Web Services & Yes \\
        \hline
        Google Quantum AI \cite{google2023quantumai} & 40 & Yes/Yes & C++, Python & Cirq, TensorFlow Quantum & Yes \\
        \hline
        IBM Quantum \cite{IBMQuantum} & 32 & Yes/Yes & Python, Swift & Qiskit & Yes \\
        \hline
        Microsoft Azure Quantum Cloud Service \cite{microsoft2023azurequantum} & 29-32 & Yes/Yes & Q\# & .NET & Yes \\
        \hline
        qBraid \cite{qBraid} & \~29-32 & Yes/Yes & Python & Cirq, Braket, Qiskit & Yes \\
        \hline
        QuTech Quantum Inspire \cite{qutech} & 26-34 & Yes/Yes & cQASM, Python & Qiskit & Yes \\
        \hline
        Orquestra \cite{Orquestra} & \~29-32 & Yes/Yes & Python & Cirq, D-Wave, PennyLane, Qiskit & Yes \\
        \hline

    \end{tabular}
\label{table:quantum_platforms}
\end{table*}

The Quantum Nearest Mean Classifier sees a package for its implementation, which allows for application over custom data, and offers parallel computing capabilities \cite{sergioli2019new}. Methods for input modeling and encoding are not provided in this repository.

In contrast, there is a sparsity in toolboxes that facilitate the implementation of practical models for dequantized algorithms. This may be primarily due to the theoretical nature of the research, the level of specificity necessary for adapting general ideas to particular tasks, and the (im)maturity of the field. As highlighted in \cite{arrazola2019quantum}, claimed complexities may not always be indicative of real-world application scenarios. The development and introduction of frameworks could potentially provide researchers with deeper insights into these issues. Additionally, they could serve as a tools for validating proposed methods through comparative or ablative studies. Similar observations apply to the various other QiML methods. However, given that these methods have seen practical implementations with direct application of the proposed methods, we can anticipate the development of dedicated toolboxes for them in the near future.

\subsection{Commercial Applications}

Several noisy intermediate-scale quantum computing architectures have been developed and commercialized and used for various applications. Cloud-based compute for quantum simulation has been employed extensively in practical research, particularly for applications requiring computational resources beyond the scope of personal computing. Common cloud-computing platforms employed in literature are outlined in Table \ref{table:quantum_platforms}, which offer high-performance simulation capabilities. IBM Quantum \cite{IBMQuantum} provides the QasmSimulator and StatevectorSimulator backends through Qiskit Aer: QasmSimulator allows for multi-shot execution of circuits, while StatevectorSimulator also returns the final statevector of the simulator after application. Both simulate up to 32 qubits in both noisy and noise-free settings. Google's qsim \cite{google2023quantumai} provides a full wavefunction quantum circuit simulator that leverages vectorized optimization and multi-threading, capable of simulating up to 40 qubits. Amazon Braket \cite{amazonbraket} provides on-demand state vector, tensor network and density matrix simulators, similar to IBM Quantum, with a current qubit limit of 34. Quantum hardware accessibility is offloaded to third-party providers. QuTech offer high performance cluster computing power via their Quantum Inspire platform \cite{qutech}, simulating up to 34 qubits and allows for inspection of the simulated quantum state. The cQASM hardware-agnostic quantum assembly language is used for constructing circuits, with a Python API also made available. Microsoft's Azure Quantum platform \cite{microsoft2023azurequantum} offers three back-end simulators, from the IonQ, Quantuum and Rigetti providers. IonQ provides a GPU-accelerated idealized simulator supporting up to 29 qubits, Quantuum provides emulators of real physical quantum models supporting up to 32 qubits, and Rigetti provides a cloud service simulator for Quil, a quantum instruction set language, supporting up to 30 qubits. Other providers, such as Orquestra \cite{Orquestra} and qBraid \cite{qBraid}, also similarly offer end-to-end software development architectures using external quantum devices, with additional front-end support from various libraries such as Cirq, Braket and Qiskit.

A few quantum-inspired-related frameworks are available such as the Fujitsu \cite{sao2019application}, NEC \cite{komatsu2021performance} and D-Wave quantum-inspired annealing services\footnote{https://docs.ocean.dwavesys.com/en/stable/}, with the latter being applied to a few, small-scale quantum-inspired \cite{mugel2022dynamic} and quantum-assisted \cite{hu2019quantum} ML applications. Outside of these services, dedicated QiML architectures have yet to see such production and adoption.


In general, there are ample toolbox options available for TN and QVAS methods, with commercial platform-as-a-service providers supplying compute power for quantum simulation. However there is a lack of available tooling outside of these areas, which limits the scope of choices for ML practitioners in exploring QiML solutions. The gap presents ample opportunities for the development, and potential commercialization, of such frameworks, which could in turn catalyze further research in the field.

\section{Open Issues}

QiML research is subject to the numerous challenges of an emerging discipline. In the practical setting, these challenges largely revolve around how QiML can be used for a broader range of applications, and in more performant ways. We identify potential issues in furthering this goal.



\begin{enumerate}

    \item \textbf{Speed and Performance:} 
    A significant challenge in present QiML methods is the dichotomy between the complexity of the methods and their performance. Works have shown the performance of QiML methods in general is worse than that of contemporary classical methods. Where they do show competitive results, this is typically caveated by slower runtimes, larger model sizes or greater incurred error \cite{tang2019overviewblog, sergioli2020quantum, liu2023tensor}. In the case of QVAS models, comparable performance is often only observed when classical models are deliberately scaled back in terms of architecture size, the number of parameters and/or number of input features. A few exceptions to this have been presented, particularly in methods designed with parameter reduction in mind \cite{patel2022quantum, konar20233}. In general, however, there is a clear necessity for research that improves the speed and performance of QiML, striking a balance between complexity and effectiveness.

    
    \item \textbf{Constraints on Input Data:}
    
    Dequantized algorithms and tensor network learning methods are predicated on the assumption that input data exhibits low rank characteristics: low linear algebraic rank and low bond dimension, respectively. However, such assumptions may not hold true in real-world scenarios, where data can often be high-dimensional and complex. While there are claims advocating for the broad applicability of QiML methods \cite{tang2021quantum}, these assertions have yet to be effectively demonstrated on large, real-world datasets. Therefore, addressing the challenge of high-dimensional data representation in quantum machine learning remains a key area for future research and development. 



    \item \textbf{Alternative Input and Embedding Methods:}
    
    Currently, methods for transforming classical data to be compatible with QiML methods are largely under-explored. Input modeling typically adheres to a few established embedding schemes, for example Equation \ref{eqn:tn_feature_mapping} for tensor networks, or density matrix-based representations commonly seen in QiML subsets. Though these appear to work in generalized settings, authors have noted potential adaptations, such as having independent mappings per feature, or promoting the mapping space to higher dimensions \cite{stoudenmire2017supervised} for TNs. Analyzing and suggesting appropriate access models for adequate comparisons between dequantized and quantum algorithms is also an emerging area of concern. As the embedding method itself often directly depends on the nature of the underlying data, investigation of the effects of different embedding schemes on factors such as performance and information entropy, researchers can gain insights into optimal methods for specific data types and tasks.

    \item \textbf{The Need for Comprehensive Tooling:}
    
    There is a significant need for comprehensive, user-friendly tooling in the field of QiML. Existing toolboxes, while useful, do not fully meet the needs of researchers and developers. They often lack the extensive array of tools and capabilities required to effectively develop new models, re-implement existing ones, or explore innovative research avenues, especially in the areas of tensor networks and other QiML methods. The development of more robust toolsets could greatly accelerate the pace of advancement in this promising field.

    \item \textbf{Effective Quantum Formalisms in QiML}
    
    In the current literature, several quantum formalisms that inspire QiML have demonstrated considerable success, especially ones that contribute towards building and utilizing quantum feature spaces. However, there remains a lack of systematic documentation identifying which quantum principles are adaptable to classical ML, the reasons for their success or potential, and indeed, which quantum concepts may not translate well or at all. The exploration of novel quantum mechanics adaptations within classical ML is an ongoing area of research. Consequently, the QiML field would significantly benefit from a methodical analysis delineating which approaches are effective and which are not.


\end{enumerate}

\section{Conclusion}\label{sec:conclusion}

Quantum-inspired Machine Learning has seen a rapid expansion in recent years, diversifying into numerous research directions, such as tensor network simulations, dequantized algorithms and other such methods that draw inspiration from quantum physics. Prior surveys have alluded to QiML, often presenting it as a facet of QML or concentrating on specific QiML subsets. In contrast, this survey provides a comprehensive examination, bringing these emergent fields together under the QiML umbrella. We have explored recent work across these areas, highlighting their practical applications, offering readers an understanding of where and how QiML has been used. This insight can potentially guide readers in exploring QiML for their specific use cases. Significantly, we strive to pin down a more precise definition of QiML, addressing the issue of vague and generic descriptions prevalent in previous work. Furthermore, this review illuminates crucial open issues in QiML, particularly pertaining to its current level of practical applicability. As we move forward, we anticipate the emergence of new QiML methods. Quantum mechanics, quantum computing and classical machine learning are expansive fields with a constant influx of emerging knowledge and techniques. The untapped potential of these fields presents a vast reservoir of methods and approaches that could further enrich QiML. The continuous evolution of these fields presents a fertile ground for novel perspectives and cross-pollination, promising to stimulate the growth and diversification of QiML.



\ifCLASSOPTIONcompsoc
  \section*{Acknowledgments}
\else
  \section*{Acknowledgment}
\fi
This work was supported by CSIRO’s Quantum Technologies Future Science Platform. Ajmal Mian is the recipient of an Australian Research Council Future Fellowship Award (project number FT210100268) funded by the Australian Government.


\ifCLASSOPTIONcaptionsoff
  \newpage
\fi



\bibliographystyle{IEEEtran}
\bibliography{refs.bib}

\begin{thebibliography}{100}
\providecommand{\url}[1]{#1}
\csname url@samestyle\endcsname
\providecommand{\newblock}{\relax}
\providecommand{\bibinfo}[2]{#2}
\providecommand{\BIBentrySTDinterwordspacing}{\spaceskip=0pt\relax}
\providecommand{\BIBentryALTinterwordstretchfactor}{4}
\providecommand{\BIBentryALTinterwordspacing}{\spaceskip=\fontdimen2\font plus
\BIBentryALTinterwordstretchfactor\fontdimen3\font minus
  \fontdimen4\font\relax}
\providecommand{\BIBforeignlanguage}[2]{{%
\expandafter\ifx\csname l@#1\endcsname\relax
\typeout{** WARNING: IEEEtran.bst: No hyphenation pattern has been}%
\typeout{** loaded for the language `#1'. Using the pattern for}%
\typeout{** the default language instead.}%
\else
\language=\csname l@#1\endcsname
\fi
#2}}
\providecommand{\BIBdecl}{\relax}
\BIBdecl

\bibitem{ciliberto2018quantum}
C.~Ciliberto, M.~Herbster, A.~D. Ialongo, M.~Pontil, A.~Rocchetto, S.~Severini,
  and L.~Wossnig, ``Quantum machine learning: a classical perspective,''
  \emph{Proceedings of the Royal Society A: Mathematical, Physical and
  Engineering Sciences}, vol. 474, no. 2209, p. 20170551, 2018.

\bibitem{cerezo2022challenges}
M.~Cerezo, G.~Verdon, H.-Y. Huang, L.~Cincio, and P.~J. Coles, ``Challenges and
  opportunities in quantum machine learning,'' \emph{Nature Computational
  Science}, vol.~2, no.~9, pp. 567--576, 2022.

\bibitem{preskill2018quantum}
J.~Preskill, ``Quantum computing in the nisq era and beyond,'' \emph{Quantum},
  vol.~2, p.~79, 2018.

\bibitem{schuld2021machine}
M.~Schuld and F.~Petruccione, \emph{Machine learning with quantum
  computers}.\hskip 1em plus 0.5em minus 0.4em\relax Springer, 2021.

\bibitem{khan2020machine}
T.~M. Khan and A.~Robles-Kelly, ``Machine learning: Quantum vs classical,''
  \emph{IEEE Access}, vol.~8, pp. 219\,275--219\,294, 2020.

\bibitem{garg2020advances}
S.~Garg and G.~Ramakrishnan, ``Advances in quantum deep learning: An
  overview,'' \emph{arXiv preprint arXiv:2005.04316}, 2020.

\bibitem{abohashima2020classification}
Z.~Abohashima, M.~Elhosen, E.~H. Houssein, and W.~M. Mohamed, ``Classification
  with quantum machine learning: A survey,'' 2020.

\bibitem{mishra2021quantum}
N.~Mishra, M.~Kapil, H.~Rakesh, A.~Anand, N.~Mishra, A.~Warke, S.~Sarkar,
  S.~Dutta, S.~Gupta, A.~Prasad~Dash \emph{et~al.}, ``Quantum machine learning:
  A review and current status,'' \emph{Data Management, Analytics and
  Innovation: Proceedings of ICDMAI 2020, Volume 2}, pp. 101--145, 2021.

\bibitem{houssein2022machine}
E.~H. Houssein, Z.~Abohashima, M.~Elhoseny, and W.~M. Mohamed, ``Machine
  learning in the quantum realm: The state-of-the-art, challenges, and future
  vision,'' \emph{Expert Systems with Applications}, p. 116512, 2022.

\bibitem{zeguendry2023quantum}
A.~Zeguendry, Z.~Jarir, and M.~Quafafou, ``Quantum machine learning: A review
  and case studies,'' \emph{Entropy}, vol.~25, no.~2, p. 287, 2023.

\bibitem{moore1995quantum}
M.~Moore and A.~Narayanan, ``Quantum-inspired computing,'' \emph{Dept. Comput.
  Sci., Univ. Exeter, Exeter, UK}, 1995.

\bibitem{steane1998quantum}
A.~Steane, ``Quantum computing,'' \emph{Reports on Progress in Physics},
  vol.~61, no.~2, p. 117, 1998.

\bibitem{narayanan1996quantum}
A.~Narayanan and M.~Moore, ``Quantum-inspired genetic algorithms,'' in
  \emph{Proceedings of IEEE international conference on evolutionary
  computation}.\hskip 1em plus 0.5em minus 0.4em\relax IEEE, 1996, pp. 61--66.

\bibitem{han2000genetic}
K.-H. Han and J.-H. Kim, ``Genetic quantum algorithm and its application to
  combinatorial optimization problem,'' in \emph{Proceedings of the 2000
  congress on evolutionary computation. CEC00 (Cat. No. 00TH8512)},
  vol.~2.\hskip 1em plus 0.5em minus 0.4em\relax IEEE, 2000, pp. 1354--1360.

\bibitem{han2002quantum}
------, ``Quantum-inspired evolutionary algorithm for a class of combinatorial
  optimization,'' \emph{IEEE transactions on evolutionary computation}, vol.~6,
  no.~6, pp. 580--593, 2002.

\bibitem{zhang2011quantum}
G.~Zhang, ``Quantum-inspired evolutionary algorithms: a survey and empirical
  study,'' \emph{Journal of Heuristics}, vol.~17, no.~3, pp. 303--351, 2011.

\bibitem{manju2014applications}
A.~Manju and M.~J. Nigam, ``Applications of quantum inspired computational
  intelligence: a survey,'' \emph{Artificial Intelligence Review}, vol.~42, pp.
  79--156, 2014.

\bibitem{varmantchaonala2022quantum}
C.~Varmantchaonala~Moudina, J.~L. Fendji Kedieng~Ebongue, and M.~Atemkeng,
  ``Quantum algorithms for combinatorial optimization problems: A comprehensive
  survey from 2000 to 2022,'' \emph{Jean Louis and Atemkeng, Marcel, Quantum
  Algorithms for Combinatorial Optimization Problems: A Comprehensive Survey
  from}, 2022.

\bibitem{ross2019review}
O.~H.~M. Ross, ``A review of quantum-inspired metaheuristics: Going from
  classical computers to real quantum computers,'' \emph{Ieee Access}, vol.~8,
  pp. 814--838, 2019.

\bibitem{han2001parallel}
K.-H. Han, K.-H. Park, C.-H. Lee, and J.-H. Kim, ``Parallel quantum-inspired
  genetic algorithm for combinatorial optimization problem,'' in
  \emph{Proceedings of the 2001 congress on evolutionary computation (IEEE Cat.
  No. 01TH8546)}, vol.~2.\hskip 1em plus 0.5em minus 0.4em\relax IEEE, 2001,
  pp. 1422--1429.

\bibitem{sun2004particle}
J.~Sun, B.~Feng, and W.~Xu, ``Particle swarm optimization with particles having
  quantum behavior,'' in \emph{Proceedings of the 2004 congress on evolutionary
  computation (IEEE Cat. No. 04TH8753)}, vol.~1.\hskip 1em plus 0.5em minus
  0.4em\relax IEEE, 2004, pp. 325--331.

\bibitem{wang2007novel}
L.~Wang, Q.~Niu, and M.~Fei, ``A novel quantum ant colony optimization
  algorithm,'' in \emph{Bio-Inspired Computational Intelligence and
  Applications: International Conference on Life System Modeling and
  Simulation, LSMS 2007, Shanghai, China, September 14-17, 2007.
  Proceedings}.\hskip 1em plus 0.5em minus 0.4em\relax Springer, 2007, pp.
  277--286.

\bibitem{yongjun2017application}
D.~Yongjun and L.~Jiying, ``The application of quantum-inspired ant colony
  algorithm in automatic segmentation of tomato image,'' in \emph{2017 2nd
  International Conference on Image, Vision and Computing (ICIVC)}.\hskip 1em
  plus 0.5em minus 0.4em\relax IEEE, 2017, pp. 341--345.

\bibitem{das2023quantum}
M.~Das, A.~Roy, S.~Maity, and S.~Kar, ``A quantum-inspired ant colony
  optimization for solving a sustainable four-dimensional traveling salesman
  problem under type-2 fuzzy variable,'' \emph{Advanced Engineering
  Informatics}, vol.~55, p. 101816, 2023.

\bibitem{somma2007quantum}
R.~Somma, S.~Boixo, and H.~Barnum, ``Quantum simulated annealing,'' 2007.

\bibitem{gharehchopogh2022quantum}
F.~S. Gharehchopogh, ``Quantum-inspired metaheuristic algorithms: Comprehensive
  survey and classification,'' \emph{Artificial Intelligence Review}, pp.
  1--65, 2022.

\bibitem{dos2012nuclear}
A.~dos Santos~Nicolau, R.~Schirru, and A.~M.~M. de~Lima, ``Nuclear reactor
  reload using quantum inspired algorithm,'' \emph{Progress in Nuclear Energy},
  vol.~55, pp. 40--48, 2012.

\bibitem{samuel1959machine}
A.~L. Samuel, ``Machine learning,'' \emph{The Technology Review}, vol.~62,
  no.~1, pp. 42--45, 1959.

\bibitem{mitchell2007machine}
T.~M. Mitchell \emph{et~al.}, \emph{Machine learning}.\hskip 1em plus 0.5em
  minus 0.4em\relax McGraw-hill New York, 2007, vol.~1.

\bibitem{goodfellow2016deep}
I.~Goodfellow, Y.~Bengio, and A.~Courville, \emph{Deep learning}.\hskip 1em
  plus 0.5em minus 0.4em\relax MIT press, 2016.

\bibitem{murphy2012machine}
K.~P. Murphy, \emph{Machine learning: a probabilistic perspective}.\hskip 1em
  plus 0.5em minus 0.4em\relax MIT press, 2012.

\bibitem{wittek2014quantum}
P.~Wittek, \emph{Quantum machine learning: what quantum computing means to data
  mining}.\hskip 1em plus 0.5em minus 0.4em\relax Academic Press, 2014.

\bibitem{domingos2012few}
P.~Domingos, ``A few useful things to know about machine learning,''
  \emph{Communications of the ACM}, vol.~55, no.~10, pp. 78--87, 2012.

\bibitem{mahesh2020machine}
B.~Mahesh, ``Machine learning algorithms-a review,'' \emph{International
  Journal of Science and Research (IJSR).[Internet]}, vol.~9, pp. 381--386,
  2020.

\bibitem{carrasquilla2020machine}
J.~Carrasquilla, ``Machine learning for quantum matter,'' \emph{Advances in
  Physics: X}, vol.~5, no.~1, p. 1797528, 2020.

\bibitem{alchieri2021introduction}
L.~Alchieri, D.~Badalotti, P.~Bonardi, and S.~Bianco, ``An introduction to
  quantum machine learning: from quantum logic to quantum deep learning,''
  \emph{Quantum Machine Intelligence}, vol.~3, pp. 1--30, 2021.

\bibitem{wu2021application}
S.~L. Wu, S.~Sun, W.~Guan, C.~Zhou, J.~Chan, C.~L. Cheng, T.~Pham, Y.~Qian,
  A.~Z. Wang, R.~Zhang \emph{et~al.}, ``Application of quantum machine learning
  using the quantum kernel algorithm on high energy physics analysis at the
  lhc,'' \emph{Physical Review Research}, vol.~3, no.~3, p. 033221, 2021.

\bibitem{riaz2023accurate}
F.~Riaz, S.~Abdulla, H.~Suzuki, S.~Ganguly, R.~C. Deo, and S.~Hopkins,
  ``Accurate image multi-class classification neural network model with quantum
  entanglement approach,'' \emph{Sensors}, vol.~23, no.~5, p. 2753, 2023.

\bibitem{Qiskit}
{Qiskit contributors}, ``Qiskit: An open-source framework for quantum
  computing,'' 2023.

\bibitem{bergholm2018pennylane}
V.~Bergholm, J.~Izaac, M.~Schuld, C.~Gogolin, S.~Ahmed, V.~Ajith, M.~S. Alam,
  G.~Alonso-Linaje, B.~AkashNarayanan, A.~Asadi \emph{et~al.}, ``Pennylane:
  Automatic differentiation of hybrid quantum-classical computations,''
  \emph{arXiv preprint arXiv:1811.04968}, 2018.

\bibitem{aimeur2006machine}
E.~A{\"\i}meur, G.~Brassard, and S.~Gambs, ``Machine learning in a quantum
  world,'' in \emph{Advances in Artificial Intelligence: 19th Conference of the
  Canadian Society for Computational Studies of Intelligence, Canadian AI 2006,
  Qu{\'e}bec City, Qu{\'e}bec, Canada, June 7-9, 2006. Proceedings 19}.\hskip
  1em plus 0.5em minus 0.4em\relax Springer, 2006, pp. 431--442.

\bibitem{tang2019overviewblog}
E.~Tang, ``An overview of quantum-inspired classical sampling,''
  \url{https://ewintang.com/blog/2019/01/28/an-overview-of-quantum-inspired-sampling/},
  2019, accessed: 2023-04-30.

\bibitem{tang2019quantum}
------, ``A quantum-inspired classical algorithm for recommendation systems,''
  in \emph{Proceedings of the 51st annual ACM SIGACT symposium on theory of
  computing}, 2019, pp. 217--228.

\bibitem{kerenidis2016quantum}
I.~Kerenidis and A.~Prakash, ``Quantum recommendation systems,'' \emph{arXiv
  preprint arXiv:1603.08675}, 2016.

\bibitem{chia2022sampling}
\BIBentryALTinterwordspacing
N.-H. Chia, A.~P. Gily\'{e}n, T.~Li, H.-H. Lin, E.~Tang, and C.~Wang,
  ``Sampling-based sublinear low-rank matrix arithmetic framework for
  dequantizing quantum machine learning,'' \emph{J. ACM}, vol.~69, no.~5, oct
  2022. [Online]. Available: \url{https://doi.org/10.1145/3549524}
\BIBentrySTDinterwordspacing

\bibitem{chepurko2022quantum}
N.~Chepurko, K.~Clarkson, L.~Horesh, H.~Lin, and D.~Woodruff,
  ``Quantum-inspired algorithms from randomized numerical linear algebra,'' in
  \emph{International Conference on Machine Learning}.\hskip 1em plus 0.5em
  minus 0.4em\relax PMLR, 2022, pp. 3879--3900.

\bibitem{bakshi2023improved}
A.~Bakshi and E.~Tang, ``An improved classical singular value transformation
  for quantum machine learning,'' \emph{arXiv preprint arXiv:2303.01492}, 2023.

\bibitem{tang2021quantum}
E.~Tang, ``Quantum principal component analysis only achieves an exponential
  speedup because of its state preparation assumptions,'' \emph{Physical Review
  Letters}, vol. 127, no.~6, p. 060503, 2021.

\bibitem{lloyd2013quantum}
S.~Lloyd, M.~Mohseni, and P.~Rebentrost, ``Quantum algorithms for supervised
  and unsupervised machine learning,'' \emph{arXiv preprint arXiv:1307.0411},
  2013.

\bibitem{lloyd2014quantum}
------, ``Quantum principal component analysis,'' \emph{Nature Physics},
  vol.~10, no.~9, pp. 631--633, 2014.

\bibitem{courrieu2008fast}
P.~Courrieu, ``Fast computation of moore-penrose inverse matrices,''
  \emph{arXiv preprint arXiv:0804.4809}, 2008.

\bibitem{geron2022hands}
A.~G{\'e}ron, \emph{Hands-on machine learning with Scikit-Learn, Keras, and
  TensorFlow}.\hskip 1em plus 0.5em minus 0.4em\relax " O'Reilly Media, Inc.",
  2022.

\bibitem{harrow2009quantum}
A.~W. Harrow, A.~Hassidim, and S.~Lloyd, ``Quantum algorithm for linear systems
  of equations,'' \emph{Physical review letters}, vol. 103, no.~15, p. 150502,
  2009.

\bibitem{gilyen2018quantum}
A.~Gily{\'e}n, S.~Lloyd, and E.~Tang, ``Quantum-inspired low-rank stochastic
  regression with logarithmic dependence on the dimension,'' \emph{arXiv
  preprint arXiv:1811.04909}, 2018.

\bibitem{chia2018quantum}
N.-H. Chia, H.-H. Lin, and C.~Wang, ``Quantum-inspired sublinear classical
  algorithms for solving low-rank linear systems,'' \emph{arXiv preprint
  arXiv:1811.04852}, 2018.

\bibitem{gilyen2022improved}
A.~Gily{\'e}n, Z.~Song, and E.~Tang, ``An improved quantum-inspired algorithm
  for linear regression,'' \emph{Quantum}, vol.~6, p. 754, 2022.

\bibitem{shao2022faster}
C.~Shao and A.~Montanaro, ``Faster quantum-inspired algorithms for solving
  linear systems,'' \emph{ACM Transactions on Quantum Computing}, vol.~3,
  no.~4, pp. 1--23, 2022.

\bibitem{rebentrost2014quantum}
P.~Rebentrost, M.~Mohseni, and S.~Lloyd, ``Quantum support vector machine for
  big data classification,'' \emph{Physical review letters}, vol. 113, no.~13,
  p. 130503, 2014.

\bibitem{ding2021quantum}
C.~Ding, T.-Y. Bao, and H.-L. Huang, ``Quantum-inspired support vector
  machine,'' \emph{IEEE Transactions on Neural Networks and Learning Systems},
  vol.~33, no.~12, pp. 7210--7222, 2021.

\bibitem{chia2019quantum}
N.-H. Chia, T.~Li, H.-H. Lin, and C.~Wang, ``Quantum-inspired sublinear
  algorithm for solving low-rank semidefinite programming,'' \emph{arXiv
  preprint arXiv:1901.03254}, 2019.

\bibitem{gilyen2019quantum}
A.~Gily{\'e}n, Y.~Su, G.~H. Low, and N.~Wiebe, ``Quantum singular value
  transformation and beyond: exponential improvements for quantum matrix
  arithmetics,'' in \emph{Proceedings of the 51st Annual ACM SIGACT Symposium
  on Theory of Computing}, 2019, pp. 193--204.

\bibitem{martyn2021grand}
J.~M. Martyn, Z.~M. Rossi, A.~K. Tan, and I.~L. Chuang, ``Grand unification of
  quantum algorithms,'' \emph{PRX Quantum}, vol.~2, no.~4, p. 040203, 2021.

\bibitem{low2019hamiltonian}
G.~H. Low and I.~L. Chuang, ``Hamiltonian simulation by qubitization,''
  \emph{Quantum}, vol.~3, p. 163, 2019.

\bibitem{jethwani2019quantum}
D.~Jethwani, F.~L. Gall, and S.~K. Singh, ``Quantum-inspired classical
  algorithms for singular value transformation,'' \emph{arXiv preprint
  arXiv:1910.05699}, 2019.

\bibitem{gharibian2022dequantizing}
S.~Gharibian and F.~Le~Gall, ``Dequantizing the quantum singular value
  transformation: Hardness and applications to quantum chemistry and the
  quantum pcp conjecture,'' in \emph{Proceedings of the 54th Annual ACM SIGACT
  Symposium on Theory of Computing}, 2022, pp. 19--32.

\bibitem{babbush2021focus}
R.~Babbush, J.~R. McClean, M.~Newman, C.~Gidney, S.~Boixo, and H.~Neven,
  ``Focus beyond quadratic speedups for error-corrected quantum advantage,''
  \emph{PRX Quantum}, vol.~2, no.~1, p. 010103, 2021.

\bibitem{cong2016quantum}
I.~Cong and L.~Duan, ``Quantum discriminant analysis for dimensionality
  reduction and classification,'' \emph{New Journal of Physics}, vol.~18,
  no.~7, p. 073011, 2016.

\bibitem{feynman2018simulating}
R.~P. Feynman, ``Simulating physics with computers,'' in \emph{Feynman and
  computation}.\hskip 1em plus 0.5em minus 0.4em\relax CRC Press, 2018, pp.
  133--153.

\bibitem{rudi2020approximating}
A.~Rudi, L.~Wossnig, C.~Ciliberto, A.~Rocchetto, M.~Pontil, and S.~Severini,
  ``Approximating hamiltonian dynamics with the nystr{\"o}m method,''
  \emph{Quantum}, vol.~4, p. 234, 2020.

\bibitem{zhao2021smooth}
Z.~Zhao, J.~K. Fitzsimons, P.~Rebentrost, V.~Dunjko, and J.~F. Fitzsimons,
  ``Smooth input preparation for quantum and quantum-inspired machine
  learning,'' \emph{Quantum Machine Intelligence}, vol.~3, no.~1, p.~14, 2021.

\bibitem{cotler2021revisiting}
J.~Cotler, H.-Y. Huang, and J.~R. McClean, ``Revisiting dequantization and
  quantum advantage in learning tasks,'' \emph{arXiv preprint
  arXiv:2112.00811}, 2021.

\bibitem{huang2022quantum}
H.-Y. Huang, M.~Broughton, J.~Cotler, S.~Chen, J.~Li, M.~Mohseni, H.~Neven,
  R.~Babbush, R.~Kueng, J.~Preskill \emph{et~al.}, ``Quantum advantage in
  learning from experiments,'' \emph{Science}, vol. 376, no. 6598, pp.
  1182--1186, 2022.

\bibitem{van2018improvements}
J.~van Apeldoorn and A.~Gily{\'e}n, ``Improvements in quantum sdp-solving with
  applications,'' \emph{arXiv preprint arXiv:1804.05058}, 2018.

\bibitem{orus2014practical}
R.~Or{\'u}s, ``A practical introduction to tensor networks: Matrix product
  states and projected entangled pair states,'' \emph{Annals of physics}, vol.
  349, pp. 117--158, 2014.

\bibitem{montangero2018introduction}
S.~Montangero, E.~Montangero, and Evenson, \emph{Introduction to tensor network
  methods}.\hskip 1em plus 0.5em minus 0.4em\relax Springer, 2018.

\bibitem{biamonte2017tensor}
J.~Biamonte and V.~Bergholm, ``Tensor networks in a nutshell,'' \emph{arXiv
  preprint arXiv:1708.00006}, 2017.

\bibitem{bridgeman2017hand}
J.~C. Bridgeman and C.~T. Chubb, ``Hand-waving and interpretive dance: an
  introductory course on tensor networks,'' \emph{Journal of physics A:
  Mathematical and theoretical}, vol.~50, no.~22, p. 223001, 2017.

\bibitem{wang2023tensor}
M.~Wang, Y.~Pan, X.~Yang, G.~Li, and Z.~Xu, ``Tensor networks meet neural
  networks: A survey,'' \emph{arXiv preprint arXiv:2302.09019}, 2023.

\bibitem{rieser2023tensor}
H.-M. Rieser, F.~K{\"o}ster, and A.~P. Raulf, ``Tensor networks for quantum
  machine learning,'' \emph{arXiv preprint arXiv:2303.11735}, 2023.

\bibitem{huggins2019towards}
W.~Huggins, P.~Patil, B.~Mitchell, K.~B. Whaley, and E.~M. Stoudenmire,
  ``Towards quantum machine learning with tensor networks,'' \emph{Quantum
  Science and technology}, vol.~4, no.~2, p. 024001, 2019.

\bibitem{araz2022classical}
J.~Y. Araz and M.~Spannowsky, ``Classical versus quantum: comparing tensor
  network-based quantum circuits on lhc data,'' \emph{arXiv preprint
  arXiv:2202.10471}, 2022.

\bibitem{eisert2010colloquium}
J.~Eisert, M.~Cramer, and M.~B. Plenio, ``Colloquium: Area laws for the
  entanglement entropy,'' \emph{Reviews of modern physics}, vol.~82, no.~1, p.
  277, 2010.

\bibitem{wall2021generative}
M.~L. Wall, M.~R. Abernathy, and G.~Quiroz, ``Generative machine learning with
  tensor networks: Benchmarks on near-term quantum computers,'' \emph{Physical
  Review Research}, vol.~3, no.~2, p. 023010, 2021.

\bibitem{grant2018hierarchical}
E.~Grant, M.~Benedetti, S.~Cao, A.~Hallam, J.~Lockhart, V.~Stojevic, A.~G.
  Green, and S.~Severini, ``Hierarchical quantum classifiers,'' \emph{npj
  Quantum Information}, vol.~4, no.~1, p.~65, 2018.

\bibitem{wall2021tree}
M.~L. Wall and G.~D'Aguanno, ``Tree-tensor-network classifiers for machine
  learning: From quantum inspired to quantum assisted,'' \emph{Physical Review
  A}, vol. 104, no.~4, p. 042408, 2021.

\bibitem{dborin2022matrix}
J.~Dborin, F.~Barratt, V.~Wimalaweera, L.~Wright, and A.~G. Green, ``Matrix
  product state pre-training for quantum machine learning,'' \emph{Quantum
  Science and Technology}, vol.~7, no.~3, p. 035014, 2022.

\bibitem{stoudenmire2018learning}
E.~M. Stoudenmire, ``Learning relevant features of data with multi-scale tensor
  networks,'' \emph{Quantum Science and Technology}, vol.~3, no.~3, p. 034003,
  2018.

\bibitem{perez2006matrix}
D.~Perez-Garcia, F.~Verstraete, M.~M. Wolf, and J.~I. Cirac, ``Matrix product
  state representations,'' \emph{arXiv preprint quant-ph/0608197}, 2006.

\bibitem{hastings2007area}
M.~B. Hastings, ``An area law for one-dimensional quantum systems,''
  \emph{Journal of statistical mechanics: theory and experiment}, vol. 2007,
  no.~08, p. P08024, 2007.

\bibitem{paeckel2019time}
S.~Paeckel, T.~K{\"o}hler, A.~Swoboda, S.~R. Manmana, U.~Schollw{\"o}ck, and
  C.~Hubig, ``Time-evolution methods for matrix-product states,'' \emph{Annals
  of Physics}, vol. 411, p. 167998, 2019.

\bibitem{stoudenmire2017supervised}
E.~M. Stoudenmire and D.~J. Schwab, ``Supervised learning with quantum-inspired
  tensor networks,'' 2017.

\bibitem{scholkopf2018learning}
B.~Scholkopf and A.~J. Smola, \emph{Learning with kernels: support vector
  machines, regularization, optimization, and beyond}.\hskip 1em plus 0.5em
  minus 0.4em\relax MIT press, 2018.

\bibitem{zhang2018quantumqmwf}
P.~Zhang, Z.~Su, L.~Zhang, B.~Wang, and D.~Song, ``A quantum many-body wave
  function inspired language modeling approach,'' in \emph{Proceedings of the
  27th ACM International Conference on Information and Knowledge Management},
  2018, pp. 1303--1312.

\bibitem{zhang2019generalized}
L.~Zhang, P.~Zhang, X.~Ma, S.~Gu, Z.~Su, and D.~Song, ``A generalized language
  model in tensor space,'' in \emph{Proceedings of the AAAI Conference on
  Artificial Intelligence}, vol.~33, no.~01, 2019, pp. 7450--7458.

\bibitem{miller2021tensor}
J.~Miller, G.~Rabusseau, and J.~Terilla, ``Tensor networks for probabilistic
  sequence modeling,'' in \emph{International Conference on Artificial
  Intelligence and Statistics}.\hskip 1em plus 0.5em minus 0.4em\relax PMLR,
  2021, pp. 3079--3087.

\bibitem{tangpanitanon2022explainable}
J.~Tangpanitanon, C.~Mangkang, P.~Bhadola, Y.~Minato, D.~G. Angelakis, and
  T.~Chotibut, ``Explainable natural language processing with matrix product
  states,'' \emph{New Journal of Physics}, vol.~24, no.~5, p. 053032, 2022.

\bibitem{novikov2018exponential}
A.~Novikov, M.~Trofimov, and I.~Oseledets, ``Exponential machines,''
  \emph{Bulletin of the Polish Academy of Sciences Technical Sciences}, pp.
  789--797, 2018.

\bibitem{schollwock2011density}
U.~Schollw{\"o}ck, ``The density-matrix renormalization group in the age of
  matrix product states,'' \emph{Annals of physics}, vol. 326, no.~1, pp.
  96--192, 2011.

\bibitem{efthymiou2019tensornetwork}
S.~Efthymiou, J.~Hidary, and S.~Leichenauer, ``Tensornetwork for machine
  learning,'' \emph{arXiv preprint arXiv:1906.06329}, 2019.

\bibitem{sun2020generative}
Z.-Z. Sun, C.~Peng, D.~Liu, S.-J. Ran, and G.~Su, ``Generative tensor network
  classification model for supervised machine learning,'' \emph{Physical Review
  B}, vol. 101, no.~7, p. 075135, 2020.

\bibitem{glasser2020probabilistic}
I.~Glasser, N.~Pancotti, and J.~I. Cirac, ``From probabilistic graphical models
  to generalized tensor networks for supervised learning,'' \emph{IEEE Access},
  vol.~8, pp. 68\,169--68\,182, 2020.

\bibitem{araz2021quantum}
J.~Y. Araz and M.~Spannowsky, ``Quantum-inspired event reconstruction with
  tensor networks: Matrix product states,'' \emph{Journal of High Energy
  Physics}, vol. 2021, no.~8, pp. 1--28, 2021.

\bibitem{zettili2009quantum}
N.~Zettili, ``Quantum mechanics: concepts and applications,'' 2009.

\bibitem{han2018unsupervised}
Z.-Y. Han, J.~Wang, H.~Fan, L.~Wang, and P.~Zhang, ``Unsupervised generative
  modeling using matrix product states,'' \emph{Physical Review X}, vol.~8,
  no.~3, p. 031012, 2018.

\bibitem{alcazar2020classical}
J.~Alcazar, V.~Leyton-Ortega, and A.~Perdomo-Ortiz, ``Classical versus quantum
  models in machine learning: insights from a finance application,''
  \emph{Machine Learning: Science and Technology}, vol.~1, no.~3, p. 035003,
  2020.

\bibitem{stokes2019probabilistic}
J.~Stokes and J.~Terilla, ``Probabilistic modeling with matrix product
  states,'' \emph{Entropy}, vol.~21, no.~12, p. 1236, 2019.

\bibitem{bradley2020modeling}
T.-D. Bradley, E.~M. Stoudenmire, and J.~Terilla, ``Modeling sequences with
  quantum states: a look under the hood,'' \emph{Machine Learning: Science and
  Technology}, vol.~1, no.~3, p. 035008, 2020.

\bibitem{shi2006classical}
Y.-Y. Shi, L.-M. Duan, and G.~Vidal, ``Classical simulation of quantum
  many-body systems with a tree tensor network,'' \emph{Physical review a},
  vol.~74, no.~2, p. 022320, 2006.

\bibitem{vidal2007entanglement}
G.~Vidal, ``Entanglement renormalization,'' \emph{Physical review letters},
  vol.~99, no.~22, p. 220405, 2007.

\bibitem{cheng2019tree}
S.~Cheng, L.~Wang, T.~Xiang, and P.~Zhang, ``Tree tensor networks for
  generative modeling,'' \emph{Physical Review B}, vol.~99, no.~15, p. 155131,
  2019.

\bibitem{liu2019machine}
D.~Liu, S.-J. Ran, P.~Wittek, C.~Peng, R.~B. Garc{\'\i}a, G.~Su, and
  M.~Lewenstein, ``Machine learning by unitary tensor network of hierarchical
  tree structure,'' \emph{New Journal of Physics}, vol.~21, no.~7, p. 073059,
  2019.

\bibitem{lecun2010mnist}
Y.~LeCun, C.~Cortes, and C.~Burges, ``Mnist handwritten digit database,''
  \emph{ATT Labs [Online]. Available: http://yann.lecun.com/exdb/mnist},
  vol.~2, 2010.

\bibitem{verstraete2004renormalization}
F.~Verstraete and J.~I. Cirac, ``Renormalization algorithms for quantum-many
  body systems in two and higher dimensions,'' \emph{arXiv preprint
  cond-mat/0407066}, 2004.

\bibitem{cirac2021matrix}
J.~I. Cirac, D.~Perez-Garcia, N.~Schuch, and F.~Verstraete, ``Matrix product
  states and projected entangled pair states: Concepts, symmetries, theorems,''
  \emph{Reviews of Modern Physics}, vol.~93, no.~4, p. 045003, 2021.

\bibitem{cheng2021supervised}
S.~Cheng, L.~Wang, and P.~Zhang, ``Supervised learning with projected entangled
  pair states,'' \emph{Physical Review B}, vol. 103, no.~12, p. 125117, 2021.

\bibitem{vieijra2022generative}
T.~Vieijra, L.~Vanderstraeten, and F.~Verstraete, ``Generative modeling with
  projected entangled-pair states,'' \emph{arXiv preprint arXiv:2202.08177},
  2022.

\bibitem{xiao2017fashion}
H.~Xiao, K.~Rasul, and R.~Vollgraf, ``Fashion-mnist: a novel image dataset for
  benchmarking machine learning algorithms,'' \emph{arXiv preprint
  arXiv:1708.07747}, 2017.

\bibitem{hubig2017generic}
C.~Hubig, I.~McCulloch, and U.~Schollw{\"o}ck, ``Generic construction of
  efficient matrix product operators,'' \emph{Physical Review B}, vol.~95,
  no.~3, p. 035129, 2017.

\bibitem{tensornetwork.org}
``Tensor network,'' \url{tensornetwork.org}, accessed: 2023-07-27.

\bibitem{novikov2015tensorizing}
A.~Novikov, D.~Podoprikhin, A.~Osokin, and D.~P. Vetrov, ``Tensorizing neural
  networks,'' \emph{Advances in neural information processing systems},
  vol.~28, 2015.

\bibitem{gao2020compressing}
Z.-F. Gao, S.~Cheng, R.-Q. He, Z.-Y. Xie, H.-H. Zhao, Z.-Y. Lu, and T.~Xiang,
  ``Compressing deep neural networks by matrix product operators,''
  \emph{Physical Review Research}, vol.~2, no.~2, p. 023300, 2020.

\bibitem{patel2022quantum}
R.~Patel, C.-W. Hsing, S.~Sahin, S.~S. Jahromi, S.~Palmer, S.~Sharma,
  C.~Michel, V.~Porte, M.~Abid, S.~Aubert \emph{et~al.}, ``Quantum-inspired
  tensor neural networks for partial differential equations,'' \emph{arXiv
  preprint arXiv:2208.02235}, 2022.

\bibitem{wang2020anomaly}
J.~Wang, C.~Roberts, G.~Vidal, and S.~Leichenauer, ``Anomaly detection with
  tensor networks,'' \emph{arXiv preprint arXiv:2006.02516}, 2020.

\bibitem{liu2023tensor}
J.~Liu, S.~Li, J.~Zhang, and P.~Zhang, ``Tensor networks for unsupervised
  machine learning,'' \emph{Physical Review E}, vol. 107, no.~1, p. L012103,
  2023.

\bibitem{selvan2020tensor}
R.~Selvan and E.~B. Dam, ``Tensor networks for medical image classification,''
  in \emph{Medical Imaging with Deep Learning}.\hskip 1em plus 0.5em minus
  0.4em\relax PMLR, 2020, pp. 721--732.

\bibitem{konar20233}
D.~Konar, S.~Bhattacharyya, T.~K. Gandhi, B.~K. Panigrahi, and R.~Jiang, ``3-d
  quantum-inspired self-supervised tensor network for volumetric segmentation
  of medical images,'' \emph{IEEE Transactions on Neural Networks and Learning
  Systems}, 2023.

\bibitem{liu2018entanglement}
Y.~Liu, X.~Zhang, M.~Lewenstein, and S.-J. Ran, ``Entanglement-guided
  architectures of machine learning by quantum tensor network,'' \emph{arXiv
  preprint arXiv:1803.09111}, 2018.

\bibitem{convy2022mutual}
I.~Convy, W.~Huggins, H.~Liao, and K.~B. Whaley, ``Mutual information scaling
  for tensor network machine learning,'' \emph{Machine learning: science and
  technology}, vol.~3, no.~1, p. 015017, 2022.

\bibitem{torralba200880}
A.~Torralba, R.~Fergus, and W.~T. Freeman, ``80 million tiny images: A large
  data set for nonparametric object and scene recognition,'' \emph{IEEE
  transactions on pattern analysis and machine intelligence}, vol.~30, no.~11,
  pp. 1958--1970, 2008.

\bibitem{hashemizadeh2020adaptive}
M.~Hashemizadeh, M.~Liu, J.~Miller, and G.~Rabusseau, ``Adaptive learning of
  tensor network structures,'' \emph{arXiv preprint arXiv:2008.05437}, 2020.

\bibitem{cohen2016expressive}
N.~Cohen, O.~Sharir, and A.~Shashua, ``On the expressive power of deep
  learning: A tensor analysis,'' in \emph{Conference on learning theory}.\hskip
  1em plus 0.5em minus 0.4em\relax PMLR, 2016, pp. 698--728.

\bibitem{khrulkov2017expressive}
V.~Khrulkov, A.~Novikov, and I.~Oseledets, ``Expressive power of recurrent
  neural networks,'' \emph{arXiv preprint arXiv:1711.00811}, 2017.

\bibitem{glasser2019expressive}
I.~Glasser, R.~Sweke, N.~Pancotti, J.~Eisert, and I.~Cirac, ``Expressive power
  of tensor-network factorizations for probabilistic modeling,'' \emph{Advances
  in neural information processing systems}, vol.~32, 2019.

\bibitem{lu2021tensor}
S.~Lu, M.~Kan{\'a}sz-Nagy, I.~Kukuljan, and J.~I. Cirac, ``Tensor networks and
  efficient descriptions of classical data,'' \emph{arXiv preprint
  arXiv:2103.06872}, 2021.

\bibitem{gao2017efficient}
X.~Gao and L.-M. Duan, ``Efficient representation of quantum many-body states
  with deep neural networks,'' \emph{Nature communications}, vol.~8, no.~1, p.
  662, 2017.

\bibitem{chen2018equivalence}
J.~Chen, S.~Cheng, H.~Xie, L.~Wang, and T.~Xiang, ``Equivalence of restricted
  boltzmann machines and tensor network states,'' \emph{Physical Review B},
  vol.~97, no.~8, p. 085104, 2018.

\bibitem{levine2019quantum}
Y.~Levine, O.~Sharir, N.~Cohen, and A.~Shashua, ``Quantum entanglement in deep
  learning architectures,'' \emph{Physical review letters}, vol. 122, no.~6, p.
  065301, 2019.

\bibitem{georgescu2014quantum}
I.~M. Georgescu, S.~Ashhab, and F.~Nori, ``Quantum simulation,'' \emph{Reviews
  of Modern Physics}, vol.~86, no.~1, p. 153, 2014.

\bibitem{daley2022practical}
A.~J. Daley, I.~Bloch, C.~Kokail, S.~Flannigan, N.~Pearson, M.~Troyer, and
  P.~Zoller, ``Practical quantum advantage in quantum simulation,''
  \emph{Nature}, vol. 607, no. 7920, pp. 667--676, 2022.

\bibitem{gujju2023quantum}
Y.~Gujju, A.~Matsuo, and R.~Raymond, ``Quantum machine learning on near-term
  quantum devices: Current state of supervised and unsupervised techniques for
  real-world applications,'' \emph{arXiv preprint arXiv:2307.00908}, 2023.

\bibitem{zhang2020recent}
Y.~Zhang and Q.~Ni, ``Recent advances in quantum machine learning,''
  \emph{Quantum Engineering}, vol.~2, no.~1, p. e34, 2020.

\bibitem{cerezo2021variational}
M.~Cerezo, A.~Arrasmith, R.~Babbush, S.~C. Benjamin, S.~Endo, K.~Fujii, J.~R.
  McClean, K.~Mitarai, X.~Yuan, L.~Cincio \emph{et~al.}, ``Variational quantum
  algorithms,'' \emph{Nature Reviews Physics}, vol.~3, no.~9, pp. 625--644,
  2021.

\bibitem{de2022survey}
G.~De~Luca, ``A survey of nisq era hybrid quantum-classical machine learning
  research,'' \emph{Journal of Artificial Intelligence and Technology}, vol.~2,
  no.~1, pp. 9--15, 2022.

\bibitem{garcia2022systematic}
D.~P. Garc{\'\i}a, J.~Cruz-Benito, and F.~J. Garc{\'\i}a-Pe{\~n}alvo,
  ``Systematic literature review: Quantum machine learning and its
  applications,'' \emph{arXiv preprint arXiv:2201.04093}, 2022.

\bibitem{preskill2012quantum}
J.~Preskill, ``Quantum computing and the entanglement frontier,'' \emph{arXiv
  preprint arXiv:1203.5813}, 2012.

\bibitem{zhou2020limits}
Y.~Zhou, E.~M. Stoudenmire, and X.~Waintal, ``What limits the simulation of
  quantum computers?'' \emph{Physical Review X}, vol.~10, no.~4, p. 041038,
  2020.

\bibitem{wu2019full}
X.-C. Wu, S.~Di, E.~M. Dasgupta, F.~Cappello, H.~Finkel, Y.~Alexeev, and F.~T.
  Chong, ``Full-state quantum circuit simulation by using data compression,''
  in \emph{Proceedings of the International Conference for High Performance
  Computing, Networking, Storage and Analysis}, 2019, pp. 1--24.

\bibitem{chen201864}
Z.-Y. Chen, Q.~Zhou, C.~Xue, X.~Yang, G.-C. Guo, and G.-P. Guo, ``64-qubit
  quantum circuit simulation,'' \emph{Science Bulletin}, vol.~63, no.~15, pp.
  964--971, 2018.

\bibitem{pan2022simulation}
F.~Pan and P.~Zhang, ``Simulation of quantum circuits using the big-batch
  tensor network method,'' \emph{Physical Review Letters}, vol. 128, no.~3, p.
  030501, 2022.

\bibitem{xu2023herculean}
X.~Xu, S.~Benjamin, J.~Sun, X.~Yuan, and P.~Zhang, ``A herculean task:
  Classical simulation of quantum computers,'' \emph{arXiv preprint
  arXiv:2302.08880}, 2023.

\bibitem{amin2018quantum}
M.~H. Amin, E.~Andriyash, J.~Rolfe, B.~Kulchytskyy, and R.~Melko, ``Quantum
  boltzmann machine,'' \emph{Physical Review X}, vol.~8, no.~2, p. 021050,
  2018.

\bibitem{havlivcek2019supervised}
V.~Havl{\'\i}{\v{c}}ek, A.~D. C{\'o}rcoles, K.~Temme, A.~W. Harrow, A.~Kandala,
  J.~M. Chow, and J.~M. Gambetta, ``Supervised learning with quantum-enhanced
  feature spaces,'' \emph{Nature}, vol. 567, no. 7747, pp. 209--212, 2019.

\bibitem{simoes2023experimental}
R.~D.~M. Sim{\~o}es, P.~Huber, N.~Meier, N.~Smailov, R.~M. F{\"u}chslin, and
  K.~Stockinger, ``Experimental evaluation of quantum machine learning
  algorithms,'' \emph{IEEE Access}, vol.~11, pp. 6197--6208, 2023.

\bibitem{payares2021quantum}
E.~Payares and J.~C. Mart{\'\i}nez-Santos, ``Quantum machine learning for
  intrusion detection of distributed denial of service attacks: a comparative
  overview,'' \emph{Quantum Computing, Communication, and Simulation}, vol.
  11699, pp. 35--43, 2021.

\bibitem{masum2022quantum}
M.~Masum, M.~Nazim, M.~J.~H. Faruk, H.~Shahriar, M.~Valero, M.~A.~H. Khan,
  G.~Uddin, S.~Barzanjeh, E.~Saglamyurek, A.~Rahman \emph{et~al.}, ``Quantum
  machine learning for software supply chain attacks: How far can we go?'' in
  \emph{2022 IEEE 46th Annual Computers, Software, and Applications Conference
  (COMPSAC)}.\hskip 1em plus 0.5em minus 0.4em\relax IEEE, 2022, pp. 530--538.

\bibitem{benedetti2019parameterized}
M.~Benedetti, E.~Lloyd, S.~Sack, and M.~Fiorentini, ``Parameterized quantum
  circuits as machine learning models,'' \emph{Quantum Science and Technology},
  vol.~4, no.~4, p. 043001, 2019.

\bibitem{mitarai2018quantum}
K.~Mitarai, M.~Negoro, M.~Kitagawa, and K.~Fujii, ``Quantum circuit learning,''
  \emph{Physical Review A}, vol.~98, no.~3, p. 032309, 2018.

\bibitem{schuld2020circuit}
M.~Schuld, A.~Bocharov, K.~M. Svore, and N.~Wiebe, ``Circuit-centric quantum
  classifiers,'' \emph{Physical Review A}, vol. 101, no.~3, p. 032308, 2020.

\bibitem{islam2022hybrid}
M.~Islam, M.~Chowdhury, Z.~Khan, and S.~M. Khan, ``Hybrid quantum-classical
  neural network for cloud-supported in-vehicle cyberattack detection,''
  \emph{IEEE Sensors Letters}, vol.~6, no.~4, pp. 1--4, 2022.

\bibitem{emmanoulopoulos2022quantum}
D.~Emmanoulopoulos and S.~Dimoska, ``Quantum machine learning in finance: Time
  series forecasting,'' \emph{arXiv preprint arXiv:2202.00599}, 2022.

\bibitem{terashi2021event}
K.~Terashi, M.~Kaneda, T.~Kishimoto, M.~Saito, R.~Sawada, and J.~Tanaka,
  ``Event classification with quantum machine learning in high-energy
  physics,'' \emph{Computing and Software for Big Science}, vol.~5, pp. 1--11,
  2021.

\bibitem{gianelle2022quantum}
A.~Gianelle, P.~Koppenburg, D.~Lucchesi, D.~Nicotra, E.~Rodrigues, L.~Sestini,
  J.~de~Vries, and D.~Zuliani, ``Quantum machine learning for b-jet charge
  identification,'' \emph{Journal of High Energy Physics}, vol. 2022, no.~8,
  pp. 1--24, 2022.

\bibitem{blance2021quantum}
A.~Blance and M.~Spannowsky, ``Quantum machine learning for particle physics
  using a variational quantum classifier,'' \emph{Journal of High Energy
  Physics}, vol. 2021, no.~2, pp. 1--20, 2021.

\bibitem{stokes2020quantum}
J.~Stokes, J.~Izaac, N.~Killoran, and G.~Carleo, ``Quantum natural gradient,''
  \emph{Quantum}, vol.~4, p. 269, 2020.

\bibitem{cong2019quantum}
I.~Cong, S.~Choi, and M.~D. Lukin, ``Quantum convolutional neural networks,''
  \emph{Nature Physics}, vol.~15, no.~12, pp. 1273--1278, 2019.

\bibitem{li2020quantum}
Y.~Li, R.-G. Zhou, R.~Xu, J.~Luo, and W.~Hu, ``A quantum deep convolutional
  neural network for image recognition,'' \emph{Quantum Science and
  Technology}, vol.~5, no.~4, p. 044003, 2020.

\bibitem{henderson2020quanvolutional}
M.~Henderson, S.~Shakya, S.~Pradhan, and T.~Cook, ``Quanvolutional neural
  networks: powering image recognition with quantum circuits,'' \emph{Quantum
  Machine Intelligence}, vol.~2, no.~1, p.~2, 2020.

\bibitem{hur2022quantum}
T.~Hur, L.~Kim, and D.~K. Park, ``Quantum convolutional neural network for
  classical data classification,'' \emph{Quantum Machine Intelligence}, vol.~4,
  no.~1, p.~3, 2022.

\bibitem{romero2017quantum}
J.~Romero, J.~P. Olson, and A.~Aspuru-Guzik, ``Quantum autoencoders for
  efficient compression of quantum data,'' \emph{Quantum Science and
  Technology}, vol.~2, no.~4, p. 045001, 2017.

\bibitem{srikumar2021clustering}
M.~Srikumar, C.~D. Hill, and L.~C. Hollenberg, ``Clustering and enhanced
  classification using a hybrid quantum autoencoder,'' \emph{Quantum Science
  and Technology}, vol.~7, no.~1, p. 015020, 2021.

\bibitem{lloyd2018quantum}
S.~Lloyd and C.~Weedbrook, ``Quantum generative adversarial learning,''
  \emph{Physical review letters}, vol. 121, no.~4, p. 040502, 2018.

\bibitem{dallaire2018quantum}
P.-L. Dallaire-Demers and N.~Killoran, ``Quantum generative adversarial
  networks,'' \emph{Physical Review A}, vol.~98, no.~1, p. 012324, 2018.

\bibitem{zeng2019learning}
J.~Zeng, Y.~Wu, J.-G. Liu, L.~Wang, and J.~Hu, ``Learning and inference on
  generative adversarial quantum circuits,'' \emph{Physical Review A}, vol.~99,
  no.~5, p. 052306, 2019.

\bibitem{benedetti2019generative}
M.~Benedetti, D.~Garcia-Pintos, O.~Perdomo, V.~Leyton-Ortega, Y.~Nam, and
  A.~Perdomo-Ortiz, ``A generative modeling approach for benchmarking and
  training shallow quantum circuits,'' \emph{npj Quantum Information}, vol.~5,
  no.~1, p.~45, 2019.

\bibitem{tian2023recent}
J.~Tian, X.~Sun, Y.~Du, S.~Zhao, Q.~Liu, K.~Zhang, W.~Yi, W.~Huang, C.~Wang,
  X.~Wu \emph{et~al.}, ``Recent advances for quantum neural networks in
  generative learning,'' \emph{IEEE Transactions on Pattern Analysis and
  Machine Intelligence}, 2023.

\bibitem{coyle2020born}
B.~Coyle, D.~Mills, V.~Danos, and E.~Kashefi, ``The born supremacy: quantum
  advantage and training of an ising born machine,'' \emph{npj Quantum
  Information}, vol.~6, no.~1, p.~60, 2020.

\bibitem{huang2021variational}
R.~Huang, X.~Tan, and Q.~Xu, ``Variational quantum tensor networks
  classifiers,'' \emph{Neurocomputing}, vol. 452, pp. 89--98, 2021.

\bibitem{guala2023practical}
D.~Guala, S.~Zhang, E.~Cruz, C.~A. Riofr{\'\i}o, J.~Klepsch, and J.~M.
  Arrazola, ``Practical overview of image classification with tensor-network
  quantum circuits,'' \emph{Scientific Reports}, vol.~13, no.~1, p. 4427, 2023.

\bibitem{haghshenas2022variational}
R.~Haghshenas, J.~Gray, A.~C. Potter, and G.~K.-L. Chan, ``Variational power of
  quantum circuit tensor networks,'' \emph{Physical Review X}, vol.~12, no.~1,
  p. 011047, 2022.

\bibitem{gili2022evaluating}
K.~Gili, M.~Mauri, and A.~Perdomo-Ortiz, ``Evaluating generalization in
  classical and quantum generative models,'' \emph{arXiv preprint
  arXiv:2201.08770}, 2022.

\bibitem{chen2020hybrid}
S.~Y.-C. Chen, C.-M. Huang, C.-W. Hsing, and Y.-J. Kao, ``Hybrid
  quantum-classical classifier based on tensor network and variational quantum
  circuit,'' \emph{arXiv preprint arXiv:2011.14651}, 2020.

\bibitem{coecke2010mathematical}
B.~Coecke, M.~Sadrzadeh, and S.~Clark, ``Mathematical foundations for a
  compositional distributional model of meaning,'' \emph{arXiv preprint
  arXiv:1003.4394}, 2010.

\bibitem{zeng2016quantum}
W.~Zeng and B.~Coecke, ``Quantum algorithms for compositional natural language
  processing,'' \emph{arXiv preprint arXiv:1608.01406}, 2016.

\bibitem{coecke2020foundations}
B.~Coecke, G.~de~Felice, K.~Meichanetzidis, and A.~Toumi, ``Foundations for
  near-term quantum natural language processing,'' \emph{arXiv preprint
  arXiv:2012.03755}, 2020.

\bibitem{meichanetzidis2020quantum}
K.~Meichanetzidis, S.~Gogioso, G.~De~Felice, N.~Chiappori, A.~Toumi, and
  B.~Coecke, ``Quantum natural language processing on near-term quantum
  computers,'' \emph{arXiv preprint arXiv:2005.04147}, 2020.

\bibitem{kartsaklis2021lambeq}
D.~Kartsaklis, I.~Fan, R.~Yeung, A.~Pearson, R.~Lorenz, A.~Toumi, G.~de~Felice,
  K.~Meichanetzidis, S.~Clark, and B.~Coecke, ``lambeq: An efficient high-level
  python library for quantum nlp,'' \emph{arXiv preprint arXiv:2110.04236},
  2021.

\bibitem{lorenz2023qnlp}
R.~Lorenz, A.~Pearson, K.~Meichanetzidis, D.~Kartsaklis, and B.~Coecke, ``Qnlp
  in practice: Running compositional models of meaning on a quantum computer,''
  \emph{Journal of Artificial Intelligence Research}, vol.~76, pp. 1305--1342,
  2023.

\bibitem{li2022quantum}
G.~Li, X.~Zhao, and X.~Wang, ``Quantum self-attention neural networks for text
  classification,'' \emph{arXiv preprint arXiv:2205.05625}, 2022.

\bibitem{sergioli2018quantum}
G.~Sergioli, E.~Santucci, L.~Didaci, J.~A. Miszczak, and R.~Giuntini, ``A
  quantum-inspired version of the nearest mean classifier,'' \emph{Soft
  Computing}, vol.~22, pp. 691--705, 2018.

\bibitem{sergioli2020quantum}
G.~Sergioli, ``Quantum and quantum-like machine learning: a note on differences
  and similarities,'' \emph{Soft Computing}, vol.~24, no.~14, pp.
  10\,247--10\,255, 2020.

\bibitem{sergioli2017quantum}
G.~Sergioli, G.~M. Bosyk, E.~Santucci, and R.~Giuntini, ``A quantum-inspired
  version of the classification problem,'' \emph{International Journal of
  Theoretical Physics}, vol.~56, pp. 3880--3888, 2017.

\bibitem{sergioli2019new}
G.~Sergioli, R.~Giuntini, and H.~Freytes, ``A new quantum approach to binary
  classification,'' \emph{PloS one}, vol.~14, no.~5, p. e0216224, 2019.

\bibitem{giuntini2023quantum}
R.~Giuntini, F.~Holik, D.~K. Park, H.~Freytes, C.~Blank, and G.~Sergioli,
  ``Quantum-inspired algorithm for direct multi-class classification,''
  \emph{Applied Soft Computing}, vol. 134, p. 109956, 2023.

\bibitem{sergioli2018quantumbiomedical}
G.~Sergioli, G.~Russo, E.~Santucci, A.~Stefano, S.~E. Torrisi, S.~Palmucci,
  C.~Vancheri, and R.~Giuntini, ``Quantum-inspired minimum distance
  classification in a biomedical context,'' \emph{International Journal of
  Quantum Information}, vol.~16, no.~08, p. 1840011, 2018.

\bibitem{sergioli2021quantum}
G.~Sergioli, C.~Militello, L.~Rundo, L.~Minafra, F.~Torrisi, G.~Russo, K.~L.
  Chow, and R.~Giuntini, ``A quantum-inspired classifier for clonogenic assay
  evaluations,'' \emph{Scientific Reports}, vol.~11, no.~1, p. 2830, 2021.

\bibitem{leporini2022efficient}
R.~Leporini and D.~Pastorello, ``An efficient geometric approach to
  quantum-inspired classifications,'' \emph{Scientific Reports}, vol.~12,
  no.~1, p. 8781, 2022.

\bibitem{bertini2023quantum}
C.~Bertini and R.~Leporini, ``Quantum-inspired applications for classification
  problems,'' \emph{Entropy}, vol.~25, no.~3, p. 404, 2023.

\bibitem{zhang2018end}
P.~Zhang, J.~Niu, Z.~Su, B.~Wang, L.~Ma, and D.~Song, ``End-to-end quantum-like
  language models with application to question answering,'' in
  \emph{Proceedings of the AAAI Conference on Artificial Intelligence},
  vol.~32, no.~1, 2018.

\bibitem{li2021quantumvideo}
Q.~Li, D.~Gkoumas, C.~Lioma, and M.~Melucci, ``Quantum-inspired multimodal
  fusion for video sentiment analysis,'' \emph{Information Fusion}, vol.~65,
  pp. 58--71, 2021.

\bibitem{zhang2018quantum}
Y.~Zhang, D.~Song, P.~Zhang, P.~Wang, J.~Li, X.~Li, and B.~Wang, ``A
  quantum-inspired multimodal sentiment analysis framework,'' \emph{Theoretical
  Computer Science}, vol. 752, pp. 21--40, 2018.

\bibitem{zhang2019quantum}
Y.~Zhang, D.~Song, P.~Zhang, X.~Li, and P.~Wang, ``A quantum-inspired sentiment
  representation model for twitter sentiment analysis,'' \emph{Applied
  Intelligence}, vol.~49, pp. 3093--3108, 2019.

\bibitem{zhang2020quantum}
Y.~Zhang, D.~Song, X.~Li, P.~Zhang, P.~Wang, L.~Rong, G.~Yu, and B.~Wang, ``A
  quantum-like multimodal network framework for modeling interaction dynamics
  in multiparty conversational sentiment analysis,'' \emph{Information Fusion},
  vol.~62, pp. 14--31, 2020.

\bibitem{li2019cnm}
Q.~Li, B.~Wang, and M.~Melucci, ``Cnm: An interpretable complex-valued network
  for matching,'' \emph{arXiv preprint arXiv:1904.05298}, 2019.

\bibitem{lowe2004distinctive}
D.~G. Lowe, ``Distinctive image features from scale-invariant keypoints,''
  \emph{International journal of computer vision}, vol.~60, pp. 91--110, 2004.

\bibitem{li2021quantumemotion}
Q.~Li, D.~Gkoumas, A.~Sordoni, J.-Y. Nie, and M.~Melucci, ``Quantum-inspired
  neural network for conversational emotion recognition,'' in \emph{Proceedings
  of the AAAI Conference on Artificial Intelligence}, vol.~35, no.~15, 2021,
  pp. 13\,270--13\,278.

\bibitem{pennington2014glove}
J.~Pennington, R.~Socher, and C.~D. Manning, ``Glove: Global vectors for word
  representation,'' in \emph{Proceedings of the 2014 conference on empirical
  methods in natural language processing (EMNLP)}, 2014, pp. 1532--1543.

\bibitem{devlin2018bert}
J.~Devlin, M.-W. Chang, K.~Lee, and K.~Toutanova, ``Bert: Pre-training of deep
  bidirectional transformers for language understanding,'' \emph{arXiv preprint
  arXiv:1810.04805}, 2018.

\bibitem{shi2021two}
J.~Shi, Z.~Li, W.~Lai, F.~Li, R.~Shi, Y.~Feng, and S.~Zhang, ``Two end-to-end
  quantum-inspired deep neural networks for text classification,'' \emph{IEEE
  Transactions on Knowledge and Data Engineering}, 2021.

\bibitem{altaisky2001quantum}
M.~Altaisky, ``Quantum neural network,'' \emph{arXiv preprint
  quant-ph/0107012}, 2001.

\bibitem{zhou2006quantum}
R.~Zhou, L.~Qin, and N.~Jiang, ``Quantum perceptron network,'' in
  \emph{Artificial Neural Networks--ICANN 2006: 16th International Conference,
  Athens, Greece, September 10-14, 2006. Proceedings, Part I 16}.\hskip 1em
  plus 0.5em minus 0.4em\relax Springer, 2006, pp. 651--657.

\bibitem{schuld2014quest}
M.~Schuld, I.~Sinayskiy, and F.~Petruccione, ``The quest for a quantum neural
  network,'' \emph{Quantum Information Processing}, vol.~13, pp. 2567--2586,
  2014.

\bibitem{patel2019novel}
O.~P. Patel, N.~Bharill, A.~Tiwari, and M.~Prasad, ``A novel quantum-inspired
  fuzzy based neural network for data classification,'' \emph{IEEE Transactions
  on Emerging Topics in Computing}, vol.~9, no.~2, pp. 1031--1044, 2019.

\bibitem{sagheer2019novel}
A.~Sagheer, M.~Zidan, and M.~M. Abdelsamea, ``A novel autonomous perceptron
  model for pattern classification applications,'' \emph{Entropy}, vol.~21,
  no.~8, p. 763, 2019.

\bibitem{UCI}
\BIBentryALTinterwordspacing
K.~Markelle, R.~Longjohn, and K.~Nottingham, ``Uci machine learning
  repository.'' [Online]. Available: \url{http://archive.ics.uci.edu/ml}
\BIBentrySTDinterwordspacing

\bibitem{konar2020quantum}
D.~Konar, S.~Bhattacharyya, T.~K. Gandhi, and B.~K. Panigrahi, ``A
  quantum-inspired self-supervised network model for automatic segmentation of
  brain mr images,'' \emph{Applied Soft Computing}, vol.~93, p. 106348, 2020.

\bibitem{zhang2022quantum}
J.~Zhang, Z.~Li, J.~Wang, Y.~Wang, S.~Hu, J.~Xiao, and Z.~Li, ``Quantum
  entanglement inspired correlation learning for classification,'' in
  \emph{Advances in Knowledge Discovery and Data Mining: 26th Pacific-Asia
  Conference, PAKDD 2022, Chengdu, China, May 16--19, 2022, Proceedings, Part
  II}.\hskip 1em plus 0.5em minus 0.4em\relax Springer, 2022, pp. 58--70.

\bibitem{tiwari2019towards}
P.~Tiwari and M.~Melucci, ``Towards a quantum-inspired binary classifier,''
  \emph{IEEE Access}, vol.~7, pp. 42\,354--42\,372, 2019.

\bibitem{tiwari2018towards}
------, ``Towards a quantum-inspired framework for binary classification,'' in
  \emph{Proceedings of the 27th ACM international conference on information and
  knowledge management}, 2018, pp. 1815--1818.

\bibitem{zhang2021interactive}
J.~Zhang, Z.~Li, R.~He, J.~Zhang, B.~Wang, Z.~Li, and T.~Niu, ``Interactive
  quantum classifier inspired by quantum open system theory,'' in \emph{2021
  International Joint Conference on Neural Networks (IJCNN)}.\hskip 1em plus
  0.5em minus 0.4em\relax IEEE, 2021, pp. 1--7.

\bibitem{tiwari2019binary}
P.~Tiwari and M.~Melucci, ``Binary classifier inspired by quantum theory,'' in
  \emph{Proceedings of the AAAI conference on artificial intelligence},
  vol.~33, no.~01, 2019, pp. 10\,051--10\,052.

\bibitem{rudolph2022generation}
M.~S. Rudolph, N.~B. Toussaint, A.~Katabarwa, S.~Johri, B.~Peropadre, and
  A.~Perdomo-Ortiz, ``Generation of high-resolution handwritten digits with an
  ion-trap quantum computer,'' \emph{Physical Review X}, vol.~12, no.~3, p.
  031010, 2022.

\bibitem{tsang2022hybrid}
S.~L. Tsang, M.~T. West, S.~M. Erfani, and M.~Usman, ``Hybrid quantum-classical
  generative adversarial network for high resolution image generation,''
  \emph{arXiv preprint arXiv:2212.11614}, 2022.

\bibitem{zhou2023hybrid}
N.-R. Zhou, T.-F. Zhang, X.-W. Xie, and J.-Y. Wu, ``Hybrid quantum--classical
  generative adversarial networks for image generation via learning discrete
  distribution,'' \emph{Signal Processing: Image Communication}, vol. 110, p.
  116891, 2023.

\bibitem{selvan2020locally}
R.~Selvan, S.~{\O}rting, and E.~B. Dam, ``Locally orderless tensor networks for
  classifying two-and three-dimensional medical images,'' \emph{arXiv preprint
  arXiv:2009.12280}, 2020.

\bibitem{pomarico2021proposal}
D.~Pomarico, A.~Fanizzi, N.~Amoroso, R.~Bellotti, A.~Biafora, S.~Bove,
  V.~Didonna, D.~L. Forgia, M.~I. Pastena, P.~Tamborra \emph{et~al.}, ``A
  proposal of quantum-inspired machine learning for medical purposes: An
  application case,'' \emph{Mathematics}, vol.~9, no.~4, p. 410, 2021.

\bibitem{azevedo2022quantum}
V.~Azevedo, C.~Silva, and I.~Dutra, ``Quantum transfer learning for breast
  cancer detection,'' \emph{Quantum Machine Intelligence}, vol.~4, no.~1, p.~5,
  2022.

\bibitem{esposito2022quantum}
M.~Esposito, G.~Uehara, and A.~Spanias, ``Quantum machine learning for audio
  classification with applications to healthcare,'' in \emph{2022 13th
  International Conference on Information, Intelligence, Systems \&
  Applications (IISA)}.\hskip 1em plus 0.5em minus 0.4em\relax IEEE, 2022, pp.
  1--4.

\bibitem{sakuma2020application}
T.~Sakuma, ``Application of deep quantum neural networks to finance,''
  \emph{arXiv preprint arXiv:2011.07319}, 2020.

\bibitem{ganguly2023implementing}
S.~Ganguly, ``Implementing quantum generative adversarial network (qgan) and
  qcbm in finance,'' \emph{arXiv preprint arXiv:2308.08448}, 2023.

\bibitem{arrazola2019quantum}
J.~M. Arrazola, A.~Delgado, B.~R. Bardhan, and S.~Lloyd, ``Quantum-inspired
  algorithms in practice,'' \emph{arXiv preprint arXiv:1905.10415}, 2019.

\bibitem{mugel2022dynamic}
S.~Mugel, C.~Kuchkovsky, E.~Sanchez, S.~Fernandez-Lorenzo, J.~Luis-Hita,
  E.~Lizaso, and R.~Orus, ``Dynamic portfolio optimization with real datasets
  using quantum processors and quantum-inspired tensor networks,''
  \emph{Physical Review Research}, vol.~4, no.~1, p. 013006, 2022.

\bibitem{coyle2021quantum}
B.~Coyle, M.~Henderson, J.~C.~J. Le, N.~Kumar, M.~Paini, and E.~Kashefi,
  ``Quantum versus classical generative modelling in finance,'' \emph{Quantum
  Science and Technology}, vol.~6, no.~2, p. 024013, 2021.

\bibitem{kondratyev2021non}
A.~Kondratyev, ``Non-differentiable leaning of quantum circuit born machine
  with genetic algorithm,'' \emph{Wilmott}, vol. 2021, no. 114, pp. 50--61,
  2021.

\bibitem{wu2021applicationvqc}
S.~L. Wu, J.~Chan, W.~Guan, S.~Sun, A.~Wang, C.~Zhou, M.~Livny, F.~Carminati,
  A.~Di~Meglio, A.~C. Li \emph{et~al.}, ``Application of quantum machine
  learning using the quantum variational classifier method to high energy
  physics analysis at the lhc on ibm quantum computer simulator and hardware
  with 10 qubits,'' \emph{Journal of Physics G: Nuclear and Particle Physics},
  vol.~48, no.~12, p. 125003, 2021.

\bibitem{wu2021applicationqsvm}
S.~L. Wu, S.~Sun, W.~Guan, C.~Zhou, J.~Chan, C.~L. Cheng, T.~Pham, Y.~Qian,
  A.~Z. Wang, R.~Zhang \emph{et~al.}, ``Application of quantum machine learning
  using the quantum kernel algorithm on high energy physics analysis at the
  lhc,'' \emph{Physical Review Research}, vol.~3, no.~3, p. 033221, 2021.

\bibitem{ngairangbam2022anomaly}
V.~S. Ngairangbam, M.~Spannowsky, and M.~Takeuchi, ``Anomaly detection in
  high-energy physics using a quantum autoencoder,'' \emph{Physical Review D},
  vol. 105, no.~9, p. 095004, 2022.

\bibitem{moussa2023application}
C.~Moussa, H.~Wang, M.~Araya-Polo, T.~B{\"a}ck, and V.~Dunjko, ``Application of
  quantum-inspired generative models to small molecular datasets,'' \emph{arXiv
  preprint arXiv:2304.10867}, 2023.

\bibitem{suryotrisongko2022evaluating}
H.~Suryotrisongko and Y.~Musashi, ``Evaluating hybrid quantum-classical deep
  learning for cybersecurity botnet dga detection,'' \emph{Procedia Computer
  Science}, vol. 197, pp. 223--229, 2022.

\bibitem{gong2022network}
C.~Gong, W.~Guan, A.~Gani, and H.~Qi, ``Network attack detection scheme based
  on variational quantum neural network,'' \emph{The Journal of
  Supercomputing}, vol.~78, no.~15, pp. 16\,876--16\,897, 2022.

\bibitem{herr2021anomaly}
D.~Herr, B.~Obert, and M.~Rosenkranz, ``Anomaly detection with variational
  quantum generative adversarial networks,'' \emph{Quantum Science and
  Technology}, vol.~6, no.~4, p. 045004, 2021.

\bibitem{yang2021decentralizing}
C.-H.~H. Yang, J.~Qi, S.~Y.-C. Chen, P.-Y. Chen, S.~M. Siniscalchi, X.~Ma, and
  C.-H. Lee, ``Decentralizing feature extraction with quantum convolutional
  neural network for automatic speech recognition,'' in \emph{ICASSP 2021-2021
  IEEE International Conference on Acoustics, Speech and Signal Processing
  (ICASSP)}.\hskip 1em plus 0.5em minus 0.4em\relax IEEE, 2021, pp. 6523--6527.

\bibitem{garcia2021quantum}
J.~J. Garc{\'\i}a-Ripoll, ``Quantum-inspired algorithms for multivariate
  analysis: from interpolation to partial differential equations,''
  \emph{Quantum}, vol.~5, p. 431, 2021.

\bibitem{an2020ensemble}
S.~An, M.~Lee, S.~Park, H.~Yang, and J.~So, ``An ensemble of simple
  convolutional neural network models for mnist digit recognition,'' 2020.

\bibitem{tanveer2021fine}
M.~S. Tanveer, M.~U.~K. Khan, and C.-M. Kyung, ``Fine-tuning darts for image
  classification,'' in \emph{2020 25th International Conference on Pattern
  Recognition (ICPR)}.\hskip 1em plus 0.5em minus 0.4em\relax IEEE, 2021, pp.
  4789--4796.

\bibitem{sinha2021consistency}
S.~Sinha and A.~B. Dieng, ``Consistency regularization for variational
  auto-encoders,'' \emph{Advances in Neural Information Processing Systems},
  vol.~34, pp. 12\,943--12\,954, 2021.

\bibitem{deng2009imagenet}
J.~Deng, W.~Dong, R.~Socher, L.-J. Li, K.~Li, and L.~Fei-Fei, ``Imagenet: A
  large-scale hierarchical image database,'' in \emph{2009 IEEE conference on
  computer vision and pattern recognition}.\hskip 1em plus 0.5em minus
  0.4em\relax Ieee, 2009, pp. 248--255.

\bibitem{muguli2021dicova}
A.~Muguli, L.~Pinto, N.~Sharma, P.~Krishnan, P.~K. Ghosh, R.~Kumar, S.~Bhat,
  S.~R. Chetupalli, S.~Ganapathy, S.~Ramoji \emph{et~al.}, ``Dicova challenge:
  Dataset, task, and baseline system for covid-19 diagnosis using acoustics,''
  \emph{arXiv preprint arXiv:2103.09148}, 2021.

\bibitem{orlandic2021coughvid}
L.~Orlandic, T.~Teijeiro, and D.~Atienza, ``The coughvid crowdsourcing dataset,
  a corpus for the study of large-scale cough analysis algorithms,''
  \emph{Scientific Data}, vol.~8, no.~1, p. 156, 2021.

\bibitem{suzuki2021qulacs}
Y.~Suzuki, Y.~Kawase, Y.~Masumura, Y.~Hiraga, M.~Nakadai, J.~Chen, K.~M.
  Nakanishi, K.~Mitarai, R.~Imai, S.~Tamiya \emph{et~al.}, ``Qulacs: a fast and
  versatile quantum circuit simulator for research purpose,'' \emph{Quantum},
  vol.~5, p. 559, 2021.

\bibitem{IBMQuantum}
\BIBentryALTinterwordspacing
IBM, ``Ibm quantum,'' 2021, accessed: August 6, 2023. [Online]. Available:
  \url{https://quantum-computing.ibm.com/}
\BIBentrySTDinterwordspacing

\bibitem{google2023quantumai}
\BIBentryALTinterwordspacing
Google, ``Google quantum ai,'' 2023, accessed: July 31, 2023. [Online].
  Available: \url{https://quantumai.google/}
\BIBentrySTDinterwordspacing

\bibitem{amazonbraket}
\BIBentryALTinterwordspacing
A.~W. Services, ``Amazon braket,'' 2020, accessed: August 6, 2023. [Online].
  Available: \url{https://aws.amazon.com/braket/}
\BIBentrySTDinterwordspacing

\bibitem{ruddigkeit2012enumeration}
L.~Ruddigkeit, R.~Van~Deursen, L.~C. Blum, and J.-L. Reymond, ``Enumeration of
  166 billion organic small molecules in the chemical universe database
  gdb-17,'' \emph{Journal of chemical information and modeling}, vol.~52,
  no.~11, pp. 2864--2875, 2012.

\bibitem{ramakrishnan2014quantum}
R.~Ramakrishnan, P.~O. Dral, M.~Rupp, and O.~A. Von~Lilienfeld, ``Quantum
  chemistry structures and properties of 134 kilo molecules,'' \emph{Scientific
  data}, vol.~1, no.~1, pp. 1--7, 2014.

\bibitem{li2019variational}
Y.~Li, J.~Hu, X.-M. Zhang, Z.~Song, and M.-H. Yung, ``Variational quantum
  simulation for quantum chemistry,'' \emph{Advanced Theory and Simulations},
  vol.~2, no.~4, p. 1800182, 2019.

\bibitem{bauer2020quantum}
B.~Bauer, S.~Bravyi, M.~Motta, and G.~K.-L. Chan, ``Quantum algorithms for
  quantum chemistry and quantum materials science,'' \emph{Chemical Reviews},
  vol. 120, no.~22, pp. 12\,685--12\,717, 2020.

\bibitem{oh2022quantum}
C.~Oh, Y.~Lim, Y.~Wong, B.~Fefferman, and L.~Jiang, ``Quantum-inspired
  classical algorithm for molecular vibronic spectra,'' \emph{arXiv preprint
  arXiv:2202.01861}, 2022.

\bibitem{oh2023quantum}
C.~Oh, L.~Jiang, and N.~Quesada, ``Quantum-inspired classical algorithm for
  graph problems by gaussian boson sampling,'' \emph{arXiv preprint
  arXiv:2302.00536}, 2023.

\bibitem{filip2020stochastic}
M.-A. Filip and A.~J. Thom, ``A stochastic approach to unitary coupled
  cluster,'' \emph{The Journal of Chemical Physics}, vol. 153, no.~21, 2020.

\bibitem{chen2021quantum}
J.~Chen, H.-P. Cheng, and J.~K. Freericks, ``Quantum-inspired algorithm for the
  factorized form of unitary coupled cluster theory,'' \emph{Journal of
  Chemical Theory and Computation}, vol.~17, no.~2, pp. 841--847, 2021.

\bibitem{sharafaldin2019developing}
I.~Sharafaldin, A.~H. Lashkari, S.~Hakak, and A.~A. Ghorbani, ``Developing
  realistic distributed denial of service (ddos) attack dataset and taxonomy,''
  in \emph{2019 International Carnahan Conference on Security Technology
  (ICCST)}.\hskip 1em plus 0.5em minus 0.4em\relax IEEE, 2019, pp. 1--8.

\bibitem{schlegl2017unsupervised}
T.~Schlegl, P.~Seeb{\"o}ck, S.~M. Waldstein, U.~Schmidt-Erfurth, and G.~Langs,
  ``Unsupervised anomaly detection with generative adversarial networks to
  guide marker discovery,'' in \emph{International conference on information
  processing in medical imaging}.\hskip 1em plus 0.5em minus 0.4em\relax
  Springer, 2017, pp. 146--157.

\bibitem{Olson2017PMLB}
\BIBentryALTinterwordspacing
R.~S. Olson, W.~La~Cava, P.~Orzechowski, R.~J. Urbanowicz, and J.~H. Moore,
  ``Pmlb: a large benchmark suite for machine learning evaluation and
  comparison,'' \emph{BioData Mining}, vol.~10, no.~1, p.~36, Dec 2017.
  [Online]. Available: \url{https://doi.org/10.1186/s13040-017-0154-4}
\BIBentrySTDinterwordspacing

\bibitem{evenbly2022number}
G.~Evenbly, ``Number-state preserving tensor networks as classifiers for
  supervised learning,'' \emph{Frontiers in Physics}, vol.~10, p. 1146, 2022.

\bibitem{aaronson2016complexity}
S.~Aaronson and L.~Chen, ``Complexity-theoretic foundations of quantum
  supremacy experiments,'' \emph{arXiv preprint arXiv:1612.05903}, 2016.

\bibitem{kerenidis2019q}
I.~Kerenidis, J.~Landman, A.~Luongo, and A.~Prakash, ``q-means: A quantum
  algorithm for unsupervised machine learning,'' \emph{Advances in neural
  information processing systems}, vol.~32, 2019.

\bibitem{di2022dawn}
R.~Di~Sipio, J.-H. Huang, S.~Y.-C. Chen, S.~Mangini, and M.~Worring, ``The dawn
  of quantum natural language processing,'' in \emph{ICASSP 2022-2022 IEEE
  International Conference on Acoustics, Speech and Signal Processing
  (ICASSP)}.\hskip 1em plus 0.5em minus 0.4em\relax IEEE, 2022, pp. 8612--8616.

\bibitem{cichocki2016low}
A.~Cichocki, N.~Lee, I.~V. Oseledets, A.-H. Phan, Q.~Zhao, and D.~Mandic,
  ``Low-rank tensor networks for dimensionality reduction and large-scale
  optimization problems: Perspectives and challenges part 1,'' \emph{arXiv
  preprint arXiv:1609.00893}, 2016.

\bibitem{carleo2018constructing}
G.~Carleo, Y.~Nomura, and M.~Imada, ``Constructing exact representations of
  quantum many-body systems with deep neural networks,'' \emph{Nature
  communications}, vol.~9, no.~1, p. 5322, 2018.

\bibitem{cai2018approximating}
Z.~Cai and J.~Liu, ``Approximating quantum many-body wave functions using
  artificial neural networks,'' \emph{Physical Review B}, vol.~97, no.~3, p.
  035116, 2018.

\bibitem{lazzarin2022multi}
M.~Lazzarin, D.~E. Galli, and E.~Prati, ``Multi-class quantum classifiers with
  tensor network circuits for quantum phase recognition,'' \emph{Physics
  Letters A}, vol. 434, p. 128056, 2022.

\bibitem{liu2018differentiable}
J.-G. Liu and L.~Wang, ``Differentiable learning of quantum circuit born
  machines,'' \emph{Physical Review A}, vol.~98, no.~6, p. 062324, 2018.

\bibitem{gong2022born}
L.-H. Gong, L.-Z. Xiang, S.-H. Liu, and N.-R. Zhou, ``Born machine model based
  on matrix product state quantum circuit,'' \emph{Physica A: Statistical
  Mechanics and its Applications}, vol. 593, p. 126907, 2022.

\bibitem{liqui}
\BIBentryALTinterwordspacing
D.~Wecker and K.~M. Svore, ``{LIQU}i|>: {A} {S}oftware {D}esign {A}rchitecture
  and {D}omain-{S}pecific {L}anguage for {Q}uantum {C}omputing,'' 2014.
  [Online]. Available: \url{arXiv:1402.4467v1}
\BIBentrySTDinterwordspacing

\bibitem{pfeifer2014ncon}
R.~N. Pfeifer, G.~Evenbly, S.~Singh, and G.~Vidal, ``Ncon: A tensor network
  contractor for matlab,'' \emph{arXiv preprint arXiv:1402.0939}, 2014.

\bibitem{dolgov2014computation}
S.~V. Dolgov, B.~N. Khoromskij, I.~V. Oseledets, and D.~V. Savostyanov,
  ``Computation of extreme eigenvalues in higher dimensions using block tensor
  train format,'' \emph{Computer Physics Communications}, vol. 185, no.~4, pp.
  1207--1216, 2014.

\bibitem{kossaifi2016tensorly}
J.~Kossaifi, Y.~Panagakis, A.~Anandkumar, and M.~Pantic, ``Tensorly: Tensor
  learning in python,'' \emph{arXiv preprint arXiv:1610.09555}, 2016.

\bibitem{Cirq2023}
\BIBentryALTinterwordspacing
C.~Developers, ``Cirq,'' 2023. [Online]. Available:
  \url{https://doi.org/10.5281/zenodo.8161252}
\BIBentrySTDinterwordspacing

\bibitem{scikittt}
P.~Gel, M.~Lücke, T.~Bake, F.~Nüske, and M.~Scherer, ``Scikit-tt tensor train
  toolbox,'' \url{https://github.com/PGelss/scikit_tt}, 2018.

\bibitem{roberts2019tensornetwork}
C.~Roberts, A.~Milsted, M.~Ganahl, A.~Zalcman, B.~Fontaine, Y.~Zou, J.~Hidary,
  G.~Vidal, and S.~Leichenauer, ``Tensornetwork: A library for physics and
  machine learning,'' 2019.

\bibitem{torchmps}
J.~Miller, ``Torchmps,'' \url{https://github.com/jemisjoky/torchmps}, 2019.

\bibitem{guerreschi2020intel}
G.~G. Guerreschi, J.~Hogaboam, F.~Baruffa, and N.~P. Sawaya, ``Intel quantum
  simulator: A cloud-ready high-performance simulator of quantum circuits,''
  \emph{Quantum Science and Technology}, vol.~5, no.~3, p. 034007, 2020.

\bibitem{pastaq}
\BIBentryALTinterwordspacing
G.~Torlai and M.~Fishman, ``\mbox{PastaQ}: A package for simulation, tomography
  and analysis of quantum computers,'' 2020. [Online]. Available:
  \url{https://github.com/GTorlai/PastaQ.jl/}
\BIBentrySTDinterwordspacing

\bibitem{broughton2020tensorflow}
M.~Broughton, G.~Verdon, T.~McCourt, A.~J. Martinez, J.~H. Yoo, S.~V. Isakov,
  P.~Massey, R.~Halavati, M.~Y. Niu, A.~Zlokapa \emph{et~al.}, ``Tensorflow
  quantum: A software framework for quantum machine learning,'' \emph{arXiv
  preprint arXiv:2003.02989}, 2020.

\bibitem{qibo_paper}
\BIBentryALTinterwordspacing
S.~Efthymiou, S.~Ramos-Calderer, C.~Bravo-Prieto, A.~P{\'{e}}rez-Salinas,
  D.~Garc{\'{\i}}a-Mart{\'{\i}}n, A.~Garcia-Saez, J.~I. Latorre, and
  S.~Carrazza, ``Qibo: a framework for quantum simulation with hardware
  acceleration,'' \emph{Quantum Science and Technology}, vol.~7, no.~1, p.
  015018, dec 2021. [Online]. Available:
  \url{https://doi.org/10.1088/2058-9565/ac39f5}
\BIBentrySTDinterwordspacing

\bibitem{fishman2022itensor}
M.~Fishman, S.~White, and E.~Stoudenmire, ``The itensor software library for
  tensor network calculations,'' \emph{SciPost Physics Codebases}, p. 004,
  2022.

\bibitem{UBS:22}
\BIBentryALTinterwordspacing
M.~Usvyatsov, R.~Ballester-Ripoll, and K.~Schindler, ``tntorch: Tensor network
  learning with {PyTorch},'' \emph{Journal of Machine Learning Research},
  vol.~23, no. 208, pp. 1--6, 2022. [Online]. Available:
  \url{http://jmlr.org/papers/v23/21-1197.html}
\BIBentrySTDinterwordspacing

\bibitem{harris2020array}
\BIBentryALTinterwordspacing
C.~R. Harris, K.~J. Millman, S.~J. van~der Walt, R.~Gommers, P.~Virtanen,
  D.~Cournapeau, E.~Wieser, J.~Taylor, S.~Berg, N.~J. Smith, R.~Kern, M.~Picus,
  S.~Hoyer, M.~H. van Kerkwijk, M.~Brett, A.~Haldane, J.~F. del R{\'{i}}o,
  M.~Wiebe, P.~Peterson, P.~G{\'{e}}rard-Marchant, K.~Sheppard, T.~Reddy,
  W.~Weckesser, H.~Abbasi, C.~Gohlke, and T.~E. Oliphant, ``Array programming
  with {NumPy},'' \emph{Nature}, vol. 585, no. 7825, pp. 357--362, Sep. 2020.
  [Online]. Available: \url{https://doi.org/10.1038/s41586-020-2649-2}
\BIBentrySTDinterwordspacing

\bibitem{psarras2021landscape}
C.~Psarras, L.~Karlsson, J.~Li, and P.~Bientinesi, ``The landscape of software
  for tensor computations,'' \emph{arXiv preprint arXiv:2103.13756}, 2021.

\bibitem{microsoft2023azurequantum}
\BIBentryALTinterwordspacing
Microsoft, ``Azure quantum,'' 2023, accessed: August 6, 2023. [Online].
  Available: \url{https://azure.microsoft.com/en-au/products/quantum}
\BIBentrySTDinterwordspacing

\bibitem{qBraid}
\BIBentryALTinterwordspacing
qBraid, ``qbraid,'' 2023, accessed: September 2, 2023. [Online]. Available:
  \url{https://www.qbraid.com/}
\BIBentrySTDinterwordspacing

\bibitem{qutech}
\BIBentryALTinterwordspacing
QuTech, ``Quantum inspire home,'' 2023, accessed: September 2, 2023. [Online].
  Available: \url{https://www.quantum-inspire.com/}
\BIBentrySTDinterwordspacing

\bibitem{Orquestra}
\BIBentryALTinterwordspacing
Z.~Computing, ``Orquestra.io,'' 2023, accessed: September 2, 2023. [Online].
  Available: \url{https://www.orquestra.io/}
\BIBentrySTDinterwordspacing

\bibitem{sao2019application}
M.~Sao, H.~Watanabe, Y.~Musha, and A.~Utsunomiya, ``Application of digital
  annealer for faster combinatorial optimization,'' \emph{Fujitsu Scientific
  and Technical Journal}, vol.~55, no.~2, pp. 45--51, 2019.

\bibitem{komatsu2021performance}
K.~Komatsu, A.~Onodera, E.~Focht, S.~Fujimoto, Y.~Isobe, S.~Momose, M.~Sato,
  and H.~Kobayashi, ``Performance and power analysis of a vector computing
  system,'' \emph{Supercomputing Frontiers and Innovations}, vol.~8, no.~2, pp.
  75--94, 2021.

\bibitem{hu2019quantum}
F.~Hu, B.-N. Wang, N.~Wang, and C.~Wang, ``Quantum machine learning with d-wave
  quantum computer,'' \emph{Quantum Engineering}, vol.~1, no.~2, p. e12, 2019.

\bibitem{kockum2014quantum}
A.~F. Kockum, \emph{Quantum optics with artificial atoms}.\hskip 1em plus 0.5em
  minus 0.4em\relax Chalmers Tekniska Hogskola (Sweden), 2014.

\bibitem{dawson2005solovay}
C.~M. Dawson and M.~A. Nielsen, ``The solovay-kitaev algorithm,'' \emph{arXiv
  preprint quant-ph/0505030}, 2005.

\bibitem{iten2016quantum}
R.~Iten, R.~Colbeck, I.~Kukuljan, J.~Home, and M.~Christandl, ``Quantum
  circuits for isometries,'' \emph{Physical Review A}, vol.~93, no.~3, p.
  032318, 2016.

\bibitem{vartiainen2004efficient}
J.~J. Vartiainen, M.~M{\"o}tt{\"o}nen, and M.~M. Salomaa, ``Efficient
  decomposition of quantum gates,'' \emph{Physical review letters}, vol.~92,
  no.~17, p. 177902, 2004.

\bibitem{glorot2010understanding}
X.~Glorot and Y.~Bengio, ``Understanding the difficulty of training deep
  feedforward neural networks,'' in \emph{Proceedings of the thirteenth
  international conference on artificial intelligence and statistics}.\hskip
  1em plus 0.5em minus 0.4em\relax JMLR Workshop and Conference Proceedings,
  2010, pp. 249--256.

\end{thebibliography}



\appendices
\section{Background}
\label{sec:background}


In this section, we provide an overview of the fundamental concepts in both quantum computing and machine learning, focusing on topics essential for understanding QiML. The reader is not expected to be already equipped with these concepts, however, a rudimentary understanding of linear algebra is presumed for ease of comprehension.

\subsection{Quantum Computing}

\subsubsection{Dirac Notation}

Dirac notation, also known as bra-ket notation, is a compact representation commonly used in quantum mechanics and quantum computing to represent quantum states, as well as perform calculations involving these states. It has since become the standard notation for working with quantum systems.

In Dirac notation, quantum states are represented by kets, denoted by $\vert\psi\rangle$, where $\psi$ is the name of the state. A quantum state $\vert\psi\rangle$ can be represented as a column vector in a vector space:

\begin{align}
\vert\psi\rangle = \begin{bmatrix} \psi_0 \\ \psi_1 \\ \cdots \\ \psi_n \end{bmatrix}
\end{align}
where $\psi_0, \psi_1, \dots , \psi_n  \in \mathbb{C}$; $\vert\psi\rangle \in \mathbb{C}^n$. The dual states, or bras, are represented as $\langle\psi\vert$ and are the complex conjugate of the transpose of the corresponding kets, denoted by the $\dagger$ symbol. For example, $\vert\psi\rangle \in \mathbb{C}^2$:

\begin{align}
\langle\psi\vert = \vert\psi\rangle^\dagger = \begin{bmatrix} \psi_0 \\ \psi_1 \end{bmatrix}^\dagger = \begin{bmatrix} \psi_0^* \ \psi_1^* \end{bmatrix},
\end{align}
where (*) denotes the complex conjugate. In quantum computing, the computational basis is a set of basis states that forms an orthonormal basis for the state space of a quantum system. For a single qubit system, the computational basis consists of two basis states: $\vert0\rangle$ and $\vert1\rangle$. These basis states can be represented as column vectors:

\begin{align}
\vert0\rangle = \begin{bmatrix} 1 \\ 0 \end{bmatrix}, \quad \vert1\rangle = \begin{bmatrix} 0 \\ 1 \end{bmatrix}
\end{align}
A general single-qubit quantum state can be represented as a linear combination of these basis states:

\begin{align}
\vert\psi\rangle = \alpha \vert0\rangle + \beta \vert1\rangle
\end{align}
where $\alpha$ and $\beta$ are complex numbers, and the normalization condition $|\alpha|^2 + |\beta|^2 = 1$ must be satisfied to maintain the state as a valid quantum state.

For a multi-qubit system with $n$ qubits, the computational basis is formed by taking the tensor product of the single-qubit basis states, resulting in $2^n$ basis states. The basis states for an $n$-qubit system are represented as $\vert i_1 i_2 \cdots i_n \rangle$, where $i_k \in {0, 1}$ for each $k = 1, 2, \dots, n$. For example, the computational basis for a two-qubit system consists of the following four basis states:

\begin{align}
\vert00\rangle, \quad \vert01\rangle, \quad \vert10\rangle, \quad \vert11\rangle
\end{align}
A general $n$-qubit quantum state can be represented as a linear combination of these computational basis states:

\begin{align}
    \ket{\psi} &= \sum_{i=0}^{2^n - 1} \alpha_i | i_1 i_2 \cdots i_n \rangle \\
    & i_k \in \{0, 1\}; \, k = 1, 2, \dots, n.
\end{align}
The index \( i \) can be thought of as the binary representation \( i_1 i_2 \cdots i_n \). $\alpha_i$ are complex coefficients, and the normalization condition $\sum_{n} |\alpha_i|^2 = 1$ must be satisfied.

\subsubsection{Inner Product}

The inner product, or scalar product, of two quantum states $\vert\psi\rangle$ and $\vert\phi\rangle$ is denoted as $\langle\psi\vert\phi\rangle$. The inner product can be calculated as the product of the corresponding bra and ket:

\begin{align}
\langle\psi\vert\phi\rangle = \sum_{i} \psi_i^* \phi_i,
\end{align}
where $\psi_i^*$ and $\phi_i$ are the $i$-th components of the states $\vert\psi\rangle$ and $\vert\phi\rangle$, respectively. The result of the inner product is a complex scalar that carries information about the overlap between the two states. In particular, the squared magnitude of the inner product, $|\langle\psi\vert\phi\rangle|^2$, represents the probability that the state $\vert\phi\rangle$ will be found in the state $\vert\psi\rangle$ upon measurement.

\subsubsection{Outer Product}

The outer product of two quantum states $\vert\psi\rangle$ and $\vert\phi\rangle$, denoted as $\vert\psi\rangle\langle\phi\vert$, produces a linear operator that acts on the Hilbert space where the quantum system that defines $\vert\psi\rangle$ and $\vert\phi\rangle$ resides. For $\ket{\psi} \in \mathbb{C}^{n}$ and $\ket{\phi} \in \mathbb{C}^{n}$, their outer product is:

\begin{equation}
\vert\psi\rangle\langle\phi\vert = \begin{bmatrix} \psi_0 \phi_0^* & \psi_0 \phi_1^* & \cdots & \psi_0 \phi_{n-1}^* \\ \psi_1 \phi_0^* & \psi_1 \phi_1^* & \cdots & \psi_1 \phi_{n-1}^* \\ \vdots & \vdots & \ddots & \vdots \\ \psi_{n-1} \phi_0^* & \psi_{n-1} \phi_1^* & \cdots & \psi_{n-1} \phi_{n-1}^* \end{bmatrix}
\end{equation}

This linear operator can be used to calculate the projection of one state onto another. A projection matrix is formed by taking the outer products of the basis states of a subspace and summing them up. Given a basis ${ \vert u_i \rangle }$ for an $m$-dimensional Hilbert space, the projection matrix $P$ is defined as:

\begin{align}
P = \sum_{i=1}^{m} \vert u_i \rangle \langle u_i \vert
\end{align}
Each term $\vert u_i \rangle \langle u_i \vert$ is an outer product, a “projector”, and the sum of these outer products forms the projection matrix $P$. $P$ inherits the idempotent ($P^2 = P$) and Hermitian ($P^\dagger = P$) properties from the outer products. When applied to a vector $\vert \psi \rangle$, the projection matrix $P$ results in the orthogonal projection $\vert \phi \rangle$ onto the subspace spanned by the basis states:

\begin{align}
\vert \phi \rangle = P \vert \psi \rangle = \sum_{i=1}^{m} \vert u_i \rangle \langle u_i \vert \psi \rangle
\end{align}
These projection matrices are essential in performing measurement. 

The outer product also allows for the convenient representation of quantum gates and transformations. For example, consider the unitary matrix $\left(\begin{smallmatrix} 0 & 1 \\ 1 & 0 \end{smallmatrix}\right)$ (The Pauli-X gate $\sigma_x$). This matrix can be considered as sum of its constituent rank 1 matrices, expressed using outer products:

\begin{equation}
\begin{split}
\vert0\rangle\langle1\vert + \vert1\rangle\langle0\vert & = \begin{bmatrix} 1 \\ 0 \end{bmatrix} \begin{bmatrix} 0 & 1 \end{bmatrix} + \begin{bmatrix} 0 \\ 1 \end{bmatrix} \begin{bmatrix} 1 & 0 \end{bmatrix} 
\\ & = \begin{bmatrix} 0 & 1 \\ 0 & 0 \end{bmatrix} + \begin{bmatrix} 0 & 0 \\ 1 & 0 \end{bmatrix} 
\\ & = \begin{bmatrix} 0 & 1 \\ 1 & 0 \end{bmatrix} \\ 
& = \sigma_x
\end{split}
\end{equation}

\subsubsection{Tensor Product}

The tensor product linearly combines two quantum states $\vert\psi\rangle$ and $\vert\phi\rangle$ to form a new composite state denoted as $\vert\psi\rangle\otimes\vert\phi\rangle$ or simply as either $\vert\psi\rangle\vert\phi\rangle$ or $\vert\psi\phi\rangle$. The tensor product is used extensively in quantum computing to describe multi-qubit systems. For instance, the tensor product of two qubits ($n = 2$) in states $\vert\psi\rangle$ and $\vert\phi\rangle$ produces a column vector of length $2^n$, written as:

\begin{equation}
\begin{split}
\vert\psi\rangle\otimes\vert\phi\rangle & = \vert\psi\rangle\vert\phi\rangle = \vert\psi\phi\rangle
\\
& = \begin{bmatrix} \alpha\ket{0} +\ \beta\ket{1} \end{bmatrix} \otimes \begin{bmatrix} \gamma\ket{0} +\ \delta\ket{1} \end{bmatrix}  \\
& = \alpha\gamma\ket{00} + \alpha\delta\ket{01} + \beta\gamma\ket{10} + \beta\delta\ket{11}
\end{split}
\end{equation}
where $|\alpha\gamma|^2 + |\alpha\delta|^2 + |\beta\gamma|^2 + |\beta\delta|^2 = 1$.
In general, the tensor product is:


\begin{itemize}
    \item not commutative: $\vert\psi\phi\rangle \neq \vert\phi\psi\rangle$
    \item associative: $(\vert\psi\rangle\otimes\vert\phi\rangle)\otimes\vert\chi\rangle = \vert\psi\rangle\otimes(\vert\phi\rangle\otimes\vert\chi\rangle)$
\end{itemize}
The tensor product can also be extended to operators, which is essential for when dealing with multi-qubit gates. For example, given the two single-qubit operators $A$ and $B$, their tensor product $A\otimes B$ will act on a two-qubit state $\vert\psi\phi\rangle$ as follows:

\begin{align}
(A\otimes B)\vert\psi\phi\rangle = (A\vert\psi\rangle)\otimes(B\vert\phi\rangle).
\end{align}

\subsubsection{Qubits and Quantum States}

Quantum systems are represented by quantum states, which are described by vectors in a complex Hilbert space. The most basic quantum system is the qubit, the quantum counterpart of the classical bit. A qubit is represented by a linear combination of basis states $\ket{0}$ and $\ket{1}$:

\begin{equation}
\ket{\psi} = \alpha\ket{0} + \beta\ket{1}
\end{equation}
where $\alpha, \beta \in \mathbb{C}$, and $\ket{0}$ and $\ket{1}$ are the vectors $\left[\begin{smallmatrix}1 \\ 0\end{smallmatrix}\right]$ and $\left[\begin{smallmatrix}0 \\ 1\end{smallmatrix}\right]$ respectively in the two-dimensional Hilbert space $\mathcal{H}^2$. $\ket{\psi}$ satisfies the normalization condition: $\braket{\psi|\psi} = |\alpha|^2 + |\beta|^2 = 1$; the coefficients $\alpha$ and $\beta$ are known as probability amplitudes, and their squared magnitudes indicate the probabilities of obtaining the corresponding basis states upon measurement, i.e., the outcome of a single qubit state is either 0 with probability $|\alpha|^2$ or 1 with probability $|\beta|^2$. Measurement on qubits is most typically performed in the computational basis, each qubit can be measured with respect to the standard states $\ket{0}$ and $\ket{1}$. Other bases can be used depending on the specific problem at hand.

When a qubit is measured, the measurement process is described by a set of projection operators. For the computational basis, the measurement operators are as follows:

\begin{align}    
M_0 = \ket{0}\bra{0} = \begin{bmatrix} 1 & 0 \\ 0 & 0 \end{bmatrix},
M_1 = \ket{1}\bra{1} = \begin{bmatrix} 0 & 0 \\ 0 & 1 \end{bmatrix}. 
\end{align}
Given a qubit state $\ket{\psi} = \alpha\ket{0} + \beta\ket{1}$, the probability of measuring the state $\ket{0}$ or $\ket{1}$ can be computed using the respective measurement operator:

\begin{align}    
P_0 = \bra{\psi}M_0\ket{\psi} = |\alpha|^2, \\
P_1 = \bra{\psi}M_1\ket{\psi} = |\beta|^2.
\end{align}
That is, the probability of obtaining state $\ket{0}$ is given by $P_0 = |\alpha|^2$ while the probability of obtaining state $\ket{1}$ is $P_1 = |\beta|^2$. The act of measuring a qubit causes it to collapse into a single state, and classical information is obtained in the form of a single classical bit, either 0 or 1. This process is inherently probabilistic, and the outcome of the measurement cannot be predicted with certainty; the state of a qubit cannot be directly observed without disturbing its state.

The quantum state of the system post-measurement is also now generally different from its initial state. The post-measurement state is given by:

\begin{align}
\ket{\psi_{m}} = \frac{M_m \ket{\psi}}{\sqrt{p(m)}}.
\end{align}
Here, $\ket{\psi_{m}}$ represents the post-measurement state, $M_m$ are the measurement operators corresponding to the chosen basis (e.g., the computational basis), $\ket{\psi}$ is the initial quantum state, and $p(m)$ is the probability of obtaining the measurement outcome $m$. The normalization factor $1/\sqrt{p(m)}$ ensures that the post-measurement state remains a valid quantum state with a total probability of 1.

\subsubsection{Qubit Representations}

Instead of column vectors, quantum states can also be represented using polar coordinates. Given a qubit state $\ket{\psi} = \alpha\ket{0} + \beta\ket{1}$, with $\alpha, \beta \in \mathbb{C}$, we can express these complex coefficients in terms of real-valued parameters:

\begin{align}
    \ket{\psi} = \cos \frac{\theta}{2} \ket{0} + e^{i\phi} \sin \frac{\theta}{2} \ket{1}, \\
    \theta, \phi \in \mathbb{R}.
\end{align}
 $\theta$ and $\phi$ can be interpreted as spherical coordinates, allowing the quantum state to be plotted as a point on the surface of a unit sphere in 3-dimensional space --- the \textit{Bloch sphere}. $\theta \in [0, \pi]$ and $\phi \in [0, 2\pi]$ are the polar and azimuthal angles, respectively. The basis states $\ket{0}$ and $\ket{1}$ correspond to the north and south poles of the sphere, while states with equal probability amplitudes (i.e., equal superpositions) lie on the equator. This geometric representation offers an intuitive way for visualizing qubits and their transformations.

\begin{figure}[!ht]
    \centering
    \centerline{\includegraphics[width=0.5\textwidth]{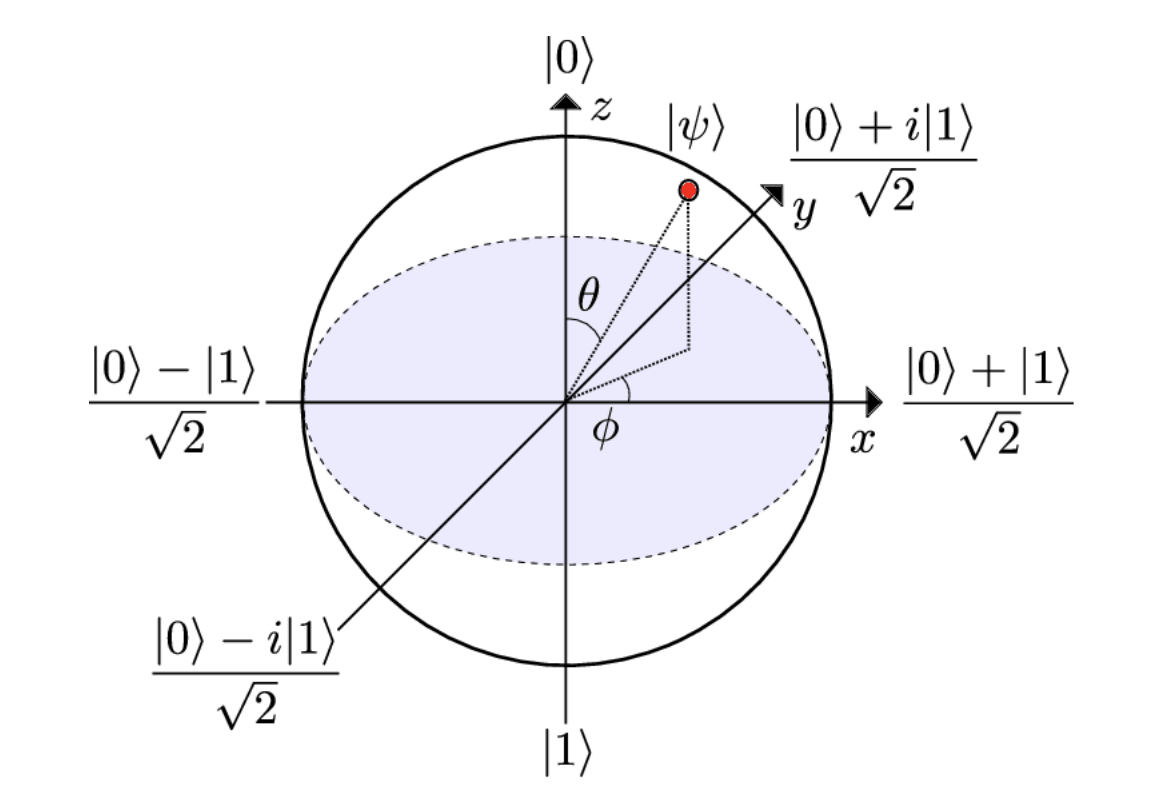}}
    \caption{Bloch sphere representation of the qubit state \cite{kockum2014quantum}.}
    \label{fig:blochsphere}
\end{figure}

\subsubsection{Quantum Gates and Circuits}

Quantum gates are the quantum equivalent of classical logic gates, which are used in classical digital circuits to manipulate bits. In a quantum computer, qubits are manipulated using quantum gates to perform quantum algorithms. A quantum gate is any unitary operator that acts on a single qubit or a set of qubits, the application of which produces a modification in the state of those qubits. These gates are represented as a unitary matrices; a single-qubit gate is represented by a $2\times2$ unitary matrix, a two-qubit gate is represented by a $4\times4$ unitary matrix. An $n$-qubit gate is thus represented by a $2^n$ matrix. A unitary matrix is any square matrix $U$ that satisfies $U^{\dagger}U = UU^{\dagger} = I$. This condition ensures that quantum gates are reversible and preserve the norm of the quantum state. Table \ref{tab:basicquantumgates} lists a few basic quantum gates.

\newsavebox{\pauliXa}
\newsavebox{\pauliXb}
\savebox{\pauliXa}{%
  \begin{quantikz}
    \qw & \gate{X} & \qw
  \end{quantikz}%
}
\savebox{\pauliXb}{%
  \begin{quantikz}
    \qw & \targ{} & \qw
  \end{quantikz}%
}
\newsavebox{\pauliX}
\savebox{\pauliX}{%
  \begin{tabular}{c}
    \usebox{\pauliXa} \\
    \text{or} \\
    \usebox{\pauliXb}
  \end{tabular}
}

\newsavebox{\pauliY}
\savebox{\pauliY}{%
  \begin{quantikz}
    \qw & \gate{Y} & \qw
  \end{quantikz}%
}

\newsavebox{\pauliZ}
\savebox{\pauliZ}{%
  \begin{quantikz}
    \qw & \gate{Z} & \qw
  \end{quantikz}%
}

\newsavebox{\hadamard}
\savebox{\hadamard}{%
  \begin{quantikz}
    \qw & \gate{H} & \qw
  \end{quantikz}%
}

\newsavebox{\phaseGate}
\savebox{\phaseGate}{%
  \begin{quantikz}
    \qw & \gate{P(\phi)} & \qw
  \end{quantikz}%
}
\newsavebox{\cnotGate}
\savebox{\cnotGate}{%
  \begin{quantikz}
    \qw & \ctrl{1} & \qw & \\
    \qw & \targ{} & \qw
  \end{quantikz}%
}

\newsavebox{\swapGate}
\savebox{\swapGate}{%
  \begin{quantikz}
    \qw & \swap{1} & \qw & \\
    \qw & \targX{} & \qw
  \end{quantikz}%
}

\newsavebox{\czGate}
\savebox{\czGate}{%
  \begin{quantikz}
    \qw & \ctrl{1} & \qw & \\
    \qw & \gate{Z} & \qw
  \end{quantikz}%
}

\begin{table}[ht!]
    \small
    \centering
    \caption{Examples of Basic Quantum Gates}
    \begin{tabular}{|p{1.5cm}|c|c|}
        \hline
        \textbf{Operator} & \textbf{Gate} & \textbf{Matrix} \\ \hline
        \multicolumn{3}{|l|}{\textbf{Single-Qubit Gates}} \\
        \hline

        Pauli-X & \usebox{\pauliX} & $X, \sigma_x = \begin{bmatrix} 0 & 1 \\ 1 & 0 \end{bmatrix}$ \\
        Pauli-Y & \usebox{\pauliY} & $Y, \sigma_y  = \begin{bmatrix} 0 & -i \\ i & 0 \end{bmatrix}$ \\
        Pauli-Z & \usebox{\pauliZ} & $Z, \sigma_z  = \begin{bmatrix} 1 & 0 \\ 0 & -1 \end{bmatrix}$ \\
        Hadamard & \usebox{\hadamard} & $H = \frac{1}{\sqrt{2}} \begin{bmatrix} 1 & 1 \\ 1 & -1 \end{bmatrix}$ \\
        Phase Gate & \usebox{\phaseGate} & $P(\phi) = \begin{bmatrix} 1 & 0 \\ 0 & e^{i\phi} \end{bmatrix}$ \\
        \hline
        \multicolumn{3}{|l|}{\textbf{Two-Qubit Gates}} \\
        \hline
        Controlled NOT (CNOT) & \usebox{\cnotGate} & $CX = \begin{bmatrix} 1 & 0 & 0 & 0 \\ 0 & 1 & 0 & 0 \\ 0 & 0 & 0 & 1 \\ 0 & 0 & 1 & 0 \end{bmatrix}$ \\
        SWAP Gate & \usebox{\swapGate} & $SWAP = \begin{bmatrix} 1 & 0 & 0 & 0 \\ 0 & 0 & 1 & 0 \\ 0 & 1 & 0 & 0 \\ 0 & 0 & 0 & 1 \end{bmatrix}$ \\
        Controlled Z (CZ) Gate & \usebox{\czGate} & $CZ = \begin{bmatrix} 1 & 0 & 0 & 0 \\ 0 & 1 & 0 & 0 \\ 0 & 0 & 1 & 0 \\ 0 & 0 & 0 & -1 \end{bmatrix}$ \\
        \hline
    \end{tabular}
    \label{tab:basicquantumgates}
\end{table}

The single qubit Pauli gates correspond to X, Y, and Z correspond to rotations of the qubit state around the x, y, and z axes of the Bloch sphere, respectively. Specifically, the Pauli-X gate corresponds to a rotation of the qubit state by $\pi$ radians around the x-axis of the Bloch sphere.
It is also often called the NOT gate, since it performs similarly to the classical NOT gate; it takes in a basis state $\ket{0}$ or $\ket{1}$ and returns the opposite state. The Pauli-Y and Z gates behave similarly; the Pauli-Y gate corresponds to a rotation of the qubit state by $\pi$ around the y-axis of the Bloch sphere while the Pauli-Z gate corresponds to a rotation of the qubit state by $\pi$ around the z-axis of the Bloch sphere. The Pauli gates are specialised cases of the generic rotation gates (Rx, Ry, Rz) in $\pi$ degrees. The Hadamard gate produces an equal superposition of the $\ket{0}$ and $\ket{1}$ states. The Phase gate is a parameterized gate that imparts a phase shift to the $\ket{1}$ state of a qubit while leaving the $\ket{0}$ state unchanged, thereby modifying the relative phase of the qubit's superposition without affecting its probability amplitudes. Superposition is discussed further in Section \ref{sec:superposition}.

The two-qubit gates perform operations on two or more qubits simultaneously, resulting in entanglement or changing the state of one qubit based on the state of another. The CNOT gate terms one of its two qubit inputs as the target, the other being the control. It performs a NOT operation on the target qubit if the control qubit is in the state $\ket{1}$. If the control qubit is in the state $\ket{0}$, the target qubit remains unchanged. The CNOT matrix presented in Table  \ref{tab:basicquantumgates} assumes the first qubit to be the control. The SWAP gate swaps the states of two qubits. Essentially, if one qubit is in state $\ket{0}$ and the other is in state $\ket{1}$, the gate will swap their states. The CZ applies a Z operation on the target qubit if the control qubit is in the state $\ket{1}$. If the control qubit is in the state $\ket{0}$, the target qubit remains unchanged. Other single and multiple qubit gates have also been used for various purposes.

\begin{figure}[]
    \centering
    \begin{quantikz}
        \qw & \gate{U} & \qw & = & & \gate{R_Z} & \gate{R_Y} & \gate{R_Z} & \qw
    \end{quantikz}
    \caption{Arbitrary single-qubit gate decomposition into Z, Y, Z rotations}
    \label{fig:universal_decomp}
\end{figure}
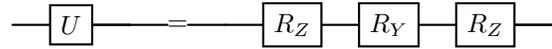

The notion of a universal quantum gate set enables the implementation of any quantum operation or circuit. A universal gate set consists of a collection of quantum gates, including both single and multi-qubit gates, which can be assembled and ordered to approximate any unitary transformation on qubits with arbitrary precision \cite{dawson2005solovay}. This property is crucial for harnessing the full potential of quantum computing, as it facilitates the construction of any quantum algorithm or circuit using a limited set of fundamental quantum gates. For example, it is possible to decompose any single-qubit gate into a sequence of three rotations around fixed axes, such as rotations around the Z, Y, and Z axes \cite{iten2016quantum}. Another widely recognized example is the combination of the CNOT gate (a two-qubit gate) and arbitrary single-qubit rotations, which together enable the creation of any quantum operation with a desired degree of accuracy \cite{vartiainen2004efficient}.


A quantum circuit serves as a blueprint for executing a series of operations or instructions on a quantum computer. Composed of quantum gates arranged sequentially, the circuit is interpreted from left to right, with each gate transforming the state of one or more qubits. Wires connect these gates, illustrating the flow of quantum information through the circuit. When no gate is present on a wire, the qubit state remains unchanged, effectively implying that the identity operator is acting on the qubit at that point.

\subsubsection{Superposition} \label{sec:superposition}

A key consequence of quantum mechanics is that quantum states are able to exist in a linear superposition, which is a fundamental departure from classical physics. A quantum system can exist in multiple states simultaneously, with the relative weights of these states determined by complex probability amplitudes. The ability of quantum systems to exist in superpositions results in a much higher level of expressivity in comparison with classical systems. 

Superposition can be manipulated using quantum gates to create more complex states. For example, a Hadamard gate can be used to put a qubit into an equal superposition of $\ket{0}$ and $\ket{1}$, given by the equation $H \ket{0} = \frac{1}{\sqrt{2}} (\ket{0} + \ket{1})$. This is the superposition of the $\ket{0}$ and $\ket{1}$ states, with equal probability amplitudes. Other gates, such as the Pauli-X gate or the phase gate, can also be used to manipulate the probability amplitudes and relative phase of a qubit in superposition. In addition, the superposition of multiple qubits can also be achieved by combining the superposition of individual qubits. For instance, a two-qubit system can be in a superposition of the four possible basis states: $\ket{\psi} = \alpha_{00} \ket{00} + \alpha_{01} \ket{01} + \alpha_{10} \ket{10} + \alpha_{11} \ket{11}$. The probability amplitudes $\alpha_{ij}$ determine the probability of measuring the corresponding basis state while satisfying the normalization condition $\braket{\psi|\psi} = 1$.

\subsubsection{Entanglement}

Quantum entanglement is the phenomenon in which quantum particles become interconnected and are described in relation to each other, regardless of their spatial separation. In a bipartite system, this “interconnectedness” is such that the state of one particle has an instantaneous effect on the state of the other particle. A system is said to be in an entangled state if it cannot be factorized into a product of individual particle states.

\begin{figure}[ht!]
    \centering
    \begin{quantikz}
        \lstick{$\ket{0}$} & \gate{H} & \ctrl{1} & \qw & \rstick[wires=2]{$\ket{\Phi^+} = \frac{1}{\sqrt{2}}(\ket{00} + \ket{11})$} \qw \\
        \lstick{$\ket{0}$} & \qw & \targ{} & \qw & \qw
    \end{quantikz}
    \caption{Quantum circuit for generating the Bell state {$\ket{\Phi^+}$}}
    \label{fig:bellstatecircuit}
\end{figure}
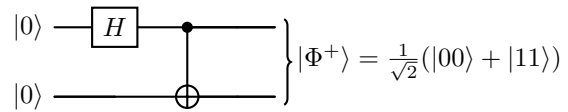

One of the most well-known examples of entangled states are the Bell states, also known as the EPR pairs, which are maximally entangled states of two qubits. These states exhibit perfect correlations, meaning that the result of measuring one qubit instantly determines the outcome of measuring the other qubit. There are four distinct Bell states, which form an orthonormal basis:

\begin{itemize}
    \item $\ket{\Phi^+} = \frac{1}{\sqrt{2}}(\ket{00} + \ket{11})$
    \item $\ket{\Phi^-} = \frac{1}{\sqrt{2}}(\ket{00} - \ket{11})$
    \item $\ket{\Psi^+} = \frac{1}{\sqrt{2}}(\ket{01} + \ket{10})$
    \item $\ket{\Psi^-} = \frac{1}{\sqrt{2}}(\ket{01} - \ket{10})$
\end{itemize}
These Bell states can be generated using a simple quantum circuit consisting of a Hadamard gate (H) followed by a CNOT gate. The circuit diagram for generating the first Bell state $\ket{\Phi^+}$ is shown in Figure \ref{fig:bellstatecircuit}. Although this is not the only way to produce the Bell states, it is a simple method as it uses the computational basis as input. The H gate is applied to the first qubit, while the CNOT gate uses the first qubit as control and the second qubit as target. The state $\ket{\Phi^+}$ has a 50\% probability of being in either the $\ket{00}$ state or the $\ket{11}$ state once measured. After this state is created and upon measuring the first qubit, if the result is 0, the second qubit will be immediately known to be 0 due to the entanglement. Similarly, if the first qubit is measured to be 1, the second qubit will also be 1. Table \ref{tab:bellstates} illustrates the transformation of the input states to the corresponding Bell states after the circuit is applied. The Bell state pairs are essential for several quantum information and communication protocols, such as quantum teleportation and superdense coding.

\begin{table}[ht!]
    \small
    \centering
    \caption{Generation of Bell states}
    \begin{tabular}{c|c|c|c}
        Input & H & CNOT & Bell State\\ \hline
        $\ket{00}$ & $\frac{1}{\sqrt{2}}(\ket{00}+\ket{10})$ & $\frac{1}{\sqrt{2}}(\ket{00}+\ket{11})$ & $\ket{\Phi^+}$ \\
        $\ket{10}$ & $\frac{1}{\sqrt{2}}(\ket{00}-\ket{10})$ & $\frac{1}{\sqrt{2}}(\ket{00}-\ket{11})$ & $\ket{\Phi^-}$\\
        $\ket{01}$ & $\frac{1}{\sqrt{2}}(\ket{01}+\ket{11})$ & $\frac{1}{\sqrt{2}}(\ket{01}+\ket{10})$ & $\ket{\Psi^+}$\\
        $\ket{11}$ & $\frac{1}{\sqrt{2}}(\ket{01}-\ket{11})$ & $\frac{1}{\sqrt{2}}(\ket{01}-\ket{10})$ & $\ket{\Psi^-}$\\ \hline
    \end{tabular}
    \label{tab:bellstates}
\end{table}

\subsubsection{Density Matrices}

Quantum state vector representations are best suited for pure states --- states with complete knowledge of their quantum characteristics. In contrast, density matrices allow for the representation of both pure and mixed states, the latter referring to statistical combinations of various quantum states that generally arise from partial or incomplete information about the system.

The density matrix $\rho$ of a quantum system is defined as an operator that acts on the system's Hilbert space. For a system in a pure state described by a normalized state vector $\ket{\psi}$, the corresponding density matrix is given by the outer product $\ket{\psi}\bra{\psi}$. For a mixed state composed of possible states $\ket{\psi_i}$ each occurring with a probability $p_i$, the density matrix is a sum of the individual state's density matrices, weighted by their respective probabilities: $\rho = \sum_i p_i \ket{\psi_i}\bra{\psi_i}$. The key characterization of $\rho$ is that it is a positive semi-definite operator with a trace equal to one. This aligns with the probabilistic interpretation of quantum mechanics, where the probability of a system being in a particular state is given by the trace of the product of the density matrix and the projector onto the state. 

The reduced density operator describes the state of a subsystem by "tracing out" or ignoring the other parts of the system. Consider some composite quantum system consisting of subsystems $A$ and $B$ in a joint state represented by the density matrix $\rho_{AB}$. Investigating the properties of subsystem $A$, independent from subsystem $B$, is done by constructing the reduced density operator for $A$. This is accomplished by taking the partial trace over subsystem $B$, written as $\rho_A = Tr_B(\rho_{AB})$, where $Tr_B$ denotes the trace over subsystem $B$'s degrees of freedom. The reduced density operator gives us all the measurable information about the subsystem of interest. The subsystem behaves as if it were in a mixed state given by the reduced density operator, irrespective of whether the overall system is in a pure or mixed state.

\subsection{Machine Learning}

Machine learning, at its core, revolves around the concept of enabling machines to learn from data, whether it's to make predictions, identify patterns, or take specific actions. The domain is vast and is commonly divided into three main approaches based on the nature of the learning process and the kind of data available: supervised, unsupervised, and reinforcement learning.

\textbf{Supervised Learning:} In supervised learning, an algorithm learns from a labeled dataset containing input-output pairs $(\textbf{x}, y)$. The goal is to learn a function $f: \textbf{x} \rightarrow y$ that maps input features to output labels.  The learning process involves minimizing a loss function $L(y, f(\textbf{x}; \theta))$, which measures the discrepancy between the true output $y$ and the predicted output $f(\textbf{x}; \theta)$, for a given parameter set $\theta$. This $\theta$ is then optimized iteratively, often through gradient descent-based methods. Two common supervised learning tasks are regression, where the output is a continuous value, and classification, where the output is a discrete class label. An effective model has thus learned a $\theta$ that produces either accurate continuous predictions for regression tasks or correct class assignments for classification tasks a sufficient proportion of the time when presented with generalized, unseen data.

\textbf{Unsupervised Learning:} Unsupervised learning algorithms learn patterns or structures from an unlabeled dataset $\textbf{x}$. The goal is to capture underlying patterns, structures, or representations in the data by identifying relationships and dependencies within the dataset. In this context, “learning” can be understood as finding a transformation function $g: \textbf{x} \rightarrow \textbf{z}$, where $\textbf{z}$ represents a lower-dimensional, structured, or otherwise more informative representation of the input data $\textbf{x}$. The learning process involves optimizing an objective function $O(g(\textbf{x}; \phi))$, which measures the quality of the transformation based on some criterion, such as preserving the intrinsic structure of the data or maximizing the compactness of clusters. The function parameters $\phi$ are iteratively optimized using unsupervised learning techniques, such as k-means clustering, hierarchical clustering, or principal component analysis (PCA). An effective unsupervised learning model can thus generalize well to new, unseen data by accurately representing the underlying patterns and structure within the discovered data distributions.

\textbf{Reinforcement Learning:} Reinforcement learning involves an agent interacting with an environment to learn optimal actions through a trial-and-error process. Unlike supervised learning, which relies on labeled data to learn input-output mappings, and unsupervised learning, which seeks to find hidden structures in unlabeled data, reinforcement learning focuses on learning through interactions and feedback in the form of rewards or penalties. Central to reinforcement learning is the Markov Decision Process (MDP) framework, a model where an agent makes decisions in states based on actions, transitions, and rewards, and operates under the premise that future states are dependent only on the current state and action. The agent seeks to learn an optimal policy function \(\pi: \textbf{s} \rightarrow \textbf{a}\), which maps states \(\textbf{s}\) to actions \(\textbf{a}\), to maximize the expected cumulative reward \(J(\pi) = \mathbb{E}\left[\sum_{t=0}^\infty \lambda^t R_t | \pi\right]\), where \(R_t\) is the reward at time \(t\) and \(\lambda \in [0, 1]\) is the discount factor (the infinite-horizon discounted return). The learning process often involves updating the policy or value function through algorithms such as Q-learning, SARSA, or policy gradients, using techniques like the action-value function \(Q^\pi(s, a)\) and Temporal Difference (TD) errors \(\delta_t\). By leveraging the MDP framework, reinforcement learning algorithms enable agents to navigate complex and uncertain environments to maximize long-term rewards.

\subsubsection{Gradient Descent}

Gradient descent is a widely prevalent optimization algorithm in machine learning and deep learning applications, serving as the backbone for minimizing a differentiable loss function $L(\boldsymbol{w})$ with respect to its parameters $\boldsymbol{w}$. Parameters are updated iteratively in the direction of the negative gradient of the loss function to minimize the loss:

\begin{equation}
\boldsymbol{w}_{t+1} = \boldsymbol{w}_t - \eta \nabla L(\boldsymbol{w}_t),
\end{equation}
where $\eta$ is the learning rate and $\nabla L(\boldsymbol{w}_t)$ represents the gradient of the loss function $L$ with respect to the parameters $\boldsymbol{w}$ at iteration $t$. Variants of gradient descent differ in how they compute the gradient:

\textbf{Batch Gradient Descent (BGD)}: The gradient is computed using the entire dataset. This approach can be computationally expensive, especially for large datasets:
\begin{equation}
\nabla L(\boldsymbol{w}_t) = \frac{1}{N} \sum_{i=1}^N \nabla L_i(\boldsymbol{w}_t),
\end{equation}
where $N$ is the number of training examples and $L_i(\boldsymbol{w}_t)$ is the loss for the $i^{th}$ example.

\textbf{Stochastic Gradient Descent (SGD)}: The gradient is approximated using a single randomly-selected training example, which leads to faster iterations but higher variance in the parameter updates:
\begin{equation}
\nabla L(\boldsymbol{w}_t) = \nabla L_i(\boldsymbol{w}_t),
\end{equation}
where $i$ is a randomly-selected index in the range $[1, N]$.

\textbf{Mini-batch Gradient Descent (MBGD)}: A compromise between BGD and SGD, the gradient is computed using a mini-batch of $b$ training examples, which balances the trade-off between computational efficiency and update variance:
\begin{equation}
\nabla L(\boldsymbol{w}_t) = \frac{1}{b} \sum_{i=1}^b \nabla L_i(\boldsymbol{w}_t),
\end{equation}
where $b$ is the mini-batch size and the mini-batch is randomly sampled from the training dataset.
Gradient descent can be further improved by incorporating adaptive learning rates or momentum-based updates, leading to variants like AdaGrad, RMSprop, Adam, and others.


\subsubsection{Key Algorithms and Techniques}


\textbf{Linear Regression:} Linear regression is a fundamental technique in machine learning and statistics, modeling the relationship between a dependent variable $y$ and one or more independent variables $\textbf{x}$. The goal is to find the best-fitting linear function that can predict the dependent variable from the independent variables. The linear model can be expressed as $y = \textbf{w}^\intercal \textbf{x} + b$, where $\textbf{w}$ is the weight vector and $b$ is the bias term. The objective function for linear regression is the least-squares loss function, which aims to minimize the sum of squared differences between the observed responses and the predictions of the linear function:

\begin{equation}
L(\textbf{w}, b) = \frac{1}{N}\sum_{i=1}^{N}(\textbf{w}^\intercal\textbf{x}_i + b - y_i)^2
\end{equation}

The optimal weights $\textbf{w}$ and bias term $b$ are found via gradient descent or by other optimization algorithms. In the case of a single independent variable, the linear function forms a straight line, and for multiple independent variables, it forms a hyperplane. Linear regression can be extended to incorporate regularization techniques like Ridge or Lasso regression, which introduce a penalty term to the objective function, helping to prevent over-fitting as well as improving generalization.

\textbf{Logistic Regression:} Logistic regression is a widely used binary classification algorithm that models the probability of an instance belonging to a specific class. It builds upon the concept of linear regression by using the logistic function, also known as the sigmoid function, to model the relationship between the dependent variable and the independent variables, given by:

\begin{equation}
\sigma(z) = \frac{1}{1 + e^{-z}}
\end{equation}

where $z = \textbf{w}^\intercal\textbf{x} + b$. This function transforms the linear combination of input features into a probability value between 0 and 1, which can then be thresholded to make binary classification decisions. The objective function for logistic regression is the cross-entropy loss, which measures the discrepancy between the predicted probabilities and the true class labels:

\begin{equation}
    \begin{aligned}
        L(\textbf{w}, b) = & -\frac{1}{N}\sum_{i=1}^{N} [y_i \log(\sigma(\textbf{w}^\intercal\textbf{x}_i + b)) \\ & + (1 - y_i) \log(1 - \sigma(\textbf{w}^\intercal\textbf{x}_i + b))].
    \end{aligned}
\end{equation}

Logistic regression can be extended to multi-class classification problems using techniques such as one-vs-rest or one-vs-one, or by adopting the softmax function in place of the sigmoid function.

\textbf{Support Vector Machines (SVM):} SVMs are a class of powerful and versatile classification algorithms that seek to find the optimal decision boundary, or hyperplane, that maximizes the margin between different classes. This is achieved by focusing on the instances that are closest to the decision boundary, known as support vectors, which play a critical role in determining the hyperplane's position and orientation. Given a training set $(\textbf{x}_i, y_i)$, where $y_i \in {-1, 1}$, the primal problem of SVM is formulated as follows:

\begin{equation}
    \begin{aligned}
        \min_{\textbf{w}, b} \frac{1}{2} |\textbf{w}|^2 \quad \text{s.t.} \quad y_i (\textbf{w}^\intercal \textbf{x}_i + b) \geq 1, \forall i
    \end{aligned}
\end{equation}

This optimization problem seeks to minimize the norm of the weight vector, which is equivalent to maximizing the margin between the classes. The margin can be understood in terms of the geometric margin, which measures the distance between the decision boundary and the nearest data points, and the functional margin, which represents the confidence in a classification decision. The dual problem can be derived using Lagrange multipliers, leading to an optimization problem that can be solved using the Sequential Minimal Optimization (SMO) algorithm, gradient descent, or other quadratic programming methods. The dual problem is expressed as follows:

\begin{equation}
    \begin{aligned}
        \max_{\boldsymbol{\alpha}} \sum_{i=1}^{N} \alpha_i - \frac{1}{2} \sum_{i=1}^{N} \sum_{j=1}^{N} \alpha_i \alpha_j y_i y_j K(\textbf{x}_i, \textbf{x}_j) \\
        \text{s.t.} \quad 0 \leq \alpha_i \leq C, \quad \sum_{i=1}^{N} \alpha_i y_i = 0.
    \end{aligned}
    \label{eqn:bg_svm}
\end{equation}
Here, $\boldsymbol{\alpha} = (\alpha_1, \alpha_2, \dots, \alpha_N)$ are the Lagrange multipliers and $C$ is a regularization parameter that controls the trade-off between maximizing the margin and minimizing the classification error.

In Equation \ref{eqn:bg_svm}, the term $K(\textbf{x}_i, \textbf{x}_j)$ represents the kernel function, used to transform data into higher-dimensional spaces, enabling the algorithm to draw more complex decision boundaries, thus handling non-linearly separable problems. The kernel function essentially maps the input vectors into a higher-dimensional feature space and computes the dot product in that space. This is equivalent to the term $(\textbf{x}_i^\intercal, \textbf{x}_j)$ in the non-kernelized version of the SVM, but with the ability to capture non-linear relations. Common kernel functions include the linear, polynomial, radial basis function (RBF), and sigmoid kernels. The choice of kernel function and its parameters significantly impacts the performance of the SVM classifier.

\textbf{Decision Trees and Random Forests:} Decision trees are a popular machine learning algorithm that learn a hierarchical structure of if-else rules to predict the target variable. The tree is constructed by recursively splitting the data based on the feature that provides the highest information gain or the lowest Gini impurity. Information gain $IG$ measures the reduction in entropy before and after the split:

\begin{equation}
IG(S, A) = H(S) - \sum_{v \in \text{Values}(A)} \frac{|S_v|}{|S|} H(S_v)
\end{equation}
where $S$ represents the set of samples, $A$ is the attribute being tested, $S_v$ is the subset of samples where attribute $A$ has value $v$, and $H(S)$ denotes the entropy of the set $S$. Entropy can be calculated using the following formula:

\begin{equation}
H(S) = -\sum_{i=1}^{C} p_i \log_2 p_i
\end{equation}
where $p_i$ is the proportion of samples in class $i$ and $C$ is the number of classes.

The Gini impurity is given by:

\begin{equation}
G(\textbf{x}) = 1 - \sum_{i=1}^{C} p_i^2
\end{equation}
where $p_i$ is the proportion of samples in class $i$ and $C$ is the number of classes.

Random forests are an extension of decision trees and represent an ensemble method, which combines multiple base learners to improve predictive performance. Ensemble methods like bagging (bootstrap aggregating) and boosting work by aggregating the predictions of several individual models. Bagging is employed in random forests, where multiple decision trees are trained on random subsets of the dataset with replacement. The final prediction is obtained by averaging (regression) or voting (classification) the outputs of these individual trees. Boosting is a sequential ensemble method that iteratively adjusts the weights of the training instances based on the errors made by the previous learners, placing more emphasis on difficult-to-classify instances. Examples of boosting algorithms include AdaBoost and Gradient Boosting. Ensemble methods, in general, take advantage of the strengths of multiple base models, making the final model more robust and accurate than individual learners.

\textbf{Clustering:} Clustering algorithms aim to group data points into clusters based on similarity or distance metrics, which allows for the identification of natural patterns or structures in the data; a key application of unsupervised learning. Two popular clustering methods are the k-means algorithm and KNN algorithm.

K-means is an iterative algorithm that seeks to minimize the within-cluster sum of squared distances by updating cluster centroids and reassigning data points. The objective function for k-means is given by:

\begin{equation}
    J = \sum_{j=1}^{k} \sum_{i=1}^{n} \|\textbf{x}_i^{(j)} - \textbf{c}_j\|^2
\end{equation}
%
where \( k \) is the total number of clusters, \( n \) is the number of data points, \( \textbf{x}_i^{(j)} \) is the \( i \)-th data point in the \( j \)-th cluster, and \( \textbf{c}_j \) is the centroid of the \( j \)-th cluster. The algorithm begins by initializing the centroids randomly and proceeds iteratively through two main steps: assignment and update. In the assignment step, each data point is assigned to the nearest centroid, and in the update step, the centroids are recomputed as the mean of the points in each cluster.

KNN is a distance-based, non-parametric method used for both classification and regression tasks. In classification, given a query point $\textbf{q}$, the algorithm identifies the $k$ nearest points to $\textbf{q}$ in the feature space and assigns the majority class label among those neighbors. For regression, the predicted value is the average of the target values of the $k$ nearest neighbors. Distance metrics such as Euclidean distance, Manhattan distance, or cosine similarity are used to measure the similarity between instances.

Other popular clustering algorithms include DBSCAN (Density-Based Spatial Clustering of Applications with Noise), which groups points based on density, and hierarchical clustering, which creates a tree-like structure of nested clusters.

\subsubsection{Neural Networks}

\textbf{Feed-forward Networks:} Feed-forward networks, also known as multilayer perceptrons (MLPs), are the most basic type of neural network. They comprise multiple layers of interconnected neurons, where each neuron calculates a weighted sum of its inputs, then applies a nonlinear activation function. The output of neuron $j$ in layer $l$ is expressed as:

\begin{equation}
a_j^{(l)} = f\left(\sum_{i=1}^{N^{(l-1)}} w_{ji}^{(l)} a_i^{(l-1)} + b_j^{(l)}\right)
\end{equation}
Here, $a_j^{(l)}$ is the activation of neuron $j$ in layer $l$, $f(\cdot)$ is the activation function, $w_{ji}^{(l)}$ is the weight connecting neuron $i$ in layer $(l-1)$ to neuron $j$ in layer $l$, and $b_j^{(l)}$ is the bias of neuron $j$ in layer $l$.

Various different activation functions have been used within neural networks. The sigmoid and tanh activation functions were prevalent in early architectures. However, they have become less common as they suffer from the vanishing or exploding gradients problem, where gradients can become increasingly small or exponentially large during backpropagation, making it difficult for the network to learn \cite{glorot2010understanding}. This issue is particularly pronounced in deep neural networks with many layers. The Rectified Linear Unit (ReLU) activation function and its variants, such as Leaky ReLU or SELU, are now commonly used for their ability to mitigate this issue. These have sparse activation; not all neurons are activated simultaneously, leading to more efficient training and better generalization in deep networks. Additionally, ReLU is piecewise linear, making it computationally faster compared to sigmoid and tanh functions.

Feed-forward neural networks can learn complex input-output relationships and have been applied in various machine learning tasks, such as regression, classification, and pattern recognition. However, they lack the specialized structures found in more advanced deep learning models like convolutional and recurrent neural networks, which excel at handling spatial and temporal data, respectively.

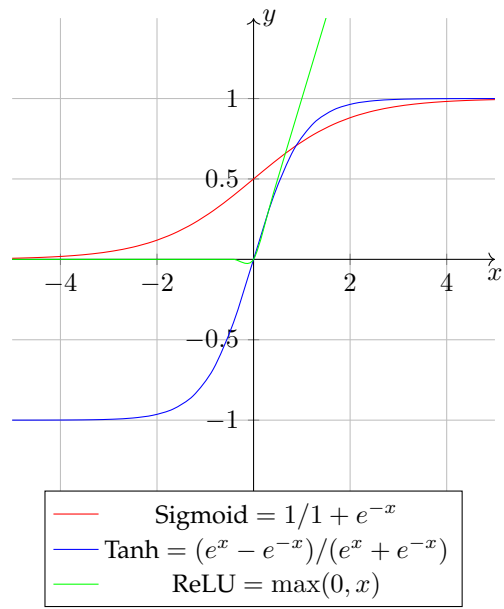
\begin{figure}[!h]
    \centering
    \begin{tikzpicture}[
      declare function={
        sigmoid(\x) = 1 / (1 + exp(-\x));
        relu(\x) = max(0, \x);
        leakyrelu(\x) = ifthenelse(\x > 0, \x, 0.01 * \x);
      }
    ]
    \begin{axis}[
      width=8cm,
      height=8cm,
      ymin=-1.5, ymax=1.5,
      xmin=-5, xmax=5,
      axis lines=center,
      axis line style={->},
      x label style={at={(axis description cs:1,0.5)},anchor=north},
      y label style={at={(axis description cs:0.5,1)},anchor=west},
      xlabel={$x$},
      ylabel={$y$},
      legend style={at={(axis description cs:0.5,-0.1)},anchor=center},
      grid=both,
      ytick={-1, -0.5, 0.5, 1},
    ]
    
    \addplot[color=red, smooth, domain=-5:5]{sigmoid(x)}; \addlegendentry{Sigmoid $= 1/1+e^{-x}$}
    \addplot[color=blue, smooth, domain=-5:5]{tanh(x)}; \addlegendentry{Tanh $=(e^{x}-e^{-x})/(e^{x}+e^{-x})$}
    \addplot[color=green, smooth, domain=-5:5]{relu(x)}; \addlegendentry{ReLU $=\max(0, x)$}
    \end{axis}
    \end{tikzpicture}
    \caption{Common Activation Functions}
\end{figure}

\textbf{Backpropagation:} The backpropagation algorithm has become the fundamental process by which neural networks are trained. It efficiently computes the gradients of the loss function concerning the network's weights and biases, enabling gradient-based optimization techniques to update the parameters and minimize the loss. The challenge in calculating gradients in a neural network stems from the large number of parameters and the complex, nested nature of the functions involved. Due to these intricacies, it is difficult to determine how a small change in a single parameter will impact the overall loss.

Given a differentiable loss function $L(\textbf{W}, \textbf{b})$, where $\textbf{W}$ denotes the weights and $\textbf{b}$ denotes the biases, the goal is to compute the gradient of the loss function with respect to each parameter, i.e., $\partial L / \partial w_{ji}^{(l)}$ and $\partial L / \partial b_j^{(l)}$. Backpropagation achieves this via the chain rule of calculus, which computes the gradients in a layer-by-layer manner, starting from the output layer and moving backward through the network. By leveraging the chain rule and the feed-forward structure of neural networks, backpropagation efficiently calculates the gradients and overcomes the challenges posed by the complex network architecture.

For each neuron $j$ in layer $l$, the error term $\delta_j^{(l)}$ is defined as the product of the gradient of the activation function, $f'(\textbf{z}_j^{(l)})$, and the weighted sum of errors from the subsequent layer:

\begin{equation}
\delta_j^{(l)} = f'(\textbf{z}_j^{(l)}) \sum_{k=1}^{N^{(l+1)}} w_{kj}^{(l+1)} \delta_k^{(l+1)}
\end{equation}
where $\textbf{z}_j^{(l)} = \sum_{i=1}^{N^{(l-1)}} w_{ji}^{(l)} a_i^{(l-1)} + b_j^{(l)}$ is the weighted input to neuron $j$ in layer $l$, and $N^{(l+1)}$ is the number of neurons in layer $(l+1)$. Using the error term $\delta_j^{(l)}$, the gradient of the loss function with respect to the parameters can be computed as:

\begin{equation}
\frac{\partial L}{\partial w_{ji}^{(l)}} = a_i^{(l-1)} \delta_j^{(l)}
\end{equation}

\begin{equation}
\frac{\partial L}{\partial b_j^{(l)}} = \delta_j^{(l)}
\end{equation}
Once the gradients are computed, the parameters are updated via gradient descent:

\begin{equation}
w_{ji}^{(l)} \leftarrow w_{ji}^{(l)} - \eta \frac{\partial L}{\partial w_{ji}^{(l)}}
\end{equation}

\begin{equation}
b_j^{(l)} \leftarrow b_j^{(l)} - \eta \frac{\partial L}{\partial b_j^{(l)}}
\end{equation}
where $\eta$ is the learning rate. This iterative process of forward propagation, gradient computation using backpropagation, and parameter updates continues until a stopping criterion, such as a maximum number of epochs or a threshold on the improvement in the loss function, is reached.

\textbf{Convolutional Neural Networks (CNNs):} CNNs are a specialized type of neural network specifically designed for tasks such as image recognition and computer vision applications. They are composed of a series of layers, including convolutional layers, pooling layers, and fully connected layers.

In a convolutional layer, the value for a neuron in the \(k\)-th feature map at position \((i, j)\) is derived by superimposing a filter onto the corresponding region of the input (or the preceding feature map). This filter has its own set of weights, and it moves across the input spatially. For each position, it multiplies its weights with the corresponding values in the input, summing them up to produce a single value. A bias is then added to this sum. The result, after possibly passing through an activation function, forms a single value in the output feature map at the current position. This process continues as the filter “slides” over the entire input, creating a new feature map that represents the presence of certain features or patterns in the input as detected by the filter. Different filters will detect different features, and thus, a convolutional layer often consists of multiple feature maps, each corresponding to a different filter.

Pooling layers, such as max pooling or average pooling, serve to aggregate information from local regions, reducing the spatial dimensions of the feature maps while preserving shift-invariance. These layers are typically placed between two convolutional layers. For each feature map, pooling can be described as taking a local neighborhood around a certain location \((i, j)\) and applying a pooling operation, like taking the maximum or average of values in that neighborhood.

Finally, fully connected layers are used to integrate the features extracted by the previous layers and produce the final output, such as class probabilities in the case of classification tasks. This architecture, with its emphasis on local connectivity and spatial hierarchy, allows CNNs to efficiently process and learn from high-dimensional data, such as images, in a way that traditional feed-forward networks cannot.

\subsubsection{Model Evaluation Metrics}

\textbf{Accuracy:} Accuracy is the proportion of correctly classified instances out of the total instances:

\begin{equation}
\text{Accuracy} = \frac{\text{Number of correct predictions}}{\text{Total number of predictions}}
\end{equation}
Accuracy is a commonly used metric for classification tasks, but it may not be ideal when dealing with imbalanced datasets. In such cases, the model might have high accuracy by simply predicting the majority class, but it might not accurately identify the minority class instances, which are often the more important ones.

\textbf{Precision, Recall, and F1 Score:} Precision is the proportion of true positive predictions out of the total positive predictions, while recall is the proportion of true positive predictions out of the total actual positive instances. The F1 score is the harmonic mean of precision and recall:

\begin{equation}
\text{Precision} = \frac{\text{True Positives}}{\text{True Positives} + \text{False Positives}}
\end{equation}
\begin{equation}
\text{Recall} = \frac{\text{True Positives}}{\text{True Positives} + \text{False Negatives}}
\end{equation}
\begin{equation}
\text{F1 Score} = \frac{2 \times \text{Precision} \times \text{Recall}}{\text{Precision} + \text{Recall}}
\end{equation}
Precision and recall are particularly useful when evaluating models on imbalanced datasets. High precision means that the model has a low false positive rate, which is crucial when false positives are costly (e.g., spam detection). High recall means that the model has a low false negative rate, which is important when false negatives are more critical (e.g., cancer detection). 

The F1 score provides a single number that balances the trade-off between precision and recall; it assumes equal importance for both, which may not be suitable for all applications.

The choice of evaluation metric depends on the specific application and the relative importance of false positives, false negatives, and the class distribution. In the unsupervised learning context, the mentioned metrics may not adequately reflect model performance due to the absence of class labels. In such cases, alternative metrics have been developed to evaluate the underlying structure or relationships captured by the models without relying on ground truth labels. Some prominent unsupervised learning metrics include mutual information, adjusted Rand index, and variation of information, to name a few. These measures evaluate the quality of the learned structure or patterns by assessing their similarity to an unknown ground truth or by assessing the internal consistency of the discovered structure itself.

\end{document}